\documentclass[final,3p,times]{elsarticle}

\journal{Neural Networks}
\usepackage{amsfonts}
\usepackage{amsmath}
\usepackage{amssymb}
\usepackage{mathtools}   
\usepackage{amsthm}
\usepackage{booktabs}
\usepackage{array,multirow,multicol}
\usepackage{longtable}
\usepackage{rotating}
\usepackage{siunitx}     

\usepackage[font=small,labelfont=bf]{caption}
\usepackage{subcaption}  

\usepackage{xfrac}
\usepackage{bm}
\usepackage{bbm}
\usepackage{wrapfig}
\usepackage{arydshln}
\usepackage{algorithm}
\usepackage[noend]{algpseudocode}
\usepackage{cancel}
\usepackage{physics}     
\usepackage{microtype}
\usepackage{textcomp}

\PassOptionsToPackage{hyphens}{url}
\usepackage{url}
\usepackage[colorlinks=false,pdfpagelabels]{hyperref}
\usepackage[capitalize,noabbrev]{cleveref}

\usepackage{amsmath,amsfonts,bm}

\def\eqref#1{equation~\ref{#1}}

\def\1{\bm{1}}

\def\va{{\bm{a}}}
\def\vb{{\bm{b}}}
\def\vc{{\bm{c}}}

\def\vg{{\bm{g}}}

\def\vm{{\bm{m}}}

\def\vq{{\bm{q}}}
\def\vr{{\bm{r}}}

\def\vu{{\bm{u}}}
\def\vv{{\bm{v}}}
\def\vw{{\bm{w}}}
\def\vx{{\bm{x}}}

\def\vpsi{{\bm{\psi}}}

\def\vomega{{\bm{\omega}}}
\def\vxi{{\bm{\xi}}}

\def\vtheta{{\bm{\theta}}}

\def\mB{{\bm{B}}}

\def\mG{{\bm{G}}}
\def\mH{{\bm{H}}}
\def\mI{{\bm{I}}}

\def\mM{{\bm{M}}}

\def\mQ{{\bm{Q}}}

\def\mS{{\bm{S}}}
\def\mT{{\bm{T}}}
\def\mU{{\bm{U}}}
\def\mV{{\bm{V}}}
\def\mW{{\bm{W}}}

\def\mXi{{\bm{\Xi}}}

\newcommand{\E}{\mathbb{E}}

\newcommand{\R}{\mathbb{R}}

\newcommand{\esssup}{\mathop{\mathrm{ess\,sup}}}

\newcommand{\supp}{\mathop{\mathrm{supp}}}

\biboptions{sort&compress}

\theoremstyle{plain}
\newtheorem{theorem}{Theorem}[section]
\newtheorem{proposition}[theorem]{Proposition}
\newtheorem{lemma}[theorem]{Lemma}

\newtheorem{corollary}[theorem]{Corollary}

\theoremstyle{definition}
\newtheorem{definition}[theorem]{Definition}
\newtheorem{assumption}[theorem]{Assumption}
\newtheorem{condition}[theorem]{Condition}

\theoremstyle{remark}

\newcommand{\lammax}{$\lambda_\mathrm{max}$}
\newcommand{\lamneg}{$|\lambda^{-}_\mathrm{max}|$}

\usepackage[most]{tcolorbox}
\tcbuselibrary{skins,breakable}
\newtcolorbox[auto counter, number within=section]{resultbox}[2][]{%
  enhanced, breakable, colback=white, colframe=black,
  fonttitle=\bfseries, title={Box~\thetcbcounter: #2}, #1}

\usepackage{lipsum}

\let\parencite=\citep
\let\textcite=\citet

\begin{document}

\begin{frontmatter}



\title{Training Instabilities Induce Flatness Bias in Gradient Descent} 

\author{\corref{cor1} Lawrence Wang}  
\ead{lawrence.wang@exeter.ox.ac.uk} 
\author{Stephen J. Roberts}
\ead{sjrob@robots.ox.ac.uk}
\cortext[cor1]{Corresponding author. }

\affiliation{organization={Department of Engineering, University of Oxford},
            addressline={Parks Road}, 
            city={Oxford},
            postcode={OX1 3PJ}, 
            state={Oxfordshire},
            country={United Kingdom}
            }

\begin{abstract}
Classical analyses of gradient descent (GD) define a stability threshold based on the largest eigenvalue of the loss Hessian, often termed \emph{sharpness}. When the learning rate lies below this threshold, training is stable and the loss decreases monotonically. Yet, modern deep networks often achieve their best performance beyond this regime.

We demonstrate that such \emph{instabilities} induce an implicit bias in GD, driving parameters toward flatter regions of the loss landscape and thereby improving generalization. The key mechanism is the \emph{Rotational Polarity of Eigenvectors} (RPE), a geometric phenomenon in which the leading eigenvectors of the Hessian rotate during training instabilities. These rotations, which increase with learning rates, promote exploration and provably lead to flatter minima. 

This theoretical framework extends to stochastic GD, where instability-driven flattening persists and its empirical effects outweigh minibatch noise. Finally, we show that restoring instabilities in Adam further improves generalization. 

Together, these results establish and understand the constructive role of training instabilities in deep learning.
\end{abstract}

\begin{graphicalabstract}
\includegraphics{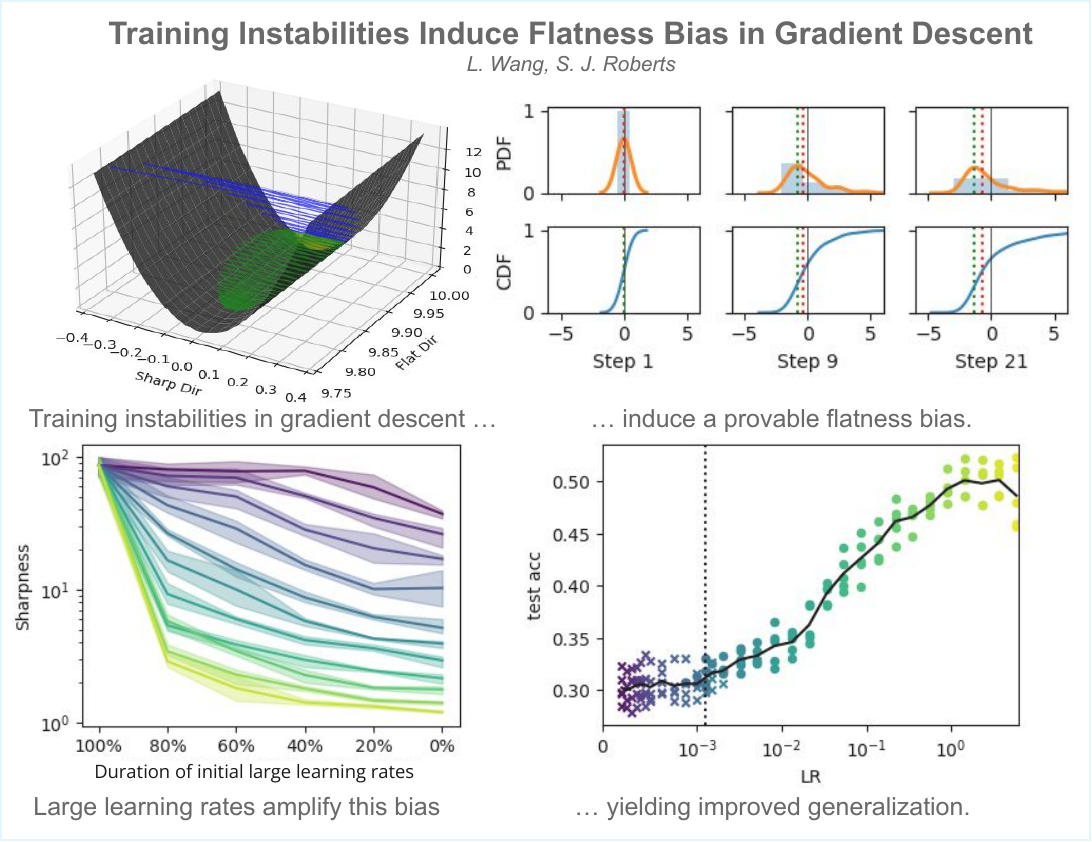}
\end{graphicalabstract}

\begin{highlights}
\item Training instabilities in gradient descent don't harm, but improve generalization.
\item Geometric mechanisms drive exploratory dynamics during unstable phases of training.
\item Exploratory dynamics during instabilities provably induce bias toward flat minima.
\item Flattening persists under stochastic gradient descent and outweighs noise effects.
\item Restoring instabilities improves generalization in adaptive optimization.
\end{highlights}

\begin{keyword}
Training Instabilities \sep Spectral Analysis \sep Flatness Bias \sep Gradient Descent \sep Generalization \sep Deep Learning


\end{keyword}
\end{frontmatter}



\section{\label{ch:intro} Introduction}

Deep neural networks achieve remarkable success across a wide range of tasks, yet their generalization performance remains highly sensitive to optimization hyperparameters, in particular the learning rate. Gradient-based methods such as stochastic gradient descent (SGD) and Adam~\cite{kingma2017adam} dominate modern practice due to their simplicity and scalability. Classical analysis, via the \emph{descent lemma}, bounds the learning rate by the local curvature (or sharpness) to ensure stable optimization and monotonic decreases in loss for convex objectives.

Recent work has challenged this view by identifying an \emph{unstable} regime of training, or training \emph{instabilities}, which operates beyond the descent-lemma bound~\cite{lewkowycz2020largelearningratephase}. Surprisingly, models trained in this regime often maintain near-threshold sharpness while achieving lower loss and improved generalization~\cite{cohen2022eos}. These findings suggest that instabilities, rather than being purely harmful, may play a constructive role in exploring different regions of the loss landscape. More surprisingly, these dynamics empirically culminate in flatter minima with improved generalization.

Our work provides a theoretical account of flattening. Following the background, we explore the formation and resolution of training instabilities in Sec.~\ref{ch:phases}. From these insights, we derive a mechanism called  \emph{Rotational Polarity of Eigenvectors} (RPE), which arises naturally in gradient descent (GD) on deep neural networks (Sec.~\ref{ch:rpe}). RPE captures geometric rotations in leading eigenvectors of the Hessian. Above the stability threshold, RPE promotes exploratory dynamics, which increases with larger learning rates, thus linking neural depth to exploration during unstable training.

Building on this, we show that RPE-effects induce an \emph{implicit flatness bias} in GD. In Sec.~\ref{ch:flatness}, we formalize this effect through the dynamics of higher-order curvature moments, demonstrating that GD during instabilities systematically accumulates bias toward flatter orientations of the loss landscape. The bias strengthens with larger learning rates and longer training, aligning theoretical predictions with observed trends in generalization. 

We validate these insights empirically in Sec.~\ref{sec:flat:cv-exp}, revealing a clear phase transition: generalization benefits emerge only once training enters the unstable regime. Moreover, large learning rates early in training confer lasting advantages, even after subsequent decay. We further extend the flatness framework to the stochastic minibatch setting in Sec.~\ref{ch:sgd}, leveraging random-matrix theory (RMT) to analyze stochastic Hessians, and finally to adaptive optimization in Sec.~\ref{ch:ada}, where we introduce \emph{Clipped-Ada}, a variant of Adam that includes the generalization benefits of instabilities.

To encourage reproducible research, the code for our experiments is made available.

\section{Background}\label{sec:bg}

\paragraph{Gradient descent and the descent lemma}
Let $\mathcal{L}(\vtheta)$ denote the loss function and consider GD with learning rate $\eta$:
\begin{equation}
    \vtheta_{t+1} = \vtheta_t - \eta \nabla \mathcal{L}(\vtheta_t) \,,
\end{equation}
where $\vtheta$ are the parameters. For an $\ell$-smooth objective, the \emph{descent lemma}~\cite{Baillon1977QuelquesPD} ensures:
\[
    \mathcal{L}(\vtheta_{t+1})
    \le \mathcal{L}(\vtheta_t)
    - \eta \!\left(1 - \tfrac{\eta \ell}{2}\right)
    \! \|\nabla \mathcal{L}(\vtheta_t)\|_2^2 \,,
\]
implying monotonic decrease for $0 < \eta < 2/\ell$. The limiting case $\eta = 2/\ell$ marks the stability boundary, motivating the practical $2 / \lambda_{\max}$ heuristic, where $\lambda_{\max}$ is the largest eigenvalue of the Hessian $\mH$. In the local quadratic approximation, GD exhibits \emph{valley jumping} beyond the stability limit: steps larger than $2/\ell$ overshoot the minimum and ascend along the opposite wall of the curvature valley, producing oscillatory or climbing dynamics around the stability boundary. 

In deep networks, however, $\ell$ varies throughout training, and empirical studies~\cite{cohen2022eos} observe that sharpness progressively increases until $\lambda_{\max}\!\approx\! 2/\eta$, termed \emph{progressive sharpening}, after which the system fluctuates near this boundary, the so-called \emph{Edge of Stability} (EoS), while the loss continues to decrease non-monotonically.

\paragraph{Sharpness and generalization}
Over-parametrized neural networks possess many near-zero-loss minima~\cite{zhang2017understanding}.  Flat minima, characterized by low curvature, are associated with better generalization, and commonly approximately measured by the largest Hessian eigenvalue~\cite{Hochreiter1997flat,keskar2017largebatch,jastrzebski2018threefactors}.
Although sharpness is not scale-invariant~\cite{dinh2017sharpminima}, it remains a useful predictor of test performance~\cite{jiang2019generalizationmeasures}.  
This connection motivates analyses that links sharpness to generalization.

\paragraph{Generalization with minibatch noise}
Under mild conditions, the dynamics of SGD with batch size $B$ can be approximated by a stochastic differential equation (SDE).  
The stochastic gradient, $\vg_\mathrm{mb}$ can be decomposed as
\begin{equation}\label{eqn:prelim:sgd-decomp}
    \vg_{\mathrm{mb}}(\vtheta)
    = \nabla \mathcal{L}(\vtheta)
    + \vxi_{\mathrm{mb}},
\end{equation}
where the noise $\vxi_\mathrm{mb}$ satisfies $\mathbb{E}[\vxi_{\mathrm{mb}}] = 0, \mathrm{Cov}(\vxi_{\mathrm{mb}}) \propto 1/B$.  
Taking the continuous-time limit $t = k\eta$ gives
\begin{equation}\label{eq:prelim:sgd-sde}
    \mathrm{d}\vtheta = -\nabla \mathcal{L}(\vtheta)\,\mathrm{d}t
    + \sqrt{2\,T_{\mathrm{eff}}(\vtheta)}\,\mathrm{d}W_t , \quad \text{where the effective temperature scales as} \quad  
    T_{\mathrm{eff}}(\vtheta)
    \propto \eta\,\frac{N}{B}\,\bar{\sigma}^2(\vtheta) \, ,
\end{equation}
with $\bar{\sigma}^2$ denoting the gradient-noise variance and $N$ the dataset size.  
In a quadratic neighborhood, the stationary distribution follows
\[
    \pi(\vtheta)
    \propto
    \exp\!\left[-\tfrac{\mathcal{L}(\vtheta)}{T_{\mathrm{eff}}}\right]
    [\det \mH(\vtheta)]^{-1/2}.
\]
Larger learning rates or smaller batches raise $T_{\mathrm{eff}}$, biasing SGD toward flatter minima.  
This \emph{SDE-view} of flatness offers a theoretical basis for the empirical correlation between instability, diffusion, and generalization~\cite{goyal2017accurate,granziol2020lr,sclocchi2023dissecting,xie2021diffusion}.

\paragraph{Related work and implementation}
With considerable recent contributions toward the theory and empirical study of GD and generalization, a broader discussion of related works appears in App.~\ref{app:related}. Details on numerical estimation of Hessian spectra and the cosine-Grassmannian metric \cite{wang2023instabilitieslosslandscape}, are provided in App.~\ref{sec:phases:tech}. Experimental details in App.~\ref{app:exp-dets}.

\section{\label{ch:phases} Two Empirical Observations on Training Instabilities}

Recent studies have reported that training instabilities coincide with pronounced rotations in the subspace spanned by the leading eigenvectors of the loss Hessian~\cite{wang2023instabilitieslosslandscape}. 
Building on these findings, we present two complementary empirical observations that motivate our theoretical analysis of instability dynamics.

First, we show that \emph{valley jumping}, the deterministic overshoot predicted by the classical descent lemma when $\eta > 2/\ell$, is the primary driver of training instabilities (Subsec.~\ref{sec:phases:emp:drivers}). 
Second, we find that the resolution of instabilities is consistently preceded by a \emph{flattening} phase in higher-order curvature moments (HOCMs) (Subsec.~\ref{sec:phases:emp:resolution-movie}). 
Together, these observations reveal that instability and flattening form two coupled phases of training dynamics: valley jumping amplifies curvature, while subsequent flattening restores stability.

For clarity, all experiments in this section use a standard multilayer perceptron (MLP) trained on Fashion-MNIST (fMNIST) with full-batch GD. Experimental details are provided in Appendix~\ref{app:exp-dets}.

\subsection{Instability phases: progressive sharpening and valley jumping}\label{sec:phases:emp:drivers}

\begin{figure}[t]
\centering
\includegraphics[width=0.9\linewidth]{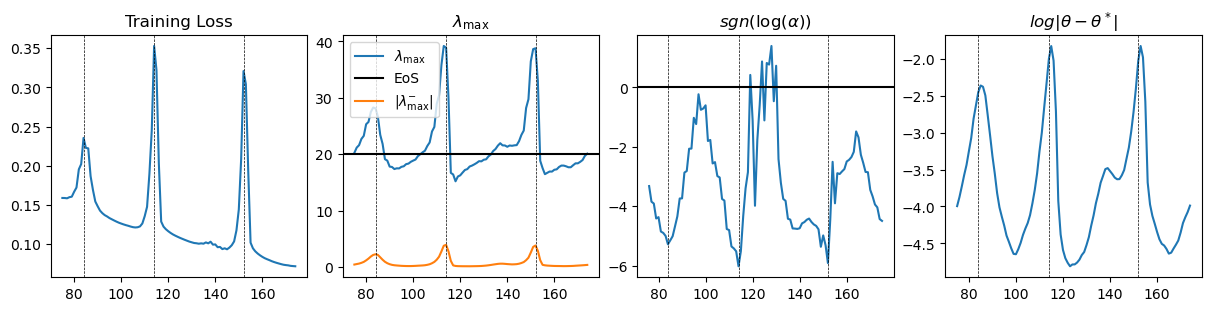}
\caption{\textbf{Stable and unstable phases of training.}
The magnitude of parameter oscillations, measured by $\|\vtheta-\vtheta^*\|_2$, cleanly separates stable and unstable phases in an MLP trained on fMNIST with GD.  
Sharpness increases until $\lambda_{\max}\!\approx\!2/\eta$ (\emph{progressive sharpening}), after which loss and curvature exhibit spikes corresponding to \emph{valley jumping}.}
\label{fig:phases:drivers-phases}
\end{figure}

During early training, sharpness increases gradually until $\lambda_{\max} \approx 2/\eta$, through progressive sharpening, after which training hovers near the EoS~\cite{cohen2022eos}. 
Beyond this point, loss and curvature exhibit short bursts, producing sudden parameter excursions consistent with the \emph{valley-jumping} behavior predicted by the descent lemma: when $\eta > 2/\ell$, steps overshoot the local minimum and climb the opposite wall of the curvature valley, yielding alternating ascent and descent across epochs.  
Fig.~\ref{fig:phases:drivers-phases} illustrates these phenomena along a representative training trajectory.

Classical stability analysis predicts divergence once $\eta \lambda_{\max}>2$, yet modern networks often remain near this boundary for extended periods without diverging.  
Empirically, we observe transient growth in the parameter distance $\|\vtheta-\vtheta^*\|_2$, where $\vtheta^*$ denotes the period-2 average of $\vtheta$, indicating large oscillatory excursions along unstable eigendirections. These excursions are the most direct empirical signature of valley jumping and mark the onset of instability.

To better understand these dynamics, we examine the auxiliary quantity
\[
    \alpha = -\nabla^{\!T}\! \mathcal{L}(\vtheta)\, \nabla \lambda_{\max} \,,
\]
which measures local changes in sharpness~\cite{damian2023selfstab}.  
We find that $\alpha$ is frequently negative during training instabilities, even as $\mathcal{L}(\vtheta)$ and $\lambda_{\max}$ spike, indicating that instability can arise even when sharpness is locally decreasing.  
This suggests that higher-order curvature effects, not captured by $\lambda_{\max}$ alone, are essential for explaining instability.

To capture this non-quadratic structure, we track the most negative eigenvalue of the Hessian, $\lambda_{\text{neg}}$, which reflects local asymmetry and higher-order curvature.  
We observe that $|\lambda_{\text{neg}}|$ systematically increases during instabilities, signaling stronger higher-order interactions, and contracts toward zero as flattening restores stability.  
This growing influence of non-quadratic curvature motivates our later analysis of HOCMs $U_t$ in Sec.~\ref{ch:flatness}.

In App. Fig.~\ref{fig:phases:control-pc}, we further intervene on the directions of gradient updates.  
Suppressing updates along eigenvectors with $\eta \lambda_i>2$ restores stability, whereas restricting updates to these unstable directions induces instability.  

Together, these results identify \emph{valley jumping}, rather than progressive sharpening, as the primary mechanism driving training instabilities.

\subsection{Resolution of instabilities through flattening}\label{sec:phases:emp:resolution-movie}

\begin{figure}[t]
\centering
\includegraphics[width=0.9\textwidth]{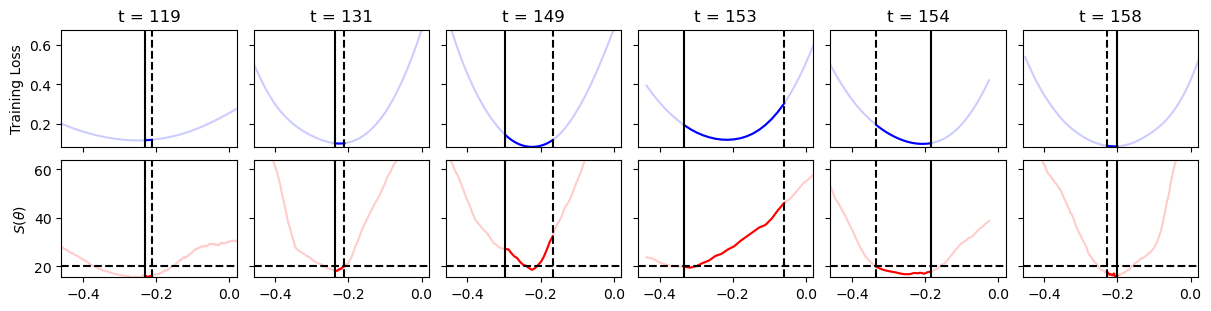}
\caption{\textbf{Loss and sharpness through a training instability.}
We plot the loss (top) and curvature (bottom; leading Hessian eigenvalue estimated via Hutchinson’s trick~\cite{hutchinson1989stochastic}) across representative snapshots. 
Fluctuations in the curve of curvature coincide with strong reorientations of leading eigenvectors, via \emph{rotations}, that precede rapid flattening and the restoration of stability.}
\label{fig:phases:movie}
\end{figure}

To understand how instabilities are resolved, we track the joint evolution of loss $\mathcal{L}(\vtheta)$ and leading curvature $\lambda_\mathrm{max}$ over an instability in Fig.~\ref{fig:phases:movie}. Snapshots are shown here, and the complete epoch-by-epoch is in App. Fig.~\ref{fig:phases:giant-movie}. 

During progressive sharpening, $\lambda_\mathrm{max}$ steadily rises while $\mathcal{L}(\vtheta)$ decreases, consistent with stable training under the descent lemma, reaching EoS at epoch~131.
Beyond the EoS, oscillations amplify and parameters expand along unstable directions. 
Ordinarily such excursions would lead to divergence, yet we observe rapid \emph{flattening} near epoch~149, where $\lambda_\mathrm{max}$ decreases sharply, counteracting parameter growth and restoring stability by epoch~154. 

Where the curve of $\lambda_\mathrm{max}$ remains relatively consistent in instability (epochs $\sim$ 131-148), unique to the restoration of stability is the sudden flattening of the curve of $\lambda_\mathrm{max}$. 
The most compelling mechanism, which we conjecture, is that that rotations of the leading Hessian eigenvectors play a critical role in enabling rapid flattening \cite{wang2023instabilitieslosslandscape}.  
In other words, the dominant subspace is immediately reoriented toward a flatter configuration, effectively `finding' a stable orientation, seemingly by chance. Subsequent epochs restart the cycle: sharpening via PS, instability, and flattening, producing self-regulating oscillations (in $\lambda_\mathrm{max}$) around the stability boundary.  

Together, these observations highlight the coupled evolution of curvature magnitude and orientation.  
Instability amplifies of higher-order curvatures, and we conjecture, promotes rotation of the dominant Hessian eigenvectors, leading to higher-order flattening which precedes and enables the resolution of instabilities.  
This interplay between curvature growth, eigenvector rotation, and higher-order flattening motivates our geometric mechanism of rotation dynamics in Sec.~\ref{ch:rpe} and our formal model of HOCMs, from which a flatness bias is derived, in Sec.~\ref{ch:flatness}.

\section{\label{ch:rpe}Rotational Polarity of Eigenvectors} 

Building on the eigenvector rotations observed in Sec.~\ref{ch:phases} and prior evidence \cite{wang2023instabilitieslosslandscape}, we now develop a theoretical model for the geometric mechanism to understand how these rotations emerge.  
Our analysis shows that under GD, instability can reverse the direction of curvature concentration: the leading eigenvector of the loss Hessian rotates away from the sharpest directions, a phenomenon we term \emph{Rotational Polarity of Eigenvectors} (RPE).  
This effect arises naturally from the chain rule applied to multiplicative parameterizations, where the coupling of layers in deep models induces systematic eigenvector rotations as depth increases.  

We formalize these dynamics in the setting of diagonal linear networks (DLNs) using a concise \emph{Sum-of-Products (SoP)} representation that exposes curvature coupling across layers.  
From this construction, we derive an analytic path from the network Hessian to $\gamma_\beta$, a quantity which describes the rate of change in $\beta$, where $\beta$ determines the polarity of rotation. 
We present theoretical results in Subsec.~\ref{sec:rpe:rot} and validate them empirically in Subsec.~\ref{sec:rpe:exp}.

\subsection{Rotations as a Consequence of Depth} \label{sec:rpe:rot}

We begin by introducing the SoP formulation for MLPs, which we model as a sum of independently parameterized DLNs, as studied in \cite{pesme2021diagonal}. Each DLN captures multiplicative structure of weights resulting from  depth. We define an $mn$-DLN in Subsubsec.~\ref{sec:rpe:rot:sop-mn}, where we present our SoP formulation in depth. We then analyze the case of $m=1$, $n=2$, and $\sigma = \mI$ in Subsubsec.~\ref{sec:rpe:rot:dln2}, where computations are exact and tractable. In this setting, we reveal salient geometrical insights, which we jointly call RPE, that govern the character rotations of the leading Hessian eigenvectors during training. We show that the strength of these rotations increase with learning rates. 

In the Appendix, we extend our derivation to more general settings: general depth ($n \in \mathbb{N}$); Rectified Linear Unit (ReLU) activations; general width ($m \in \mathbb{N}$); and DLNs with shared parameters. An outline of extensions is found in App.~\ref{sec:rpe:rot:dln-ext} and the detailed derivations in Appendices~\ref{app:rpe:dln-n}-\ref{app:rpe:dln-shared}. 

\subsubsection{The Sum-of-Products decomposition of a multilayer perceptron} \label{sec:rpe:rot:sop-mn}

Consider a standard MLP with input $ \vx \in \mathbb{R}^N $, depth $D$, and layer-wise parameters:
\[
    f_\mathrm{MLP}(\vx) = \mW_D \sigma_{D-1} \big( \cdots \sigma_1(\mW_1 \vx + \vb_1) \cdots \big) + \vb_D \, , 
\]

\noindent
where each hidden layer $ d \in \{1, \dots, D\} $ has weights $ \mW_d \in \mathbb{R}^{h_d \times h_{d-1}} $, biases $ \vb_d \in \mathbb{R}^{h_d} $, activation functions $\sigma_d$. $h_d$ denotes the size of activations to each $\sigma_d$, where $h_0 = N$. 

To study curvature dynamics in a tractable setting, we now introduce a simplified formulation of MLPs under mild conditions. In the linearization limit of $\sigma_d$s, we may adopt a path-wise view of MLPs: the network can be expressed as a sum over computational paths from input to output. Each path contributes a term involving a product of weights along that path, modulated by an input-dependent coefficient.

We model each path as a \emph{DLN} - a minimal network that captures multiplicative interactions across layers \cite{pesme2021diagonal}. A DLN with scalar parameters $ \vartheta_1, \dots, \vartheta_n $ reflects a computation path across depth $D = n$:
\[
    f_{\mathrm{DLN}}(\vartheta_1, \dots, \vartheta_n) = \vtheta = \prod_{i=1}^n \vartheta_i \, ,
\]
where how each $\vartheta_i$ corresponds to a scalar weight at layer $i$ \footnote{In general, we use $\vartheta$ for (scalar) parameters of a DLN and $\vtheta$ for the (vector) parameters of a deep neural network. }. This multiplicative form reflects the vertical structure of depth: parameters across layers interact through products rather than additive composition.

As a result, MLPs can be written using the SoP formulation, which is an $mn$-DLN, which is a sum of $m$ independently parametrized DLNs, each of depth at most $n$:
\[
    f_{\mathrm{MLP}}(\vx) = \sum_{j=1}^m f_{\mathrm{DLN}}(\vtheta_j) = \sum_{j=1}^m \prod_{i=1}^{n_j} \vartheta_{i,j} \, .
\]
Here, each DLN $j$ has its own parameter set $ \{\vartheta_{i,j}\}_{i=1}^{n_j} $, where $n_j \leq n$ is the depth of DLN $j$. For simplicity, we model each DLN with distinct parameters, to isolate the effects of multiplicative depth, and we consider the more realistic setting of shared-parameter in App.~\ref{app:rpe:dln-shared}. We also omit the dependence on scalar input $ x \in \mathbb{R} $ \emph{w.l.o.g.}, absorbing it into a constant coefficient for each DLN term. 

While our theoretical setup considers scalar-valued DLN paths for analytical simplicity, the analysis naturally extends to networks with vector-valued inputs and scalar loss functions, including MSE and CE. Our curvature derivations apply to each input independently, and the resulting Hessian structure, and associated RPE behavior, are expected to hold across minibatches or aggregate losses. 

We note that the SoP formulation of MLPs, while veritable, remains a \emph{theoretical simplification}, and not a general claim about all MLPs. For example, in empirical MLPs, parameters are shared across paths and activations are nonlinear. Our abstraction allows us to study depth-induced multiplicativity in isolation, disentangled from these effects. While our theory does not model all aspects of empirical MLPs, we observe the full characterized behavior in practice (see Subsec.~\ref{sec:rpe:exp}). For a full derivation and discussion on the SoP formulation of MLPs, see App.~\ref{app:rpe:sop}. 

\subsubsection{Eigenvector rotations of a DLN model:} \label{sec:rpe:rot:dln2}

We now study the dynamics of GD in a single DLN with $m=1$, $n=2$, and $\sigma=\mI$, which computes the product
\[
    \vtheta = \vartheta_1 \vartheta_2 \, ,
\]
reflecting the multiplicative structure of each computational path. We write the loss using the function $z$ as:
\[
    \mathcal{L}(\vartheta_1, \vartheta_2) = z(\vtheta) \, ,
\]
where $ z: \mathbb{R} \to \mathbb{R}_{\geq 0} $ is a convex, twice-differentiable function with a unique minimum at some scalar $ \vtheta^* \in \mathbb{R} $.

We restrict our attention to local neighborhoods where the second-order Taylor approximation of the loss provides an accurate local model. This allows analysis of curvature dynamics via the Hessian without requiring global quadratic structure. Importantly, we do not assume $ \vtheta^* = 0 $; our analysis and results hold for arbitrary $\vtheta^*$, and all expressions (e.g., the derivatives $z'$, $z''$) are evaluated at the current $\vtheta$. While our toy example sets $ \vtheta^* = 0 $ for simplification, this assumption is not used in any derivation and is unnecessary for correctness, as our derivations rely on the gradients of the Hessian structure of GD updates - all of which are centered about the local minima. The parameter-space dynamics of GD depend on the multiplicative mapping $ \vtheta = \vartheta_1 \vartheta_2 $, and this structure alone drives the apparent rotational dynamics.

We proceed to compute the Hessian of the loss with respect to the DLN parameters. Let $z' := \left. \frac{d z}{d \vtheta} \right|_{\vtheta} $ and $z'' := \left. \frac{d^2 z}{d \vtheta^2}\right|_{\vtheta}$. Using the chain rule, we obtain the full $2 \times 2$ Hessian in our $n=2$ case:
\[
\mH =
\begin{bmatrix}
\frac{\partial^2 L}{\partial \vartheta_1^2} & \frac{\partial^2 L}{\partial \vartheta_1 \partial \vartheta_2} \\
\frac{\partial^2 L}{\partial \vartheta_2 \partial \vartheta_1} & \frac{\partial^2 L}{\partial \vartheta_2^2}
\end{bmatrix}
=
\begin{bmatrix}
z'' \vartheta_2^2 & z'' \vartheta_1 \vartheta_2 + z' \\
z'' \vartheta_1 \vartheta_2 + z' & z'' \vartheta_1^2
\end{bmatrix} \, .
\]

\noindent
The eigenvalues and eigenvectors of $\mH$ describe the principal curvature directions of the loss landscape. Let $(\lambda_\mathrm{max}, \vv_1)$ and $(\lambda_2, \vv_2)$ be the eigenvalue-eigenvector pairs of $\mH$, where $\lambda_\mathrm{max} \geq \lambda_2$. Identify $\mathbb{R}^2$, ($n=2$), with the coordinate axes aligned with $\vartheta_1$ and $\vartheta_2$. Each Hessian eigenvector $\vv_i \in \mathbb{R}^2$ is written in this basis as:
\[
    \vv_i = \begin{bmatrix} v_{i,1} \\ v_{i,2} \end{bmatrix} \, .
\]
where $v_{i,1}$ (resp.\ $v_{i,2}$) is the component of $\vv_i$ along the $\vartheta_1$ (resp.\ $\vartheta_2$) axis. We also consider that $\vartheta_2^2 > \vartheta_1^2$ w.l.o.g.. In this case, $\mathcal{L}(\vtheta)$ is more sensitive to changes in $\vartheta_1$ than $\vartheta_2$, or equivalently that $\vartheta_1$ is the smaller but \emph{sharper} parameter. To quantify the orientation of the leading eigenvector $\vv_1$, we define and use the ratio:
\[
    R_{n=2} := \left| \frac{v_{1,1}}{v_{1,2}} \right| \, ,
\]
which captures the relative alignment of $\vv_1$ to the $\vartheta_1$ (sharper) axis. We use the subscript ${n=2}$ to denote the particular form in this case and to differentiate from alternative derivations in the App.~\ref{app:rpe}. Larger values of $R_{n=2}$ indicate stronger alignment with $\vartheta_1$, the sharper parameter. Note that $R_{n=2}$ is invariant to the sign of $\vv_1$ and captures the eigenvector orientation through the ratio of coordinates. 

As a short summary, the Hessian $\mH$ reflects both local curvature (via $z''$) and gradient alignment (via $z'$). The off-diagonal term includes $z'$, which couples the current loss value with curvature asymmetry between parameters. As GD evolves, these quantities change, inducing rotations in $\vv_1$. By tracking $R_{n=2}$, we can measure how much the leading curvature direction aligns with one parameter over the other.

Following this, we derive a closed-form expression for $R_{n=2}$ in terms of the parameters and loss derivatives. Solving the eigenvalue equation for $\mH$ yields:
\[
    R_{n=2} = \left| \frac{v_{1,1}}{v_{1,2}} \right| = \beta + \sqrt{\beta^2 + 1} \, ,
\]
where we define the scalar quantity:
\[
    \beta := \left| \frac{z'' (\vartheta_2^2 - \vartheta_1^2)}{2(z' + z'' \vtheta)} \right| \, .
\]
This expression is exact and valid whenever $z' + z'' \vtheta \neq 0$. The derivation for $R_{n=2}$ and its sign invariance is provided in App.~\ref{app:rpe:dln2:R}.

The quantity $\beta$ reflects the degree of anisotropy in curvature across parameters. The numerator, $z''(\vartheta_2^2 - \vartheta_1^2)$, measures the difference in sharpness between directions, while the denominator, $2(z' + z'' \vtheta)$, normalizes this difference relative to the local gradient. Large values of $\beta$ imply that curvature is strongly concentrated along one parameter (typically the sharper one), which pulls the leading eigenvector of the Hessian toward that axis. Conversely, when curvature is more balanced, $\beta$ is small and the eigenvector becomes more isotropic. Since $R$ increases monotonically with $\beta$, changes in $\beta$ directly govern the rotation of the leading eigenvector. 

We proceed to analyze how $\beta$ evolves during gradient descent. Since $R_{n=2}$ is a monotonic function of $\beta$, the rate of change in $\beta$ directly reflects how the leading eigenvector of the Hessian rotates. We define the relative change in variables $\gamma_x := \frac{x_{t+1} - x_t}{x_t}$ for arbitrary $x$, and thus write $\gamma_\beta$ as:
\[
    \gamma_\beta := \frac{\beta + \Delta \beta}{\beta} \, ,
\]
where $\Delta \beta$ is the change in $\beta$ after one GD update. Using the GD dynamics under a fixed learning rate $\eta$, we derive (see App.~\ref{app:rpe:dln2:gamma} for details):
\[
\gamma_\beta \approx \frac{\gamma_{|\vartheta_2^2 - \vartheta_1^2|}}{\gamma_\vtheta}
= \left| \frac{1 - \eta^2 z'^2}{1 - \zeta \eta z' + \eta^2 z'^2} \right| \, ,
\]
where \emph{temporarily}, we define $\zeta := \left( \frac{\vartheta_1}{\vartheta_2} + \frac{\vartheta_2}{\vartheta_1} \right) > 2$ for notational clarity (the definition of $\zeta$ will not be used beyond this point). This expression for $\gamma_\beta$ is not exact, but it can predict the direction of eigenvector rotation across learning rates.

The expression for $\gamma_\beta$ reveals how the rotation of the leading eigenvector depends on the learning rate $\eta$ and effective curvature scale $z'$. When $\gamma_\beta > 1$, $\beta$ increases and the eigenvector aligns more strongly with the sharper parameter (e.g., $\vartheta_1$). When $\gamma_\beta < 1$, $\beta$ decreases and the eigenvector rotates away from sharp directions. The function $\gamma_\beta$ has a vertical asymptote when $\gamma_\vtheta \to 0$, i.e., when the change in $\vtheta$ vanishes. This update is the \emph{Newton step}, common to second-order optimizers, where we reach the minimum in one step for quadratic settings.

As a simple illustration, consider a quadratic loss $z(\vtheta)=\tfrac{1}{2}\vtheta^2$.  
Here, the Newton-step asymptote occurs at $\eta=1/\lambda$, while the rotation threshold $\gamma_\beta=1$-signaling reversal of eigenvector orientation-appears slightly before the Edge of Stability at $\eta=\eta_{\mathrm{eos}}(1-\xi)$, where $\xi$ depends on the relative scales of $\vartheta_1$ and $\vartheta_2$.  
This example is provided purely for intuition; the subsequent analysis does not assume a quadratic form or rely on this specific parameterization.

We highlight two common choices for the local loss $z$, summarized below:
\begin{align*}
    \text{(MSE)} \quad 
    & z(\vtheta)=\tfrac{1}{2}(\vtheta-y)^2, 
      && z'=\vtheta-y, 
      && z''=1, \\
    \text{(Binary CE)} \quad 
    & z(\vtheta)=\log\!\bigl(1+e^{-y\vtheta}\bigr), 
      && z'=-\tfrac{y}{1+e^{y\vtheta}}, 
      && z''=\sigma_\mathrm{ce}(y\vtheta)\bigl(1-\sigma_\mathrm{ce}(y\vtheta)\bigr),
\end{align*}
where $\sigma_\mathrm{ce}$ denotes the logistic sigmoid.  
These derivatives substitute directly into expressions for $\mH$, $\beta$, and $R_{n=2}$ without affecting subsequent analysis.  
For multiclass cross-entropy, $z''$ represents the variance of class probabilities along the chosen logit direction.  
Unlike MSE, CE is asymmetric about its minima, so its curvature is more accurately captured through a two-step update analysis (see App.~\ref{app:rpe:dln-relu}).

\begin{figure}[t]
\centering
\includegraphics[width=0.7\textwidth]{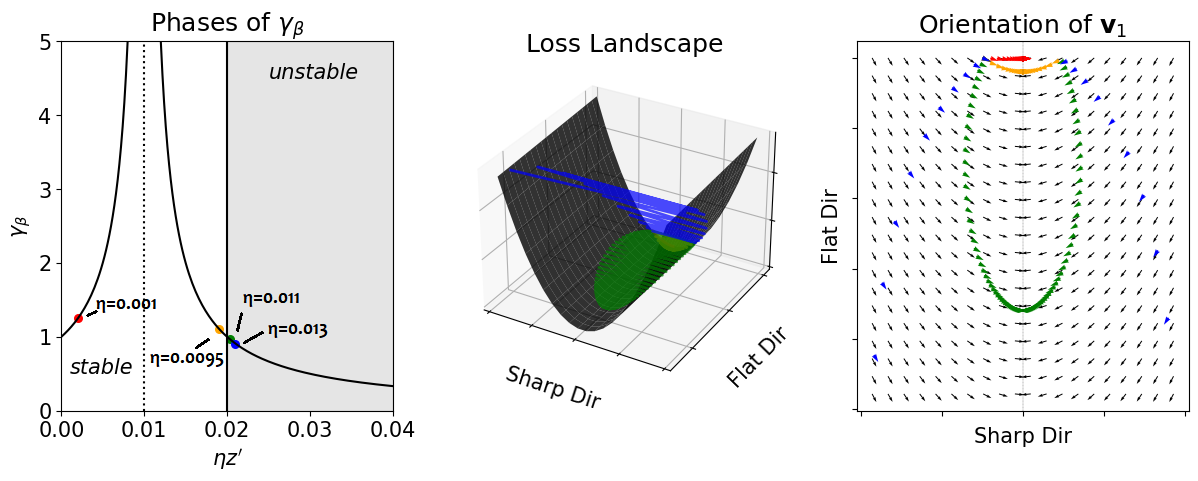}
\caption[Optimization trajectories in a $2$-parameter DLN with varying $\eta$]{\textbf{Optimization trajectories in a $2$-parameter DLN with varying $\eta$}. Left: the regimes of $\gamma_\beta$ vs learning rate. Middle: the loss landscape. Right: rotation of the leading Hessian eigenvector $\vv_1$. Stable regimes (orange/red/green) show convergence to sharper axis; unstable regimes (blue) show divergence.}
\label{fig:rpe:cartoon}
\end{figure}

From the above, $\gamma_\beta$ partitions two distinct regimes where eigenvector alignment is either reinforced or reversed:

\begin{itemize}
    \item \emph{Stable regime:} When $\gamma_\beta > 1$, the value of $\beta$ increases, and the leading eigenvector $\vv_1$ aligns increasingly with $\vartheta_1$ - the sharper parameter. This regime corresponds to learning rates $\eta$ such that:
    \[
        \eta z' < \frac{2 \vartheta_1 \vartheta_2}{\vartheta_1^2 + \vartheta_2^2} =: \eta_{\mathrm{eos}}(1 - \xi) \, ,
    \]
    where we define $\eta_{\mathrm{eos}} := \frac{2 \vartheta_1}{\vartheta_2 z'}$ and $\xi := \frac{\vartheta_1^2}{\vartheta_1^2 + \vartheta_2^2}$.

    \item \emph{Unstable regime (instabilities):} When $\gamma_\beta < 1$, $\beta$ decreases and the leading eigenvector rotates \emph{away} from the sharper axis. This regime begins when:
    \[
        \eta > \eta_{\mathrm{eos}}(1 - \xi) \, .
    \]
\end{itemize}

\noindent For the instability regime to exist, we require $\eta_{\mathrm{eos}} < 1/z'$, a condition that which simplifies to:
\[
    \frac{2 \vartheta_1}{\vartheta_2 z'} < \frac{1}{z'} \quad \Rightarrow \quad \vartheta_2^2 > 2 \vartheta_1^2 \, .
\]
This mild constraint is satisfied whenever the parameters are sufficiently imbalanced - a condition that often holds in practice, where empirical neural networks are ill-conditioned \cite{granziol2020lr, papyan2019spectrum}.

As GD progresses, the behavior of the leading eigenvector depends critically on the learning rate. Below the EoS, $\vv_1$ increasingly tracks the sharper parameter. Above it, this direction destabilizes, and $\vv_1$ rotates away. These dynamics underpin RPE, and are visualized empirically in Fig.~\ref{fig:rpe:cartoon}. Therefore, RPE consists of these following behaviors:
    \begin{enumerate}
      \item \emph{Opposite rotation:} The axis of rotation for the leading eigenvectors of the Hessian during training are opposite for stable and unstable training.
      \item \emph{Monotonicity:} In regimes above and below the stability threshold, more extreme (larger when above or smaller when below) learning rates lead to increased levels of rotation.
    \end{enumerate}
\noindent
Extension to more general settings is outlined in App.~\ref{sec:rpe:rot:dln-ext}, and detailed derivations are found in Apps.~\ref{app:rpe:dln-n}-\ref{app:rpe:dln-shared}. 


\paragraph{Toy example}
Fig.~\ref{fig:rpe:cartoon} visualizes GD trajectories on a toy example: a two-parameter DLN with initialization $(\vartheta_1,\vartheta_2)=(-0.1,10)$ and quadratic loss $z(\vtheta)=\tfrac{1}{2}\vtheta^2$.  
We vary the learning rate $\eta$ and examine the resulting curvature dynamics.  
The left panel shows $\gamma_\beta$ as a function of $\eta$, highlighting the transition at the edge of stability $\eta_{\mathrm{eos}}$; the middle panel depicts anisotropic curvature arising from the multiplicative structure; and the right panel tracks the orientation (and rotation) of the leading Hessian eigenvector $\vv_1$ across iterations. 

In the \emph{stable regime} ($\eta < \eta_{\mathrm{eos}}$), $\vv_1$ gradually aligns with the sharper parameter axis, consistent with increasing curvature concentration.  
In the \emph{unstable regime} ($\eta > \eta_{\mathrm{eos}}$), the rotation reverses direction. $\vv_1$ moves away from $\vartheta_1$ even though $\vartheta_1$ remains smaller in magnitude, confirming the predicted \emph{opposite rotation} behavior of RPE.  
These observations demonstrate that curvature alignment in DLNs evolves in opposite directions across stability regimes, providing a geometric interpretation of instability resolution.  
This depth-induced eigenvector rotation forms the theoretical foundation for the flatness bias analyzed in Sec.~\ref{ch:flatness}.  

\subsection{Rotational Polarity of Eigenvectors in Empirical Multilayer Perceptrons}\label{sec:rpe:exp}

\begin{figure}[t]
\centering
\includegraphics[width=\textwidth]{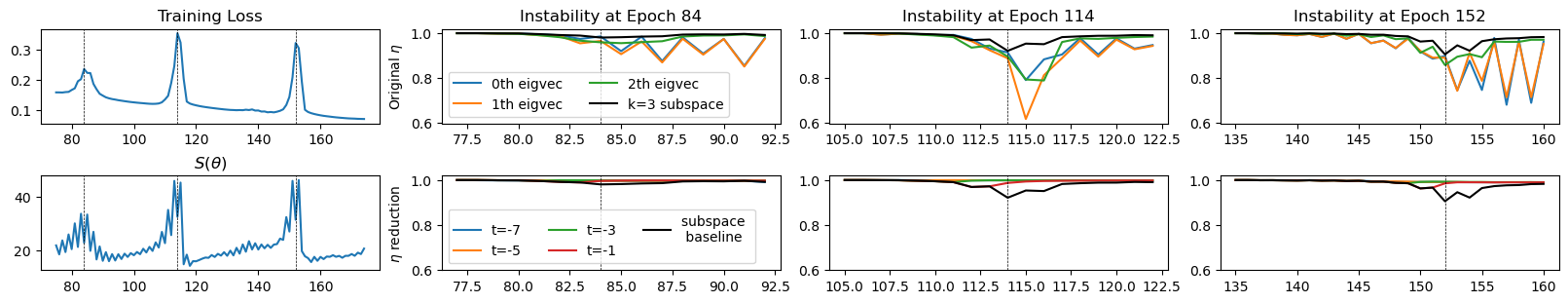}
\caption[\textbf{RPE, Opposite Rotation in MLPs.}]{\textbf{RPE, Opposite Rotation in MLPs.}  
During instabilities, the sharpest Hessian eigenvectors rotate away (Top), while stability reverts these rotations (Bottom).  
We track the similarity of the top-$k$ eigenvectors and the subspace they span in an MLP on fMNIST across three instability cycles.  
\textbf{Left:} $\mathcal{L}(\vtheta)$ and $\lambda_{\max}$ ($S(\vtheta)$).  
\textbf{Top:} Similarities of the leading eigenvectors (colored) and the top-3 subspace (black).  
\textbf{Bottom:} Subspace baseline (black) with colored trajectories for each $\eta$ reduction. }
\label{fig:rpe:instab_rot}
\end{figure}

We next examine whether RPE for MLPs trained on fMNIST, extending insights from the theoretical DLN model.  
Fig.~\ref{fig:rpe:instab_rot} shows the evolution of the leading Hessian eigenvectors across three instability episodes.  
Each curve measures the similarity of leading eigenvectors relative to a baseline defined at the start of each episode. 

In the upper panel, we observe gradual, monotonic rotations as $\mathcal{L}(\vtheta)$ approaches its peak values, mirroring the behavior predicted by the DLN theory.  
These rotations are smooth and distributed across multiple eigenvectors, reflected in a global decrease in similarity rather than abrupt swaps.  
After each instability resolves, individual eigenvectors remain dissimilar even as the subspace similarity stays high, implying that reorientations within the principal subspace, which are new linear combinations of sharp directions, may contribute to recovery.  

In the lower panel, we intervene on the learning rate before instability resolution, setting $\eta_\mathrm{low}=0.2\eta$.  
This reduction immediately reverses the previously observed rotations, demonstrating that stability and instability phases induce \emph{opposite rotations} of the Hessian eigenvectors.

\begin{figure}[t]
\centering
\includegraphics[width=\textwidth]{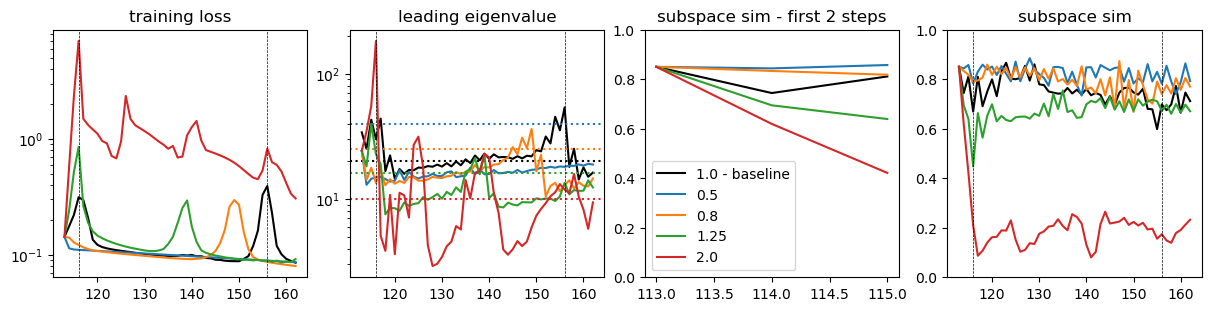}
\caption[\textbf{RPE, Monotonicity in MLPs.}]{\textbf{RPE, Monotonicity in MLPs.}  
We intervene on the learning rate $\eta_\mathrm{int}\!\in\![0.5\eta,\,0.8\eta,\,\eta,\,1.25\eta,\,2.0\eta]$ three epochs before instability resolution.  
\textbf{First:} Training loss; \textbf{Second:} $\lambda_{\max}$ (dotted lines mark new stability thresholds);  
\textbf{Third:} $\mathrm{rot}_{\mathrm{fp}}$ (first two steps); \textbf{Fourth:} $\mathrm{rot}_{\mathrm{fp}}$ (all steps).  
Larger learning rates yield stronger instabilities and greater rotational motion, confirming the monotonicity property of RPE.}
\label{fig:rpe:mono_rot}
\end{figure}

In Fig.~\ref{fig:rpe:mono_rot}, we intervene $3$ epochs before instability resolution, scaling the learning rates as $[0.5\eta,\,0.8\eta,\,\eta,\,1.25\eta,\,2.0\eta]$, and track training loss, $\lambda_{\max}$, and rotational similarity $\mathrm{rot}_{\mathrm{fp}}$.  
Larger learning rates lead to stronger instabilities, spikes in both loss and sharpness, and smaller learning rates suppress these oscillations and restore progressive sharpening.  
Additionally, within instabilities, the scale of the learning rate also governs the degree of rotational motion: higher $\eta$ values produce greater eigenvector rotations between epochs, confirming the \emph{monotonicity} property of RPE.  

This relationship persists across extended training and multiple instability cycles, indicating that the learning rate directly controls the extent of rotational exploration in the loss landscape.  
Although this exploratory bias is not formally part of the RPE theory, we interpret it as an empirical consequence of the two theoretical behaviors described above: depth-induced eigenvector rotations that, together, facilitate transitions toward flatter regions of the loss landscape.

Together, these results demonstrate that RPE captures two key empirical regularities: \emph{opposite rotation} and \emph{monotonicity}, linking curvature dynamics to the stability of training.  
This geometric mechanism underlies the implicit flatness bias analyzed next in Sec.~\ref{ch:flatness}.

\section{\label{ch:flatness}The Implicit Flatness Bias of Gradient Descent} 


Sec.~\ref{ch:rpe} introduced RPE, which characterizes how the leading Hessian eigenvectors rotate during unstable training. Evidence in Sec.~\ref{ch:phases} suggest that training instabilities may be \emph{resolved} through a flattening of HOCMs. Building on these insights, we derive a dynamical system for GD under instability, using $U_t$, a term for HOCMs, to capture curvature variability beyond the quadratic approximation. Together, these results uncover an implicit \emph{flatness bias} in GD, which predicts that improved generalization emerges only within the unstable training regime.

We briefly introduce our theoretical setting in Subsec~\ref{subsec:flat:intro}, and defer a detailed discussion to App.~\ref{sec:flat:dynamics}, where additional empirical evidence supporting our characterization of HOCMs may be found (App.~\ref{app:flat:fmnist-ut}). 

In Subsec~\ref{sec:flat:dyn-system}, we present our major theoretical results that indicate a flatness bias. Thm.~\ref{thm:flat:point-contraction} formalizes contraction (a leftward bias) in the cumulative distribution function (CDF) for atoms, owing to state-dependent noise. We introduce further regularity conditions to derive median contraction for initially unimodal distributions in Thm~\ref{thm:flat:median-drift-unimodal}. For expositional clarity, we defer detailed proofs of these results to the Appendix.  Outlines of the proofs are provided in App.~\ref{app:flat:proof-outlines}, and full proofs detailed in Appendices~\ref{app:flat:pm-contract-proof}-\ref{app:flat:median-drift-unimodal-proof}. Finally, App.~\ref{sec:flat:xt-sim} includes a numerical simulation of $X_t \coloneqq \log U_t$, illustrating the key properties of the flatness bias in GD. 

\subsection{Conceptual motivation and problem setting}
\label{subsec:flat:intro}

We use insights from RPE to motivate theoretical settings for an analysis of GD. Empirically, instabilities trigger sharp yet transient reductions in curvature magnitude (Sec.~\ref{ch:phases}), showing the existence of a self-correcting process that drives models toward flatter regions after periods of instability.
To study this mechanism, we express GD as a coupled dynamical system that tracks two scalar quantities:

\begin{itemize}
    \item The \emph{instability surplus} $\delta_t$, which measures how far the current step size exceeds the local stability threshold; and
    \item An HOCMs term, $U_t$, which quantifies deviations from quadratic curvature along the sharpest Hessian direction.
\end{itemize}

The interaction between these variables forms a minimal model of curvature dynamics under instability.  
$\delta_t>0$ corresponds to the \emph{unstable regime}, where the effective step size temporarily overshoots the stability limit and induces exploratory rotations of the dominant Hessian eigenvectors (RPE).  
These rotations, in turn, alter the local curvature profile, leading to the characteristic flattening observed in empirical loss landscapes.

Motivated by the descent lemma and classical stability theory~\cite{nesterov2004convex,cohen2022eos}, we define:
\begin{definition}[Instability surplus]\label{def:flat:delta}
The instability surplus is
\[
\delta_t := \eta \lambda_{1,t} - C_{\mathrm{eos},t},
\]
where $\eta$ is the learning rate, $\lambda_{1,t}$ the leading Hessian eigenvalue, and $C_{\mathrm{eos},t}>0$ the local stability limit (equal to $2$ for a quadratic loss).  
\end{definition}
We focus on the unstable regime of training $\delta_t>0$, where eigenvector rotations promote exploration. 
As predicted by RPE, when $\delta_t \le 0$, training is stable and the exploratory mechanisms vanish, allowing us to treat stability as an absorbing boundary for instabilities (see Asp.~\ref{asp:exit} below).

To capture deviations beyond the quadratic approximation, we introduce a positive scalar $U_t>0$ that captures HOCMs about the sharpest directions $v_{1,t}$. 
Larger $U_t$ indicates stronger departures from quadratic behavior. 
Our analysis treats $U_t$ abstractly; it does \emph{not} depend on a specific construction, only on positivity and its multiplicative update structure introduced below. 
One convenient instantiation is a directional fourth-derivative envelope:
\[
U_t := 
\E_{v\in\mathcal C_t}\sup_{0<\rho\le\rho_0}
\frac{\bigl|\mathcal L(\vtheta_t+\rho v)+\mathcal L(\vtheta_t-\rho v)
-2\mathcal L(\vtheta_t)-\tfrac12\lambda_{1,t}\rho^2\bigr|}{\rho^4},
\]
with a cone $\mathcal C_t$ around $v_{1,t}$ and local radius $\rho_0$. We stress that our derivations do not depend on the exact form of $U_t$. 

RPE theory and empirical observations (App.~\ref{app:flat:fmnist-ut}) suggest that higher-order curvature measures exhibit strong fluctuations during instability but both RPE theory nor empirical evidence supports consistent drift. This agnosticism toward drift motivates a balanced stochastic process whose variability scales with the level of instability.  
We therefore model $U_t$ as a mean-zero multiplicative process: random fluctuations with magnitudes governed by the surplus $\delta_t$. 
This leads to the stochastic term $\xi_t f_{\log U}(\delta_t)$ introduced below, where $\xi_t\!\in\!\{-1,1\}$ are i.i.d.\ Rademacher signs capturing signed, mean-zero variability, and $f_{\log U}(\delta_t)>0$ encodes how instability strength amplifies curvature fluctuations. 
Working with $\log U_t$ ensures positivity and reveals the multiplicative nature of these curvature updates.

\subsection{Coupled curvature-instability dynamics}
\label{subsec:flat:coupling}
Having defined $\delta_t$ and $U_t$, we now model their mutual dependence through a stochastic coupling that captures the essential curvature-instability feedback observed in practice. The monotonicity of RPE (Sec.~\ref{ch:rpe}) shows that eigenvector rotations grow strictly with the instability surplus~$\delta_t$ and vanish when training reenters stability.  
To capture this behavior, we construct a minimal stochastic feedback model linking $\delta_t$ and $U_t$.  
Rather than model the full dependence of $\lambda_{1,t}$ on past states, we assume local coupling governed by two deterministic response functions:
\[
f_{\log U}:\R_{>0}\to\R_{>0}, \qquad
f_\lambda:\R_{>0}\to\R_{>0},
\]
representing, respectively, how $\lambda_\mathrm{max}$ modulates HOCMs' variability and how HOCMs influence $\lambda_\mathrm{max}$. Specifically: 
\begin{enumerate}
    \item \emph{Instability $\rightarrow$ HOCMs' variability.}  
    Larger surpluses $\delta_t$ increase the magnitude of fluctuations in $\log U_t$, encoded by a strictly increasing $f_{\log U}(\delta_t)$.
    This captures how stronger instability amplifies HOCMs dynamics.
    \item \emph{HOCMs $\rightarrow$ sharpness.}  
    Higher $U_t$ values raise the expected sharpness $\E[\lambda_{1,t}\mid U_t]$, modeled via a nondecreasing $f_\lambda(U_t)$.
    This coupling closes the loop, as greater curvature feeds back into $\delta_t = \eta \lambda_{1,t} - C_t$.
\end{enumerate}
Working with $\log U_t$ ensures positivity and exposes the multiplicative character of curvature updates.

Combining both directions yields the coupled dynamics:
\begin{equation}\label{eqn:flat:Ut-dynamics-short}
\log U_{t+1} = \log U_t + \xi_t f_{\log U}(\delta_t),
\qquad
\delta_t = \eta f_\lambda(U_t) - C_t + \eta \zeta_t,
\end{equation}
where $\xi_t\!\in\!\{-1,1\}$ are i.i.d.\ Rademacher signs representing mean-zero curvature fluctuations and $\zeta_t$ is a zero-mean noise term in the curvature estimate.  
Larger surpluses $\delta_t$ thus produce proportionally larger multiplicative changes in $U_t$, while smaller $U_t$ feed back to reduce $\lambda_{1,t}$ and hence $\delta_t$, forming the essential feedback loop behind instability dynamics.

When $\delta_t>0$, GD operates beyond the EoS, producing curvature rotations (RPE) and transient excursions in sharpness.  
These oscillations inject variability into $U_t$, which subsequently decreases with high probability, reducing expected $\lambda_{1,t}$ and restoring stability.  
This self-regulating cycle: instability $\to$ rotation $\to$ flattening $\to$ stability, constitutes the dynamical origin of the \emph{implicit flatness bias} of instability GD that we formalize in the next subsections. 

\paragraph{Initialization and exit}

We initialize in instability, $\delta_0>0$, so that the feedback loop is active. Once $\delta_t\le0$, the loop is empirically quiescent (as rotations are negligible). The process is therefore absorbed in at the boundary:

\begin{assumption}[Exit by construction]\label{asp:exit}
If $\delta_t\le0$, then $U_{t'}\equiv U_t$ for all $t'\ge t$.
\end{assumption}

\noindent
This assumption matches RPE/EoS observations and avoids specifying a separate stable-regime dynamic.    
Together, $\delta_t$ and $U_t$ form a minimal yet empirically faithful description of curvature feedback of GD during instabilities.

\subsection{Dynamical System for $X_t$}\label{sec:flat:dyn-system}
We now study the dynamical system in Eqn.~\ref{eqn:flat:Ut-dynamics-short}. The previous subsection outlined the following \emph{standing conditions}:

\begin{condition}[Standing Conditions]\label{cond:flat:standing}
 \textcolor{white}{.}

\begin{itemize}
\item \textbf{(C1)} \emph{Boundary learning rate:} 
There exists $\eta_{\mathrm{eos},t}>0$ such that one GD step at $\eta_{\mathrm{eos},t}$ leaves $(\lambda_{1,t},v_{1,t})$ unchanged to first order.  
This follows from the coexistence of stable and divergent regimes (App.~\ref{app:rpe:dln-relu}).  

\item \textbf{(C2)} \emph{Positivity and log-space updates:} 
$U_t>0$ and 
\[
\log U_{t+1}=\log U_t+\xi_t\,f_{\log U}(\delta_t),
\]
where $\xi_t$ are i.i.d.\ Rademacher signs (empirically supported, App.~\ref{app:flat:fmnist-ut}) and $f_{\log U}:\R_{>0}\!\to\!\R_{>0}$ deterministic.  

\item \textbf{(C3)} \emph{Increasing $f_{\log U}$:} 
$f_{\log U}$ is finite and strictly increasing:
$0<\delta_a<\delta_b\Rightarrow 0<f_{\log U}(\delta_a)<f_{\log U}(\delta_b)$.

\item \textbf{(C4)} \emph{Sharpness monotonicity:} 
$\lambda_{1,t}=f_\lambda(U_t)+\zeta_t$, with $\E[\zeta_t|U_t]=0$ and 
$U_a<U_b\Rightarrow\E[\lambda_{1,t}|U_t=U_a]\le\E[\lambda_{1,t}|U_t=U_b]$.

\item \textbf{(C5)} \emph{Initial instability:} 
$\delta_0>0$.

\item \textbf{(C6)} \emph{Exit by construction:}
If $\delta_t\le0$, then $U_{t'}\equiv U_t$ for all $t'\ge t$. Asp.~\ref{asp:exit}, now stated as an explicit condition. 
\end{itemize}
\end{condition}

\noindent
Writing $X_t:=\log U_t$ converts the multiplicative updates into a state-dependent additive random walk:
\begin{equation}\label{eqn:flat:X-updates}
    X_{t+1}=X_t+\xi_t\,\alpha(X_t), 
    \qquad 
    \xi_t\in\{-1,+1\}\ \text{i.i.d. symmetric},
\end{equation}
where $\alpha(x):=\E[f_{\log U}(\delta_t)\mid X_t=x]$ is positive, finite, and strictly increasing (from the standing conditions). 

\paragraph{Stochastic Minibatching} 
We note that the required properties of $\alpha$ may be derived from a relaxed \emph{expected} statement of \textbf{C3}. 
Write \textbf{C3E}: $0<\delta_a<\delta_b\Rightarrow 0<\E[f_{\log U}(\delta_a)]< \E[f_{\log U}(\delta_b)]$.
\textbf{C3E} is a relaxed condition which is used instead when we consider RPE in the stochastic minibatch setting for SGD. We present a detailed RMT argument in App.~\ref{sec:sgd:fb-sgd-sim}. 

\paragraph{Notation}
We use one-step maps $T_\pm(x):=x\pm\alpha(x)$ and, for a sign string $s\in\{+,-\}^t$, $T_s:=T_{s_t}\!\circ\!\cdots\!\circ\!T_{s_1}$. 
The one-step kernel is $K(x,\cdot):=\frac12\delta_{T_+(x)}+\frac12\delta_{T_-(x)}$, with $K^{(t)}(x,\cdot)=2^{-t}\!\sum_s\!\delta_{T_s(x)}$.  
Let $F_t(u):=\Pr(X_t\le u)$, the CDF, $f_t:=F_t'$, the density, and denote the essential supremum function of $X_0$ as $\esssup X_0$. 

As $\E[\xi_t|X_t]=0$, $X_t$ is a martingale, yet state-dependent step sizes induce an effective asymmetry: in any paired two-step pattern ($+-$ or $-+$), the negative step is evaluated at a larger argument and thus has greater magnitude, biasing the process leftward. 
Our main theorem formalizes this atomic leftward drift induced by state-dependent step sizes.

\begin{theorem}[Point-mass contraction (PMC)]\label{thm:flat:point-contraction}
Under Cond.~\ref{cond:flat:standing} and Eqn.~\ref{eqn:flat:X-updates}, for any point-mass $X_0=u$ and all $t\ge1$,
\[
    F_t(u)\ \ge\ \tfrac12 \,,
\]
with strict inequality at even $t$. 
Hence every even-time median $m_t$ satisfies $m_t<u$.
\end{theorem}

Proof in App.~\ref{app:flat:pm-contract-proof}. Paired $+, -$ steps therefore produce a net left drift even without explicit bias. PMC does not extend directly to arbitrary densities because medians are nonlinear functionals. To generalize PMC beyond the atomic setting, we provide mass-balance conditions ensuring immediate contraction of the median.

\begin{corollary}[Mass conditions for median contraction]\label{cor:flat:mass-conditions-median}
Under Cond.~\ref{cond:flat:standing} and Eqn.~\ref{eqn:flat:X-updates}, let $m_0$ be a median of $X_0$. Define
\[
a^- := \lim_{x\uparrow m_0}\alpha(x), 
\qquad 
a^+ := \lim_{x\downarrow m_0}\alpha(x).
\]
Partition the domain into
\[
A=(-\infty,m_0-2a^-],\quad 
B=(m_0-2a^-,m_0],\quad 
C=(m_0,m_0+2a^+],\quad 
D=(m_0+2a^+,\infty),
\]
and let $A,B,C,D$ also denote their respective $X_0$-masses. Then
\[
F_2(m_0)\ \ge\ A+\tfrac34 B+\tfrac14 C.
\]
In particular, if $B<C$ then $F_2(m_0)>\tfrac12$, and any median $m_2$ satisfies $m_2<m_0$.
\end{corollary}

Proof in App.~\ref{app:flat:mass-conditions-median-proof}. 
This result formalizes a simple intuition: if slightly more mass lies below the initial median than above, then after two steps the median must contract leftward. We next derive the contraction of an envelope that upper-bounds the median.

\begin{theorem}[Upper-envelope drift]\label{thm:flat:quantile-drift}
Under Cond.~\ref{cond:flat:standing} and Eqn.~\ref{eqn:flat:X-updates}, let $b:=\esssup X_0<\infty$ and $c:=\alpha_{\min}\alpha'_{\min}>0$.  
For even $t\ge0$, let $m_t(x)$ be a median of $K^{(t)}(x,\cdot)$ and define $Z_t:=\esssup_{x\in\supp(X_0)} m_t(x)$.  
Then:
\begin{enumerate}\itemsep0.15em
\item[\textnormal{(i)}] \textbf{Strict linear decay:}\quad $Z_{t+2}\le Z_t-c$, hence $Z_t\le b-\tfrac{t}{2}c$.
\item[\textnormal{(ii)}] \textbf{Median domination:}\quad $F_t(Z_t)\ge\tfrac12$, so every even-time median of $X_t$ lies $\le Z_t$.
\item[\textnormal{(iii)}] \textbf{Finite crossing:}\quad For $N:=\lceil (b-m_0)/c\rceil$, $Z_{2N}\le m_0$ and for even $t\ge2N$, $Z_t\le m_0-\tfrac{t-2N}{2}c$.
\end{enumerate}
\end{theorem}

Proof in App.~\ref{app:flat:quantile-drift-proof}. Thus, a deterministic envelope decays at least linearly and crosses the initial median in finite time, which allows us to conclude eventual median and quantile drift for arbitrary distributions. 
However, worst-case median drift alone does not characterize the overall shape of $f_t$ to enough specificity. We introduce mild conditions for additional control: unimodal initializations and $1$-Lipschitz regularity on $\alpha$. These conditions allow us to derive persistent skewness, and a general median drift in $X_t$ for a general flatness bias for GD in instabilities. 

\begin{condition}[Unimodal initialization]\label{cond:flat:unimodal}
$X_0$ is absolutely continuous with density $f_0$, nondecreasing on $(-\infty,m^*]$ and nonincreasing on $[m^*,\infty)$ for some mode $m^*$.
\end{condition}

\begin{condition}[$1$-Lipschitz control of $\alpha$]\label{cond:flat:alpha-regularity}
On the visited range, $\alpha\in C^1$ with $0<\alpha'(x)\le L_\alpha<1$.
\end{condition}

These conditions lead to order preservation of the map $T$, which then allows comparison of local densities and yields monotone tails up to a narrow central band. The next lemmas use order preservation to establish shape preservation and controlled monotonicity with high probability outside a central band. 

\begin{lemma}[Two-sided monotonicity]\label{lem:flat:edge-strict-descent}
Under Eqn.~\ref{eqn:flat:X-updates} and Conds.~\ref{cond:flat:standing}, \ref{cond:flat:unimodal}, \ref{cond:flat:alpha-regularity}, fix $t\ge1$ and $w>0$. With
\[
y_{-}(t):=T_{-\cdots-}(m^*), \qquad y_{+}(t):=T_{+\cdots+}(m^*),
\]
$\hat f_t^{(w)}$ is nondecreasing on $(-\infty,y_-(t)-w]$ and nonincreasing on $[y_+(t),\infty)$. If $f_0$ has no positive-length plateaus, these are strict a.e. Moreover,
\begin{equation}\label{eqn:flat:strict-descent-far-enough}
y_+(t)-y_-(t)\ \le\ 2\!\sum_{j=0}^{t-1}\sup_{\text{range at step }j}\alpha\ \le\ 2t\,\alpha_{\max},
\end{equation}
with $\alpha_{\max}:=\sup_x\alpha(x)$.
\end{lemma}

\begin{lemma}[CLT-controlled monotone tails]\label{lem:flat:clt-descent}
Under Eqn.~\ref{eqn:flat:X-updates} and Conds.~\ref{cond:flat:standing}, \ref{cond:flat:unimodal}, \ref{cond:flat:alpha-regularity}, fix $t\ge1$, $w>0$, $\kappa\ge1$. Define:
\[
W_t(\kappa):=\frac{2}{1-L_\alpha}\,\alpha_{\rm loc}\,\kappa\sqrt t,\qquad
B_t(\kappa):=[\,m^*-W_t(\kappa)-w,\ m^*+W_t(\kappa)\,].
\]
If $\{x_1,x_2\}\cap B_t(\kappa)=\varnothing$ and $x_1<x_2$, then
\[
\hat f_t^{(w)}(x_2)-\hat f_t^{(w)}(x_1)\ \le\ \frac{4}{w}e^{-\kappa^2/2}
\quad\text{(right side),}
\]
with the reversed inequality on the left. Thus outside $B_t(\kappa)$, $\hat f_t^{(w)}$ is monotone up to $O(e^{-\kappa^2/2})$.
\end{lemma}

Proofs for these lemmas are in App.~\ref{app:flat:strict-descent-proof},\ref{app:flat:clt-descent-proof}. Lemma~\ref{lem:flat:edge-strict-descent} offers a strict monotonicity result, but the bands grow linearly in $t$, which is limiting in application. Lemma~\ref{lem:flat:clt-descent} introduces tighter probabilistic monotonicity. The width of the central bands are controlled through the factor $\kappa$, which also upper bounds the probability of outliers. In other words, $\kappa$ introduces a variable tradeoff in the certainty of monotonicity versus its domain of applicability.

Using these shape constraints, the following result shows that the mode always precedes the median under evolution. 

\begin{proposition}[Median dominates mode]\label{prop:flat:median-dominate-mode}
Under Eqn.~\ref{eqn:flat:X-updates} and Conds.~\ref{cond:flat:standing}, \ref{cond:flat:unimodal}, \ref{cond:flat:alpha-regularity}, if $m^*$ is a global mode and $m_t$ any median of $X_t$, then $m^*\le m_t$, with strictness unless $f_t$ has a plateau at $m_t$.
\end{proposition}

Proof in App.~\ref{app:flat:median-dominate-mode-proof}. The mode-median ordering forms the basis of skew properties of this distribution, indicating a right-skew under further conditions. 

Our second major theoretical result combines order preservation, CLT-controlled tails, and local drift bounds to derive monotone contraction of even-time medians with high probability. 

\begin{theorem}[Even-time median drift]\label{thm:flat:median-drift-unimodal}
Under Eqn.~\ref{eqn:flat:X-updates} and Conds.~\ref{cond:flat:standing}, \ref{cond:flat:unimodal}, \ref{cond:flat:alpha-regularity}, let $m_t$ be a median of $X_t$. Then:
\begin{enumerate}\itemsep0.15em
\item[\textnormal{(a)}] \textbf{High-probability decrease:}\quad with high probability, $m_{t+2}<m_t$.
\item[\textnormal{(b)}] \textbf{Quantified local drift:}\quad if $\alpha(m_t)>0$ and 
\[
\underline\alpha'(m_t):=\inf_{u\in[m_t-r_t,m_t+r_t]}\alpha'(u)>0
\]
for some $r_t>0$ with $\Pr\{|X_t-m_t|\le r_t\}\ge p_t>0$, then
\[
m_{t+2}\ \le\ m_t - p_t\,\alpha(m_t)\,\underline\alpha'(m_t).
\]
\end{enumerate}
\end{theorem}

Proof in App.~\ref{app:flat:median-drift-unimodal-proof}. Collectively, these results establish that state-dependent symmetric updates produce a systematic leftward contraction of $X_t=\log U_t$ in density. 
In probabilistic terms, instability drives the distribution of $U_t$ toward smaller values, which are flatter orientations of the loss landscape.  
This contraction and induced right-skew in $X_t$ form the mathematical basis of the implicit flatness bias for GD in instabilities.

\medskip

\subsection{Implicit Bias for Flatness in Gradient Descent during Instabilities} \label{sec:flat:flatness-summary}

The theoretical results in Subsec.~\ref{sec:flat:dyn-system} provide an account of how eigenvector rotations during instability, characterized via RPE, provably induces a flatness bias in GD.  
Under the mild conditions (Conds.~\ref{cond:flat:standing}, \ref{cond:flat:unimodal}, \ref{cond:flat:alpha-regularity}), the stochastic updates of $X_t=\log U_t$ behave as a state-dependent random walk whose even-time medians contract monotonically (Thm.~\ref{thm:flat:median-drift-unimodal}).  
This contraction formalizes a systematic reduction in the density and typical curvature magnitude: instability increases curvature variability, but the feedback mechanism drives the mass of $U_t$ downward over time. 

Moreover, the distribution of $X_t$ becomes increasingly right-skewed.  
High-probability unimodality (Lems.~\ref{lem:flat:edge-strict-descent}, \ref{lem:flat:clt-descent}) combined with the mode-median ordering (Prop.~\ref{prop:flat:median-dominate-mode}) implies that while the center of mass shifts left, vanishing probability mass spread over widening support in the right tail.  
In practice, GD spends progressively more time in regions of lower higher-order curvature moments, $U_t$, formalizing the implicit flatness bias observed empirically.

In App.~\ref{sec:flat:xt-sim} we conduct a numerical simulation of one realization of $X_t$, which align closely with these predictions. In the well-behaved regime, we observe unimodality, leftward drift of both median and mode, median dominance, and a gradual widening of the central distribution with mass concentrated on smaller values of $X_t=\log U_t$. 

In the simulations, when the $1$-Lipschitz condition on $\alpha$ is relaxed, unimodality may break, yet median drift and right-skew persist, accompanied by faster expansion of the central width. With bimodal initializations, both the median and mode continue to drift left, though a secondary peak remains, showing that unimodality is not guaranteed. 

Together, these behaviors illustrate the \emph{flatness bias of GD during instabilities}: over time, the process allocates more probability mass to regions where $U_t$ is smaller, while the support in the flanks increase, producing a widening, and increasingly right-skewed distribution. Smaller $U_t$ corresponds to a flatter effective landscape, supporting the claim that GD (during instabilities) biases training toward flatter regions of the loss surface and thereby improves generalization.

\section{Flatness Bias of GD on Computer Vision}
\label{sec:flat:cv-exp}

Our theoretical analysis predicts an implicit bias toward flatter eigen-directions in gradient descent trajectories, strengthened by large learning rates and sustained instability. We now test this prediction on CIFAR10, a standard computer vision benchmark, examining how instability influences generalization and flatness. We find that improvements in generalization occur \emph{only} when training operates in the unstable regime, and that the resulting minima are markedly flatter. All experiments in this section are conducted in a deterministic setting, which is computationally expensive. To mitigate this, we use reduced datasets for the main study and benchmark on the full CIFAR10 at the end. 

\subsection{Measuring flatness}

Our flatness-bias theorems predict that, after training in GD instabilities, HOCMs (via $U_t$) decrease, indicating movement toward flatter regions of the loss landscape. To test this empirically, we require a measure of curvature beyond the order of $\lambda_{\max}$. 

Recall from Sec.~\ref{ch:phases} that gradient descent exhibits two distinct behaviors:
\begin{itemize}
    \item \emph{Progressive sharpening:} in stable training, $\lambda_{\max}$ gradually increases as the model drifts along flat directions;  
    \item \emph{Instabilities:} as $\lambda_{\max}$ approaches the stability threshold $2/\eta$, sharpness and loss spike due to valley-jumping.
\end{itemize}
The former reflects slow geometric drift within a basin, while the latter arises from curvature transitions near instability.

In low learning-rate regimes without instabilities, the increase in $\lambda_{\max}$ arises solely from progressive sharpening. Although the mechanisms behind progressive sharpening remain under study, we conjecture that reducing HOCMs also weakens this effect. Thus, the terminal curvature from stable training, where eigenvectors remain largely fixed, can serve as a proxy for HOCM contributions. 

During instabilities, our flatness-bias theory predicts a long-term \emph{progressive flattening}, where $U_t$ decreases as trajectories move toward flatter regions. By tuning the magnitude and duration of large initial learning rates, we can modulate this implicit bias. Subsequent stable training with a small $\eta$ then produces final curvatures, allowing us to quantify cumulative flattening from earlier instabilities through reductions in progressive sharpening. 

\subsection{Experimental setup}

Generalization is assessed via the gap between training and test accuracy after training completion, which we define as reaching $>99.99\%$ train accuracy. We conduct experiments on CIFAR10 using VGG-like networks~\cite{simonyan2015vgg}, and on fMNIST, using cross-entropy loss. All runs employ full-batch GD in a fully-deterministic setting. This means that no stochastic data augmentation is used, and batch normalization is replaced with GhostBatchNorm~\cite{hoffer2017trainlonger}, which is computed deterministically with a fixed batch size of $1024$.  
To manage resources, we train hundreds of small VGG10s on a $5$k ($10\%$) CIFAR10 subset, then repeat the key experiments on the full $50$k dataset. 
(For details, see App.~\ref{app:exp-dets}.)

\subsection{Effect of large learning rates}

\begin{figure}[t]
  \centering 
  \includegraphics[width=0.40\textwidth]{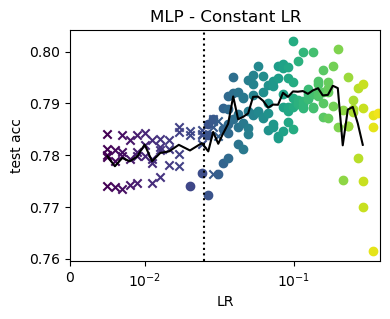}
  \includegraphics[width=0.44\textwidth]{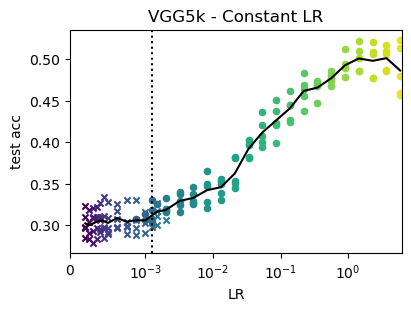}
  \caption[Generalization improves only in instability.]{\textbf{Generalization improves only in instability.} Validation accuracy vs.\ learning rate for MLPs on fMNIST (left) and VGG10s on CIFAR10-5k (right). The dotted line marks the stability threshold $2/\lambda_{\max}$.}
  \label{fig:flat:goldilocks}
\end{figure}

We sweep learning rates $\eta$ from $10^{-4}$ to the divergence boundary, sampling exponentially with task-specific scaling (scaling factor $m=1.1$ for fMNIST, $1.6$ for CIFAR10-5k).  
Validation accuracy remains flat below the stability limit and rises sharply once $\eta$ enters the unstable regime, revealing a pronounced generalization benefit only during \emph{instabilities}. The transition mirrors the RPE phase distinction and its monotonic dependence on learning rate.
Excessively large $\eta$ eventually degrades performance, yielding a \emph{Goldilocks zone} of learning rates where training instabilities aid regularization without leading to divergence.  

The extent of this zone depends on task complexity: for CIFAR10, the beneficial range of $\eta$ spans nearly an order of magnitude, whereas fMNIST shows a narrower window-yet both exhibit the same qualitative trend. These results indicate that instability-driven flattening under GD benefits even simple, nearly linearly separable problems.

\begin{figure}[H]
  \centering
  \begin{subfigure}{0.35\textwidth}
    \includegraphics[width=\linewidth]{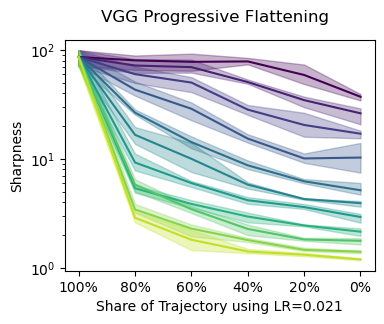}
    \caption{Flattening}
  \end{subfigure}\hfill
  \begin{subfigure}{0.35\textwidth}
    \includegraphics[width=\linewidth]{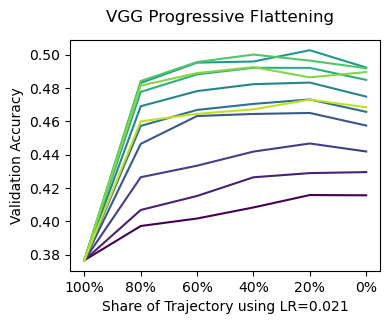}
    \caption{Generalization}
  \end{subfigure}\hfill
  \begin{subfigure}{0.1\textwidth}
    \includegraphics[width=\linewidth]{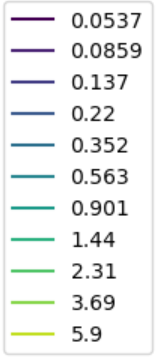}
    \caption*{$\eta$}
  \end{subfigure}
  \caption[Progressive flattening and generalization on CIFAR10.]{\textbf{Progressive flattening and generalization on CIFAR10.} VGG10s trained on CIFAR10-5k with large $\eta$, switching to $\eta_\mathrm{small}=0.021$ at different stages. Longer exposure to large $\eta$ yields flatter minima and better generalization.}
  \label{fig:flat:cifar-flat}
\end{figure}

\subsection{Learning rate reduction}

Reducing the learning rate provides an effective probe of accumulated flattening. We train VGG10s with large $\eta$ until training accuracy exceeds thresholds $60\%, 70\%, 80\%, 90\%$, and then we switch to low stable learning rates $\eta_\mathrm{small}=0.021$
As shown in Fig.~\ref{fig:flat:cifar-flat}, longer initial training with large $\eta$ leads to lower final sharpness and higher test accuracy. 
Even extreme $\eta$ values can improve generalization when reduced near the end of training, aligning with common learning-rate scheduling practices, though our results reveal that the benefit is sensitive to timing.

\subsection{Benchmark on the full dataset}

To consolidate the preceding trends, we repeat experiments on the full CIFAR10-50k dataset. To remain in the deterministic setting, we train VGG19s without data augmentation, then include static augmentations (crops and flips) to form a $10\times$ augmented CIFAR10-500k dataset~\cite{geiping2022fbgd}. Finally, we evaluate ResNet20s on the same augmented set.  
Results (Table~\ref{fig:flat:cifar50k}) confirm that larger learning rates consistently improve generalization. 
For these models, we omit explicit stability-limit computation due to cost but intentionally select large $\eta$ values to ensure instability. The best performance occurs for ResNet20 at $\eta=3.2$; higher learning rates (e.g.\ $6.4$) diverge.  

Overall, the empirical evidence supports a wide \emph{Goldilocks region} several orders of magnitude above the classical stability bound yet below divergence, where instability-induced flattening yields the best generalization. Practically, we recommend exploring learning rates substantially beyond the descent-lemma limit (e.g.\ $\eta \approx 0.4$ in~\textcite{geiping2022fbgd}) to fully realize this flatness bias in practice. 

\begin{figure*}[t]
  \centering
  \begin{tabular}[b]{c|cccccc}
    $\eta$ & 0.1 & 0.2 & 0.4 & 0.8 & 1.6 & 3.2 \\
    \hline\hline
    VGG19 on CIFAR10-50k & 67.1 & 68.4 & 70.5 & 73.4 & 72.7 & 73.6\\
    VGG19 on CIFAR10-500k & 78.1 & 80.6 & 81.7 & 82.9 & 84.0 & 84.1\\
    ResNet20 on CIFAR10-500k & 83.7 & 85.6 & 86.1 & 86.5 & 87.0 & \textbf{87.3}\\
  \end{tabular}
  \captionof{table}[Large learning rates improve generalization on CIFAR10.]{\textbf{Large learning rates improve generalization on CIFAR10.} Validation accuracy across architectures and datasets.}
  \label{fig:flat:cifar50k}
\end{figure*}

\section{\label{ch:sgd}Flatness Bias under SGD}

Here, we consider instability-induced flatness bias in the stochastic minibatch setting for SGD.  
Theoretically, our framework carries over naturally: random-matrix arguments show that the leading eigenstructure of the minibatch Hessian closely matches that of the full-batch Hessian \cite{benaych2010rmt, granziol2020lr}, requiring only monotonicity in expectation in $f_{\log U}$, which our theory already accounts for (for details, see discussion in Subsec.~\ref{sec:flat:dyn-system} and RMT arguments in App.~\ref{sec:sgd:fb-sgd-sim}). 

Empirically, large learning rates induce strong benefits to generalization. Our empirical study in this section finds that this benefit outweighs what is predicted by the SDE-view of SGD, indicating that instability-driven benefits take a dominant role in generalization with SGD. Additionally, we validate that \emph{progressive flattening}, the flattening of HOCMs measured via stable training proxies, persist under SGD. Experimental details are provided in Appendix~\ref{app:exp-dets}. 

\subsection{Comparable generalization under the SDE-view of SGD}

Under the SDE-view, minibatch SGD is modeled as an SDE whose drift follows gradient flow and whose diffusion arises from gradient-estimation noise~(Eqn.~\ref{eq:prelim:sgd-sde}).
For an isotropic scalarization of the gradient-noise covariance, the \emph{effective temperature} scales as: 
\begin{equation}
T_{\mathrm{eff}}(\vtheta) \;\propto\; \eta \;\frac{N}{B}\;\bar{\sigma}^2(\vtheta),
\label{eq:teff_base}
\end{equation}
where $N$ is the dataset size, $B$ the minibatch size, and $\bar{\sigma}^2(\vtheta)$ a scalarization (e.g., trace) of the gradient-noise covariance $ \Sigma(\vtheta) = \operatorname{Cov_i}(\nabla \ell(\vtheta, x_i))$. We introduce a noise-scale parameter $\alpha$, and write the discrete update as:
\begin{equation}
\vtheta_{t+1} \;=\; \vtheta_t - \eta\,\vg_t(\vtheta_t),
\qquad 
\vg_t(\vtheta_t) \;=\; \nabla \mathcal{L} (\vtheta_t) + \alpha \vxi_t,
\label{eq:update_alpha}
\end{equation}
where $\vxi_t$ is zero-mean gradient noise with covariance $\operatorname{Cov}(\vxi_t \mid \vtheta_t) = \Sigma(\vtheta_t)$.  
In the SDE limit with $\mathrm{d}t=\eta$, the diffusion coefficient scales linearly with $\alpha$, so that 
\begin{equation}
T_{\mathrm{eff}}(\vtheta) \;\propto\; \eta \;\frac{N}{B}\;\alpha^2\,\bar{\sigma}^2(\vtheta).
\label{eq:teff_alpha}
\end{equation}
In this view, trajectories with larger effective temperature $T_{\mathrm{eff}}\!\propto\!\eta\alpha^2$ should explore flatter regions and generalize better; those conserving $\eta\alpha^2$ (with fixed $N/B$) should behave equivalently.

\subsection{SDE-view fails to explain empirical generalization}

\begin{figure}[t]
  \centering
  \includegraphics[width=0.7\linewidth]{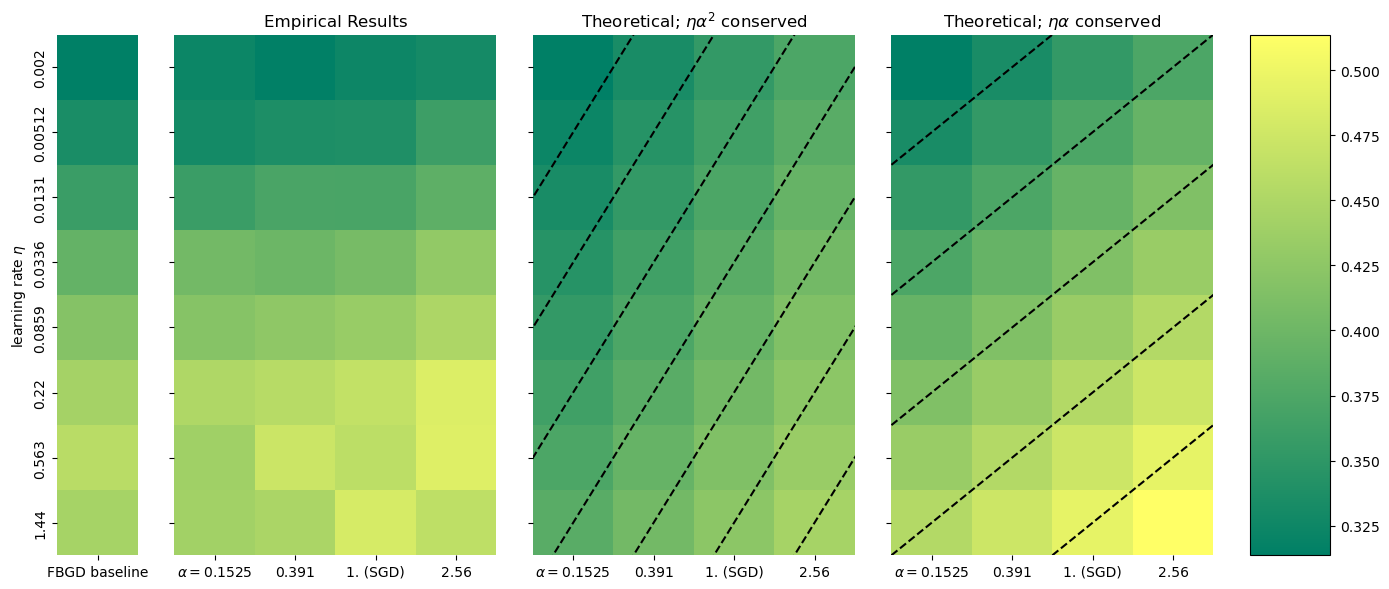}
  \caption[Generalization performance in SGD for varying $\eta$ and $\alpha$.]{\textbf{Generalization performance in SGD for varying $\eta$ and $\alpha$.}  
  VGGs on CIFAR5k trained with controlled noise-scale $\alpha$.  
  \textbf{Far-left:} GD baseline; \textbf{Left:} empirical results; \textbf{Centre:} expected generalization under SDE-view ($\eta\alpha^2$ conserved); \textbf{Right:} proxy with constant absolute noise ($\eta\alpha$ conserved); \textbf{Far-right:} empirical color-map of generalization.}
  \label{fig:sgd:fbst-noise-eta}
\end{figure}

\begin{figure}[t]
  \centering
  \includegraphics[width=1.\linewidth]{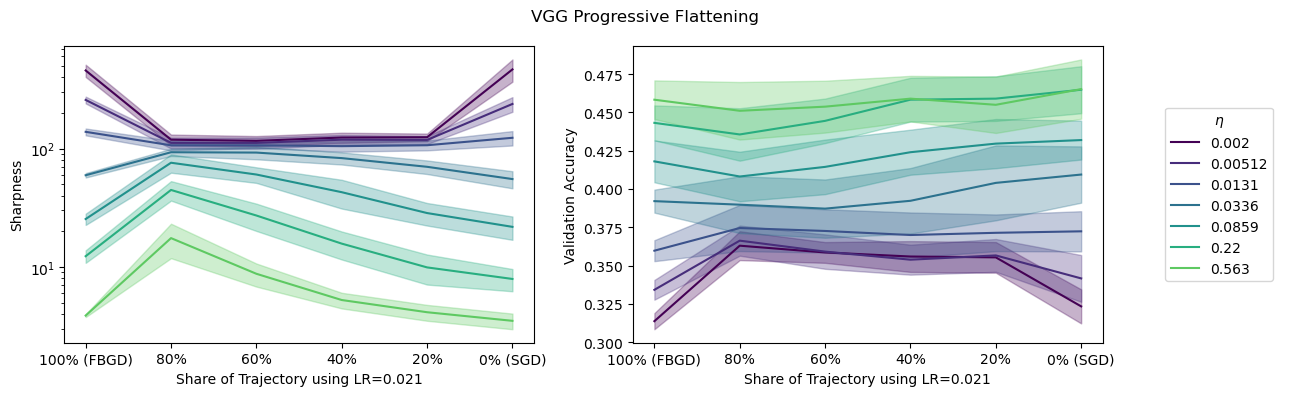}
  \caption[Progressive flattening in SGD.]{\textbf{Progressive flattening in SGD.}  
  Models trained with large-$\eta$ SGD then switched to low-$\eta$ GD at different fractions of training.  
  Left: final sharpness; Right: test accuracy; mean and $1\sigma$ over 5 seeds.}
  \label{fig:sgd:fbst-prog-flat}
\end{figure}

We train VGG networks on CIFAR5k across grids of $(\eta,\alpha)$, where $\alpha=0$ recovers full-batch GD and $\alpha=1$ yields standard SGD. Each epoch computes a full-batch gradient plus scaled minibatch noise, providing precise control over $\alpha$.  
Fig.~\ref{fig:sgd:fbst-noise-eta} compares empirical generalization (left) with SDE-view predictions ($\eta\alpha^2$ conserved) and a proxy assuming constant absolute noise ($\eta\alpha$ conserved). 
For instance, matching $(\eta,\alpha)=(0.002,1)$ to $\alpha=0.39$ implies $\eta\approx0.013$. 
However, empirically, this higher learning rate yields \emph{better} generalization, directly contradicting the diffusion prediction.

Across all regimes, increasing $\eta$ improves generalization even when $\alpha$ is reduced to maintain constant $\eta\alpha^2$, contradicting diffusion predictions. $\eta$-driven effects dominate noise-driven ones, indicating that effective temperature alone cannot explain generalization.

These results confirm that instability-driven effects, such as anisotropic noise–curvature coupling and non-equilibrium dynamics, ignored in diffusion approximations, dominate generalization behavior in SGD.

\subsection{Progressive flattening persists under SGD}

We next test whether the flattening effect persists in stochastic training.  
VGGs on CIFAR10-5k are trained with large-$\eta$ SGD and then switched to low-$\eta$ GD at various stages; the batch size is $32$ ($<1\%$ of the training set).  
We use $\eta_\mathrm{low}=0.021$ for the final stable phase, following the setup of Sec.~\ref{sec:flat:cv-exp}.  
Learning-rate scaling follows the linear rule between $\eta$ and batch size~\cite{goyal2017accurate,granziol2020flatness} (details in App.~\ref{app:sgd:scaling}).

Fig.~\ref{fig:sgd:fbst-prog-flat} shows final sharpness and validation accuracy as the switch point varies.  
Pure GD and SGD reach similar sharpness, yet large-$\eta$ SGD consistently generalizes better.
Longer exposure to large-$\eta$ phases yields flatter minima and higher test accuracy, with full-SGD performing best.

These results show that generalization gains under SGD arise not from gradient noise alone, but from instability-driven dynamics that persist even in stochastic training.

\section{\label{ch:ada}Leveraging Instabilities for Adaptive Optimizers}

Having established the implicit bias from training instabilities in GD and SGD, we now examine whether similar effects can be harnessed in adaptive methods such as RMSprop~\parencite{hinton2012rms} and Adam~\parencite{kingma2017adam}.  
These adaptive optimizers can be viewed as preconditioned forms of gradient descent (see App.~\ref{app:adaptive:precond-gd}), where the adaptive rescaling of gradients suppresses updates along sharp directions.  
This suppression improves stability, but may hinder the beneficial flattening bias from training in instabilities.  
To address this, we introduce an \emph{instability-aware} variant of adaptive optimization, which we call \emph{Clipped-Ada} (CA), that selectively restores sharp-direction dynamics, yielding improved generalization in practice. 

\subsection{Clipped-Ada: restoring instabilities in a controlled manner}

Following \textcite{kingma2017adam}, we write the update for Adam as:
\[
    \vtheta_{t+1} = \vtheta_t - \eta \cdot \frac{\vm_t}{\sqrt{\vq_t} + \epsilon \mI}
\]
where $\vm_t$ and $\vq_t$ are exponential moving averages of the gradient and squared gradient \footnote{We use the notation $\vq$ to denote the $2$nd-order moment estimate, which is traditionally denoted by $\vv$ in the literature, to avoid confusion with the global notation $\vv$ for eigenvectors.}, respectively, and $\epsilon$ is a small constant for numerical stability in division. The updates to $\vm_t$ and $\vq_t$ are controlled by hyperparameters $\beta_1$ and $\beta_2$, respectively, which commonly are set to $\beta_1=[0.9]$ and $\beta_2=[0.99, 0.999]$ for Adam. Setting the hyperparameter $b_1=0$, which means $\vm_t=\vg_t$, recovers RMSprop, which can be seen as a special case of Adam. 

App.~\ref{app:adaptive:precond-gd} shows how these optimizers can be viewed as preconditioned gradient-descent, and the role of adaptive optimization in suppressing training instabilities. To relax the suppression of instabilities, we introduce \emph{Clipped-Ada}, a unified adaptive optimizer that is leverages the flatness bias from training instabilities.  
Crucially, CA caps each element of the second-moment vector $\vq_t = \{p_{i,t}\}$:
\begin{equation}
    q^{\text{mod}}_{i,t} = \min(p_{i,t}, q^{\text{thresh}}), \quad \text{to get the CA update rule:} \quad
    \vtheta_{t+1} = \vtheta_t - \eta \cdot \frac{\vm_t}{\sqrt{\vq^{\text{mod}}_t} + \epsilon}\, ,
\end{equation}
where the division is coordinate-wise.  
The modified preconditioner therefore limits curvature suppression when $p_{i,t}$ exceeds the clipping threshold $q^{\text{thresh}}$, effectively re-enabling larger steps along over-regularized directions.

We can express the corresponding effective curvature as:
\begin{equation}
    \lambda_{\text{eff}}(\vv_1) \;\approx\; 
    \lambda_\mathrm{max} \Big/ \sum_i v_{1,i}^2 \,(q^{\text{mod}}_{i,t}+\epsilon) \,.
\end{equation}
Instability arises when $\eta\,\lambda_{\text{eff}}(\vv_1) > 2$, or equivalently when the threshold $q^{\text{thresh}}$ is small enough that adaptive suppression is constrained.  
In this sense, $q^{\text{thresh}}$ serves as a tunable parameter controlling the degree of instability:
\begin{itemize}
    \item $q^{\text{thresh}} = \infty$: standard adaptive optimization (full suppression);
    \item $q^{\text{thresh}} > 1$: mild suppression;
    \item $q^{\text{thresh}} = 1$: GD-like stability thresholds;
    \item $q^{\text{thresh}} < 1$: enhanced step sizes and instabilities.
\end{itemize}
Derivations of the stability behavior of CA is in App.~\ref{app:adaptive:clipped-ada-deriv}. 
CA thus interpolates between the strong stability of adaptive optimizers and the unrestrained dynamics of GD, providing a tunable parameter controlling instabilities during training.

\subsection{Empirical generalization with Clipped-Ada}
\label{sec:ada:experiments}

We evaluate CA using VGG networks on CIFAR10-5k with RMSprop and Adam in both full-batch (GD) and minibatch (SGD) regimes.  
Learning rates are swept in the range $10^{-4}$–$10^{-3}$, and the clipping threshold $q^{\text{thresh}}$ is varied over a broad grid.  
Each $(\eta, q^{\text{thresh}})$ configuration is repeated over five random seeds. For details, see Appendix~\ref{app:exp-dets}.

\begin{figure}[t]
    \centering
    \includegraphics[width=0.48\textwidth]{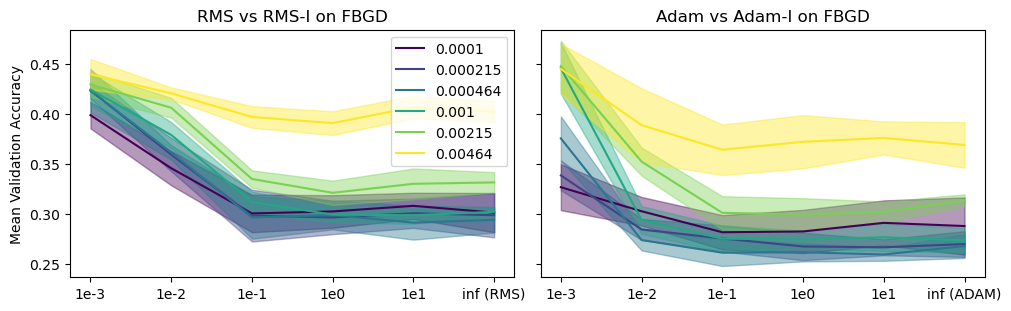}
    \includegraphics[width=0.48\textwidth]{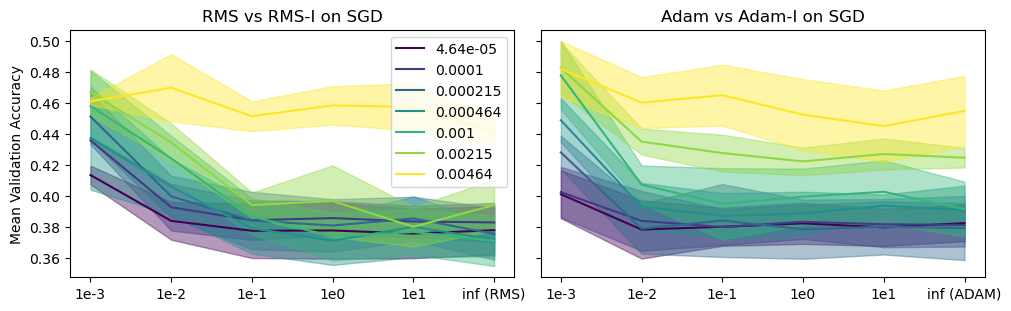}
    \caption[Performance of Adaptive Optimizers on CIFAR10-5k.]{\textbf{Performance of Adaptive Optimizers on CIFAR10-5k.}  
    VGGs trained with RMSprop and Adam across learning rates $\eta$ and clipping thresholds $q^{\text{thresh}}$.  
    Each pane shows mean ± 1 SD over five seeds. }
    \label{fig:ada:fb-gen}
\end{figure}

As shown in Fig.~\ref{fig:ada:fb-gen}, both optimizers exhibit improved generalization with larger learning rates, paralleling the trends observed for GD and SGD (Fig.~\ref{fig:flat:goldilocks}).  
Adam shows a narrower gap between small- and large-$\eta$ runs, consistent with stronger suppression of sharp-direction updates.  
Varying $q^{\text{thresh}}$ modulates this suppression: tighter clipping (smaller $q^{\text{thresh}}$) restores instabilities in a controlled manner along dominant curvature directions and improves validation accuracy, while very loose clipping behaves like standard adaptive optimization.

Comparing full-batch and minibatch results, the qualitative behavior is consistent: large learning rates and moderate clipping yield the best out-of-sample performance in both regimes.  
These findings reinforce the view from Sec.~\ref{ch:sgd} that instability-driven dynamics, rather than stochastic noise alone, are central to the generalization benefits of modern optimizers.  
By tuning $q^{\text{thresh}}$, CA is a simple method to leverage these effects in practice.

\section{\label{ch:conclusion}Conclusion} 

We have shown that training \emph{instabilities} need not be detrimental. 
Viewed through the geometry of Hessian dynamics, they can actively reorient the subspaces explored by gradient-based optimizers. 
By identifying the \emph{Rotational Polarity of Eigenvectors} and formalizing a state-dependent dynamical system for higher-order curvature moments, we derived a set of distributional consequences: median contraction; persistent right-skew; and monotone tails, that together constitute an implicit \emph{flatness bias} in gradient descent under instability. 
This theory explains why, in practice, models trained with sufficiently large learning rates tend to migrate toward flatter, more trainable regions.

Empirically, we verified the predicted eigenvector rotations and their monotonic dependence on learning rate, observed progressive flattening in both full-batch GD and minibatch SGD, and identified a `Goldilocks' regime in which large learning rates improve generalization without inducing irreversible divergence. 
Extending the analysis to SGD, we found that the SDE minibatch-noise arguments alone underpredict the benefits of large learning rates, indicating that anisotropic effects and curvature–rotation couplings are essential to modern training dynamics. 
Finally, we propose \emph{instability-aware} variants of adaptive optimization (Clipped-Ada) that restore instabilities in a controlled manner and improve generalization.

This study is limited by the restricted scope to multilayer perceptrons for the theoretical derivations of RPE and the reliance on finite-sample estimates of higher-order curvature moments for empirical validation of $U_t$ dynamics. 
Future work should extend both the theoretical framework and empirical evaluations to more complex, large-scale architectures, and develop instability-aware optimizers under broader conditions.

However, our results suggest that training instabilities are not a pathology to avoid, but a mechanism to harness in deep learning optimization. 
Practically, this points to schedules that embrace larger learning rates early and consolidate with smaller rates once flatter geometry is located, routines that are fortunately already common in practice. 

More importantly, our work invites a conceptual re-examination of the role of rotations and instability in deep learning optimization. 
We hope this study encourages further theoretical and empirical work on the interaction between geometry and dynamics in modern networks. 
In particular, higher-order analyses that capture non-quadratic effects, and dynamical studies that explicitly model the time evolution of curvature and orientation, appear crucial for understanding and harnessing instability in large-scale deep neural networks.


\section*{Author Contributions}

\emph{LW}: Conceptualization, Methodology, Software, Validation, Formal Analysis, Investigation, Data Curation, Writing-Original Draft, Writing-Reviewing \& Editing, Visualization, Project administration.  

\emph{SJR}: Conceptualization, Resources, Supervision, Writing-Reviewing \& Editing, Project administration, Funding acquisition. 

\section*{Acknowledgements}

LW gratefully acknowledges support from the Engineering and Physical Sciences Research Council [grants EP/T517811/1 \& EP/W524311/1].



\appendix
\newpage
\makeatletter
\renewcommand\thesection{\Alph{section}}           
\renewcommand\thesubsection{\Alph{section}.\arabic{subsection}} 
\makeatother




\section{Related Works}\label{app:related}

\paragraph{Large learning rates, flatness, and generalization} 
Several works have explored the impact of large learning rates on generalization and loss landscape geometry. \citet{keskar2017largebatch} and \citet{jastrzebski2018threefactors} found that increasing the learning rate could improve generalization by driving optimization toward flatter minima, which are empirically associated with better generalization. Extending this perspective, \citet{mohtashami2023gdlargelr} proved that gradient descent with large learning rates can escape local minima and converge to global optima in certain non-convex settings.

Our work aligns with and deepens this line of inquiry. Specifically, we demonstrate that learning rates large enough to induce instability result in a provable flattening effect on the loss landscape, formalized in Theorems \ref{thm:flat:point-contraction} \ref{thm:flat:median-drift-unimodal}. Furthermore, we empirically verify that this flattening correlates with improved generalization. Closely related is the work of \citet{wang2022largelearningratetames}, who analyzed GD under large learning rates in a homogeneous matrix factorization setting, establishing convergence and a balancing implicit bias. Our analysis generalizes this by introducing the concept of rotational polarity of eigenvectors (RPE), a novel, geometry-driven mechanism that explains how instabilities can bias optimization trajectories toward flatter regions of the loss surface.

\paragraph{Large initial learning rates} 
Numerous studies have highlighted the benefits of employing large learning rates in the early stages of training. \citet{you2019doeslearningratedecay} suggested that high initial learning rates can accelerate convergence and assist in escaping suboptimal basins. In a complementary view, \citet{jastrzebski2020breakeven} argued that large learning rates reduce gradient variance and improve the conditioning of the covariance matrix of gradients.

\citet{li2020initiallarge} further distinguished between the roles of large and small learning rates, observing that large learning rates favor the acquisition of generalizable features, while small learning rates promote memorization. More recently, \citet{andriushchenko2023sgdlargestepsizes} proposed the concept of loss stabilization, where early large learning rates result in generalization-enhancing regularization phases.

Our theoretical contributions align with these findings: we prove that instabilities caused by large learning rates lead to an accumulation of flattening over time. This effect, when occurring early in training, has the potential to guide optimization toward regions in parameter space that promote generalization. However, contrasting perspectives exist. For instance, \citet{li2020intrinsiclr} introduced the idea of the intrinsic learning rate, the product of the learning rate and the weight decay coefficient, as a more consistent predictor of training dynamics and generalization than the raw learning rate alone.

\paragraph{The catapult mechanism}
The catapult effect was first described by \citet{lewkowycz2020largelearningratephase} as a phenomenon where learning rates beyond the stability threshold push models from sharp to flatter basins, enhancing generalization. \citet{zhu2024quadraticcatapult} formalized this idea using neural quadratic models, while \citet{zhu2024catapultssgd} emphasized the implicit feature-learning advantages driven by such dynamics.

Our findings are consistent and builds upon these insights. Specifically, we show that within the instability regime, gradient descent has a provable bias toward flatness, which we attribute to the mechanism of rotational polarity of eigenvectors. Additionally, \citet{damian2023selfstab} explored stabilization in the presence of a single unstable eigenvalue, attributing convergence to cubic curvature terms in the local Taylor expansion. Our analysis engages with higher-order terms as well, but we do not rely on the progressive sharpening condition ($\alpha \coloneq -\nabla L(\theta) \cdot \nabla S(\theta) > 0$). Instead, we focus on the rotational dynamics of the Hessian's leading eigenvectors and their implicit regularization effects.

\paragraph{Edge of Stability and sharpness dynamics}
\citet{cohen2022eos} introduced two key phenomena in deep learning optimization:\begin{enumerate}
    \item \textit{Progressive Sharpening}:"So long as $S(\theta) \leq 2/\eta$, GD has an overwhelming tendency to continually increase $S(\theta)$". 
    \item \textit{Edge of Stability}:"Once $S(\theta)\approx 2/\eta$, it sits at, or just above, the stability threshold. Additionally, although the \emph{descent lemma} does not guarantee a decrease in $L(\theta)$, it nonetheless continues to decrease, albeit non-monotonically".
\end{enumerate}

These observations inspired follow-up theoretical analyzes. \citet{Li2022sharpnessgd} modeled the transition phases of sharpness. \citet{zhu2023understanding} and \citet{ahn2023learning} analyzed how the edge-of-stability regime affects learning thresholds and neuron behavior. \citet{kalra2025universalsharpness} introduced a UV-model framework for a phase-analysis of sharpness dynamics and its relation to chaos.

Our work is consistent with this family of research, particularly in observing that instability does not prevent generalization but may promote it. Most notably, our findings build on \citet{arora2022gd_eos}, who showed that in stable regimes, the gradient aligns with the leading Hessian eigenvector $\vv_1$. We show that during instability, this alignment reverses, a key aspect of the rotational polarity effect, resulting in directional rather than scalar regularization. Importantly, we do not aim to model the sharpening phase itself, but rather focus on the instability that follows it.

\paragraph{Warmup, divergence, and stability tradeoffs} 
The benefits and risks of instability have led to broader discussions about learning rate schedules. \citet{kalra2024warmup} and \citet{gilmer2022a_instabilities} recommend warmup strategies to improve initialization and reduce early divergence. While their goal is to minimize instability, our findings nuance this view: we show that instability can serve as a powerful implicit regularizer when models are robust to divergence.

Our empirical results suggest a tradeoff: instability should be avoided in the earliest, most fragile stages of training but deliberately embraced later, once the model is resilient. This perspective is supported by theoretical work from \citet{ahn2022unstableconvergence} and \citet{chen2024unstableGD}, who show that GD can still converge in unstable regimes.

\paragraph{Implicit biases in GD and SGD} 
Various implicit biases in optimization dynamics have been proposed. \citet{barrett2022implicitgradientregularization} describe a bias in GD that penalizes directions with high gradient magnitudes. Related to our work, \citet{xie2021diffusion} showed that SGD exponentially favors flat minima. Other works, such as \citet{xing2018walksgd} and \citet{smith2021implicitregsgd}, suggest that the path and averaging behavior of SGD play a role in its implicit regularization.

Our work contributes a new type of flatness bias through rotations, which arises specifically under instability and acts to rotate the descent direction away from sharp curvature directions. This is a novel mechanism that describes a unique dynamic geometric effect.

\paragraph{Sharpness as a generalization metric}
The relationship between sharpness and generalization remains debated. \citet{dinh2017sharpminima} criticized sharpness metrics for their lack of scale invariance, while \citet{kwon2021asam} proposed scale-invariant alternatives. \citet{kaur2023gen} found that standard optimizers may avoid sharpness pathologies, and \citet{andriushchenko2023modernsharpness} highlighted the dataset dependence of this relationship.

Despite these concerns, \citet{jiang2019generalizationmeasures} found that sharpness, while imperfect, remains one of the most predictive metrics of generalization. Our results align with this conclusion: we show that instability-induced flattening is associated with improved generalization, providing both a theoretical mechanism and empirical support for sharpness as a useful, though not infallible, measure of generalization performance. 

\paragraph{Full-batch GD in practice}
Finally, \citet{geiping2022fbgd} demonstrated that state-of-the-art results on CIFAR10 can be achieved without stochasticity, using full-batch gradient descent alone. Our results mirror this observation: by using large learning rates to induce instability, we achieve strong generalization in a full-batch setting, without explicit regularization. This confirms that training dynamics alone, specifically those involving instability and choice of learning rates, can be sufficient for effective generalization.

\newpage
\section{Technical Tools for Hessian Spectral Computations}\label{sec:phases:tech}

We build on existing techniques for Hessian analysis. Notably, we optimize the numerical stability of existing implementations of the Lanczos algorithm with the Modified Parlett-Kahan (MPK) re-orthoganalization. We introduce a metric for eigenvector comparison, extending the method from vector-based comparisons to subspace comparisons. 

\subsection{Stable Lanczos iteration via Modified Parlett-Kahan}\label{subsec:phases:tech:lanczos-mpk}

The Lanczos algorithm generates a Krylov basis for a real symmetric operator, such as the Hessian $\mH$, by short three-term recurrences. In exact arithmetic, the Lanczos vectors remain mutually orthogonal; in floating point, gradual loss of orthogonality degrades the tridiagonalization and the resulting spectral estimates. A common remedy is \emph{step-wise reorthogonalization} with a Gram-Schmidt procedure, which may suffer from significant numerical instability, especially when considering Hessians for deep neural networks which are frequently degenerate. 

In the study of numerically stable reorthogonalization, Parlett and Kahan (PK) proposed a practical rule, colloquially known as `twice is enough’. This rule monitors the reduction in norm after an orthogonalization sweep and, if necessary, performs a single reorthogonalization \cite{parlett1980communication}. Abaffy and Fodor adapted this idea, yielding a modified-PK (MPK) test with explicit accept/reorthogonalise/declare-dependent thresholds \cite{Abaffy20152TP}. 

Our first minor technical contribution leverages this connection to form \textbf{Lanzcos-MPK}: we use the same decision rule \emph{inside} Lanczos, which strongly improved the numerical stability of such algorithms. At each step we orthogonalize the candidate vector $\vw$ against the current basis $\mU_i=[\vu_0,\dots,\vu_i]$, compute:
\[
  \gamma \;\coloneq\; \frac{\|\vw_\perp\|_2}{\|\vw\|_2}, 
  \qquad \vw_\perp \;=\; \vw - \mU_i(\mU_i^\top \vw)
\]
and accept if $\gamma \geq \gamma_{\max}$; if not, we reorthogonalize once (`twice is enough’). If the second pass still yields $\gamma<\gamma_{\min}$, then we declare (near) linear dependence and \emph{restart}. In double precision, we use $\gamma_{\max}=\sqrt{1/2}$ by the PK analysis and a small floor $\gamma_{\min}\in[10^{-14},10^{-12}]$ depending on scaling. This adds negligible cost to the existing Lanczos steps. We used the default choice of $\gamma_\mathrm{max}=\sqrt{1/2}$ in \cite{Abaffy20152TP} and chose $\gamma_\mathrm{min}=10^{-14}$ to account for floating point errors. 

\subsection{Loss-basin similarities with cosine-Grassmannian distance}\label{subsec:phases:tech:grassman-cos}

Following \textcite{Ghorbani2019losslandscape}, our empirical studies revolve around \emph{loss basins}. As a secondary technical contribution, we develop a quantitative approximation of the similarity between such basins by comparing their informative (outlier) eigensubspaces. This is motivated by the observation that in practice, the remaining eigen-directions of deep networks are often flat or degenerate \parencite{granziol2020lr,papyan2019threelevel}.

Let $\mV^{*}_a \in \mathbb{R}^{p \times m_a}$ and $\mV^{*}_b \in \mathbb{R}^{p \times m_b}$ be column-orthonormal bases for the informative eigensubspaces of two models (or checkpoints). Define
$m_{\min} \coloneq \min(m_a,m_b)$. To compare the subspaces $\mathcal{S}_a = \mathrm{span}(\mV^{*}_a)$ and $\mathcal{S}_b = \mathrm{span}(\mV^{*}_b)$, we use the \emph{Grassmannian (geodesic) distance} on $\mathrm{Gr}(m_{\min},p)$ \parencite{ye2016schubert}. 

The principal angles $\{\phi_i\}_{i=1}^{m_{\min}} \subset [0,\tfrac{\pi}{2}]$ between $\mathcal{S}_a$ and $\mathcal{S}_b$ are defined via the singular value decomposition:
\[
  (\mV^{*}_a)^{\!\top} \mV^{*}_b
  \;=\; \mathcal{U}\,\Sigma\,\mathcal{V}^{\!\top}, 
  \qquad
  \Sigma = \mathrm{diag}(\sigma_1,\ldots,\sigma_{m_{\min}}),
  \qquad
  \cos(\phi_i) = \sigma_i \, .
\]

\noindent
The Grassmannian (geodesic) distance is:
\[
  h_{m_{\min}}(\mV^{*}_a,\mV^{*}_b)
  \;=\;
  \Biggl(\,\sum_{i=1}^{m_{\min}} \phi_i^2 \Biggr)^{\!1/2} \, .
\]

\noindent
To obtain a similarity measure in $[0,1]$ that reduces to cosine similarity in the one-dimensional case ($m_a=m_b=1$), we define:
\[
  \hat h_{m_{\min}}(\mV^{*}_a,\mV^{*}_b)
  \;=\;
  \cos \Biggl(\frac{h_{m_{\min}}(\mV^{*}_a,\mV^{*}_b)}{\sqrt{m_{\min}}}\Biggr) \, .
\]

\noindent
Finally, we summarize this with a \textbf{cosine-Grassmannian misalignment score}:
\begin{equation} \label{eqn:phases:Grassmanian} 
  S(\mV^*_a,\mV^*_b) \;=\; 1 - \hat h_{m_{\min}}(\mV^*_a,\mV^*_b), 
  \qquad S(\mV^*_a,\mV^*_b) \in [0,1] \, ,
\end{equation}
which increases as the subspaces diverge.

\begin{algorithm}[H]
\caption{\textbf{Lanczos-MPK}: Lanczos with MPK selective reorthogonalization and restarts}
\label{alg:phases:lanczos-mpk}
\begin{algorithmic}[1]
\Require matrix-vector oracle $G(\vu)=\mH\vu$ (via Pearlmutter's trick); dimension $n$; order $m$; Random Number Generator (RNG); $\gamma_{\max}=\sqrt{1/2}$; $\gamma_{\min}=10^{-14}$; $\texttt{max\_attempts} \coloneq z$;  
\Ensure $T\in\mathbb{R}^{m\times m}$ tridiagonal, $\mU=[\vu_0,\dots,\vu_{m-1}]\in\mathbb{R}^{n\times m}$ with orthonormal columns

\Function{normalize}{$\vx$} \State \Return $\vx/\|\vx\|_2$ 
\EndFunction

\Function{RandomVector}{$n,\texttt{RNG}$} \State \Return i.i.d.\ $N(0,1)$ from \texttt{RNG} \EndFunction

\\

\Function{MPK\_Orth}{$\vw,\, \mU$}
  \If{$\text{cols}(\mU)=0$} \Return $(\vw,\,1)$ \EndIf
  \State $\vw_0 \gets \vw$; \quad $\vc \gets \mU^\top \vw$; \quad $\vw_\perp \gets \vw - \mU \vc$
  \State $\gamma \gets \dfrac{\|\vw_\perp\|_2}{\|\vw_0\|_2}$ \Comment{$\triangleright$ PK test statistic}
  \State \Return $(\vw_\perp,\, \gamma)$
\EndFunction

\\

\Function{Lanczos\_MPK}{$\mathrm{oracle} \ G, n, m, \texttt{RNG}$}
  \For{$\texttt{attempt}=1$ \textbf{to} $z$}
    \State $\mT \gets 0_{m\times m}$; \quad $\mU \gets 0_{n\times m}$; \quad $\beta \gets 0$
\State $\vu_0 \gets \textproc{normalize}(\textproc{RandomVector}(n,\textproc{RNG}))$;    
\State $\mU[:,0] \gets \vu_0$
\For{$i=0$ \textbf{to} $m-1$}
      \State $\vu \gets \mU[:,i]$; \; $\vu_{\mathrm{prev}} \gets \begin{cases}0 & i=0\\ \mU[:,i-1] & i>0\end{cases}$
      \State $\vw \gets G(\vu)$ \Comment{$\triangleright$ using the matrix-vector product oracle}
      \State $\va \gets \langle \vw, \vu\rangle$; \; $\vw \gets \vw - \va \vu - \beta \vu_{\mathrm{prev}}$; \; $\mT[i,i]\gets \va$
      \State $\vw_{\perp},\, \gamma \gets \Call{MPK\_Orth}{\vw,\, \mU[:,0\!:\!i]}$ \label{line:mpk-first}
      \If{$\gamma < \gamma_{\max}$}
        \State $\vw_{\perp},\, \gamma \gets \Call{MPK\_Orth}{\vw_{\perp},\, \mU[:,0\!:\!i]}$ \Comment{$\triangleright$ PK: \emph{twice is enough}}
        \If{$\gamma < \gamma_{\min}$}
          \State \textbf{goto} \textsc{Restart} \Comment{$\triangleright$ declare dependence; restart the run}
        \EndIf
      \EndIf
      \State $\beta \gets \|\vw_{\perp}\|_2$
      \If{$i+1<m$}
        \State $\mT[i,i+1]\gets \beta$; \; $\mT[i+1,i]\gets \beta$
        \State $\mU[:,i+1]\gets \vw_{\perp}/\beta$
      \EndIf
    \EndFor
    \State \textbf{return} $(\mT,\mU)$
    \State \textbf{break}
    \Statex
    \State \textsc{Restart:} \textbf{continue} \Comment{$\triangleright$ new random start; preserves $m$ on success}
  \EndFor
  \\
  \State \textbf{error} \emph{Lanczos failed to build an order-$m$ basis after $z$ restarts.}
\EndFunction

\end{algorithmic}
\end{algorithm}

\section{\lamneg\ as a proxy for local curvature non-linearity}
\label{app:phases:lamneg-nonlin}

Following \parencite{granziol2020flatness, papyan2019threelevel, papyan2019spectrum}, we decompose the Hessian into a low-rank `signal' component $\mS$ (outliers, 
aligned with the \emph{generalized Gauss-Newton (GGN)} matrix) and a high-dimensional bulk component
$\mW$, commonly modeled with random matrix theory (RMT):
\[
    \mH = \mS + \mW \, .
\]
Under RMT, the spectrum of $\mW$ converges to the Wigner semicircle law with (Hessian) variance $\sigma_\epsilon^2$:
\[
\rho(\lambda) \;=\; \frac{1}{2\pi\sigma_\epsilon^2}
\sqrt{4\sigma_\epsilon^2 - \lambda^2}\ \, ,
\]
where $\rho(\lambda)$ is the density of eigenvalues.

The semicircle bulk has support:
\([\,\lambda_-, \lambda_+] = [-2\sigma_\epsilon,\, 2\sigma_\epsilon]\) \, .
Thus,
\[
\lambda_- = -2\sigma_\epsilon, 
\qquad \sigma_\epsilon^2 = \tfrac{1}{4}\lambda_-^2 \, .
\]
Hence, the negative edge $\lambda_-$ is a direct proxy for the bulk variance.

The quadratic Taylor model of the loss around parameters $\vtheta$ is:
\[
f(\vtheta + \vx) \;\approx\; f(\vtheta) + \vg^\top \vx + \tfrac{1}{2}\vx^\top \mH \vx \, ,
\]
where $\vg$ are the gradients and $\vx=x \vu$ is a small expansion in direction $\vu$. The next term involves the third derivative tensor $\mT$. Along a unit vector $\vu$, we write the tensor $\mT$:
\[
\mT[\vu,\vu,\vu] \;=\; \frac{d}{dx}\Big( \vu^\top \mH(\vtheta+x\vu)\,\vu \Big)\Big|_{x=0} \, .
\]

\noindent
Using the decomposition $\mH = \mS + \mW$, we can write:
\[
T[\vu,\vu,\vu] \;=\; 
\frac{d}{dx}\Big(\vu^\top \mS(\vtheta+x\vu)\,\vu\Big)\Big|_{x=0}
\;+\;
\frac{d}{dx}\Big(\vu^\top \mW(\vtheta+x\vu)\,\vu\Big)\Big|_{x=0} \, .
\]
The first term reflects changes in the low-rank signal $\mS$, which is dominated by the outliers. Empirically, these outlier directions (aligned with the GGN) are relatively stable across training and across minibatches, so their contribution to higher-order terms is small. In contrast, the bulk component $\mW$ is high-dimensional and fluctuates strongly: its variance scale $\sigma_\epsilon^2$ sets the typical size of curvature changes. Therefore, when discussing non-quadraticness, we can approximate:
\[
T[\vu,\vu,\vu] \;\approx\; 
\frac{d}{dx}\Big(\vu^\top \mW(\vtheta+t\vu)\,\vu\Big)\Big|_{x=0} \, ,
\]
so the bulk controls the third- and higher-order contributions.

The scale of third-order effects is therefore approximately proportional to $\sigma_\epsilon$. We define a dimensionless `non-quadraticness' index, $\mathcal{N}$, relative to the quadratic terms from the signal (outliers $\mS$):
\[
\mathcal{N}(\vu) \;\propto\; 
\frac{\mathrm{RMS}(\mT[\vu,\vu,\vu])}{\vu^\top \mS \vu}
\;\propto\; \frac{\sigma_\epsilon}{\vu^\top \mS \vu} \, ,
\]
where $\mathrm{RMS}$ denotes the root-mean-square magnitude. This construction makes $\mathcal{N}(\vu)$ dimensionless: it measures the scale of cubic fluctuations relative to the quadratic signal in the same direction. Substituting $\sigma_\epsilon = -\lambda_-/2$, this gives:
\[
\boxed{\;\;\mathcal{N}(\vu)\;\propto\;
\frac{-\lambda_-}{2\,\vu^\top \mS \vu}\;} \, ,
\]
which increases monotonically as $\lambda_-$ becomes more negative.
Note that non-quadraticness is a relative measure, so we divide by the scale of quadratic signal. $\lambda_-$ itself only controls the numerator. 

We deduce the following relationship:
\[
\lambda_- \downarrow
\;\;\Rightarrow\;\; 
\sigma_\epsilon^2 \uparrow
\;\;\Rightarrow\;\;
\text{higher-order influence}\uparrow
\;\;\Rightarrow\;\;
\text{local quadraticness}\downarrow \, .
\]
Thus, $\lambda_-$ can be seen as a proxy for non-quadratic behavior in the loss landscape. 
We note these results are not formal, but illustrate a qualitative relationship between $\lambda_-$ (the lower edge of the Wigner semicircle of $\rho(\lambda)$ for $\mW$ in this section) and the empirical \lamneg, which we measure in Section~\ref{ch:phases}.

\section{Additional Experiments for Section~\ref{ch:phases}}

\subsection{Suppressing updates along eigenvectors of leading eigenvalues}
\begin{figure}[H]
\centering
\includegraphics[width=0.9\linewidth]{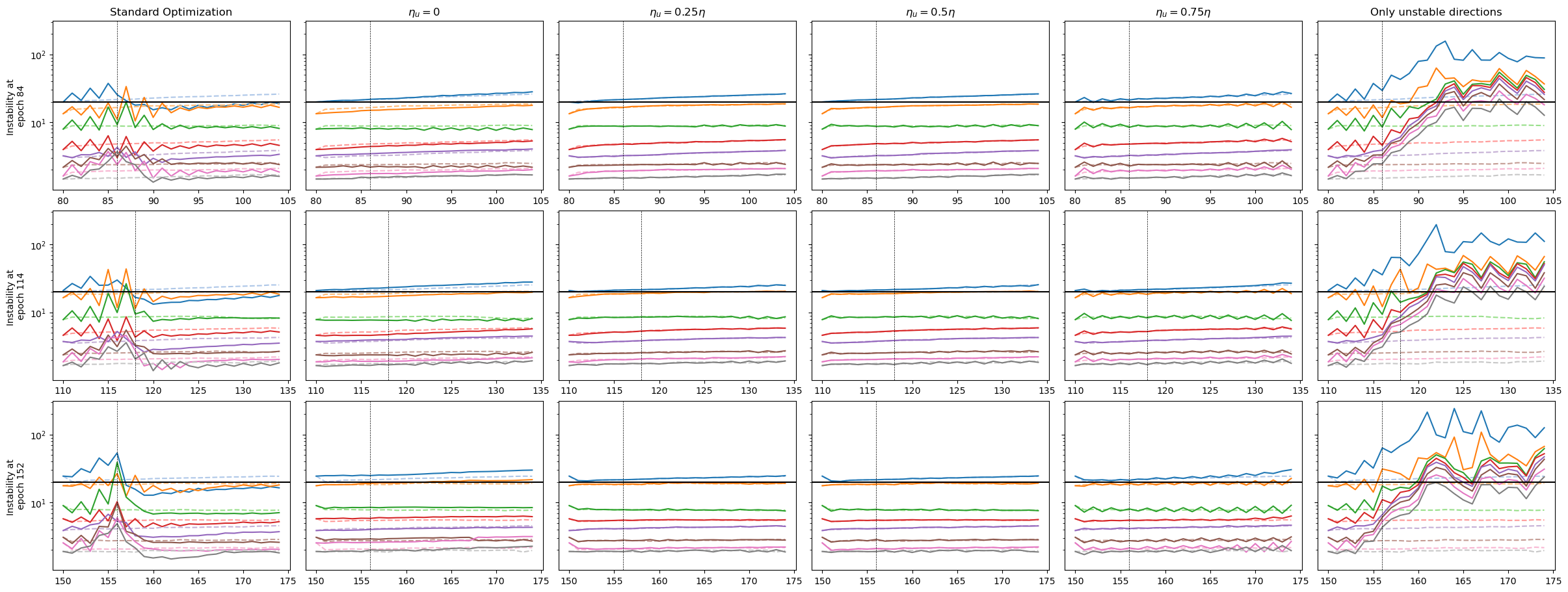}
\caption{\textbf{Effect of reducing learning rates along unstable directions.}
Suppressing updates along eigenvectors with $\eta\lambda_i>2$ stabilizes training and aligns $\lambda_{\max}$ with the gradient-flow baseline (dashed), whereas restricting updates to only unstable directions rapidly reintroduces divergence.}
\label{fig:phases:control-pc}
\end{figure}
\subsection{Complete Evolution of Figure \ref{fig:phases:movie}} \label{app:phases:movie}

\begin{figure}[H] 
\centering
\includegraphics[width=1.\linewidth]{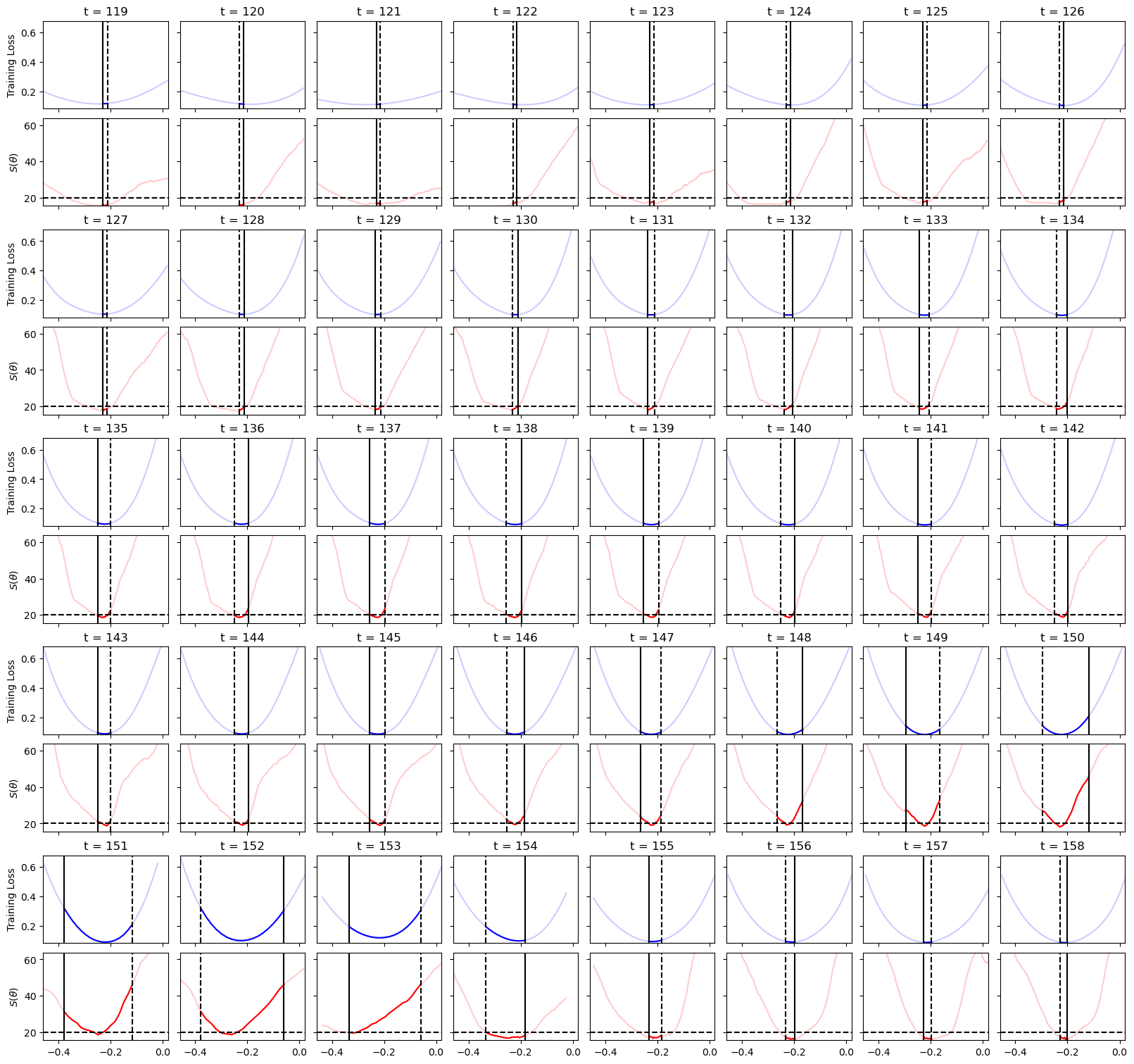}
\caption[Loss and \lammax\ curves through a training instability, all snapshots. ]{\textbf{Loss and \lammax\ curves through a training instability, all snapshots. } We plot the loss (top) and curvature (bottom, leading eigenvalue along the gradient direction, estimated with Hutchinson's trick \cite{hutchinson1989stochastic}) curves at snapshots through an instability. Note, the bottom is somewhat confusingly labeled as $S(\vtheta)$, which represents sharpness when $\vtheta$ is an argument of $S$. The radical changes in the curvature of sharpness suggests that the leading Hessian eigenvectors undergo significant changes, or `rotations', during instability. The dotted/solid vertical lines indicate the positions of previous/current parameters, respectively. Growth in $(\vtheta - \vtheta^*)\vv_i$ along the sharpest Hessian eigenvectors leads to exploration of the peripheries of the local minima, driving up $\mathcal{L}(\vtheta)$ and $S(\vtheta)$ in the process. As the instability develops, the $S(\vtheta)$ curve undergoes large changes until a flat region is found to enable a return to stability.}
\label{fig:phases:giant-movie}
\end{figure}

\newpage
\section{\label{app:rpe} Proofs, derivations, and additional experiments for Section~\ref{ch:rpe}: Rotational Polarity of Eigenvectors}

%

\subsection{Multilayer perceptrons as a sum-of-products} \label{app:rpe:sop}

We provide a derivation of the sum-of-products (SoP) formulation introduced in Subsection~\ref{sec:rpe:rot}, showing how a standard fully connected MLP reduces to a summation of products in the case of linear activations and a scalar output. This motivates our abstraction of the MLP as a sum of DLNs.

\paragraph{Full MLP form} Consider an MLP with depth $D$, weight matrices \( \mW_d \in \mathbb{R}^{h_d \times h_{d-1}} \), biases \( \vb_d \in \mathbb{R}^{h_d} \), and a uniform activation function \( \sigma_d = \sigma \). Let the input be \( \vx \in \mathbb{R}^N \) and assume a scalar output, i.e., $h_D = 1$. The MLP output is:
\[
    f(\vx) = \mW_D \sigma \left( \mW_{D-1} \sigma \left( \cdots \sigma(\mW_1 \vx + \vb_1) \cdots \right) + \vb_{D-1} \right) + \vb_D \, .
\]

\paragraph{Identity activations} In the case \( \sigma = \mI \), the MLP simplifies to:
\[    
    f(\vx) = \mW_D \mW_{D-1} \cdots \mW_1 \vx + \sum_{j=1}^{D} \left( \mW_D \cdots \mW_{j+1} \vb_j \right) + \vb_D  \,.
\]

\noindent
This expression decomposes into two components: A sum over products of weights along \emph{input paths}, and a sum over products of weights along \emph{bias paths}. 

We can write the function as:
\begin{align}
    f(\vx) = \underbrace{\sum_{i=1}^p \left( \prod_{\text{path through } x_i} w_{i, m_D \dots m_1} \right) x_i}_{\textit{input contributions}} 
    + \underbrace{\sum_{j=1}^D \sum_{k=1}^{h_j} \left( \prod_{\text{path through } b_{j,k}} w_{k, s_D \dots s_{j+1}} \right) b_{j,k}}_{\textit{bias contributions}} \, . \label{eqn:sumprod}
\end{align}

\noindent
Here, \( w_{i, m_D \dots m_1} \) is the product of scalar weights along a single path from $x_i$ to the output, and \( w_{k, s_D \dots s_{j+1}} \) is the product of weights along a path from the bias unit $b_{j,k}$ to the output. 

For each of the terms, we have:
\begin{align}
    w_{i, m_D \dots m_1} x_i &= \left( w_{D, m_D m_{D-1}} \cdots w_{1, m_1 m_i} \right) x_i \,,\nonumber \\
    w_{k, s_D \dots s_{j+1}} b_{j,k} &= \left( w_{D, s_D s_{D-1}} \cdots w_{j+1, s_{j+1} s_k} \right) b_{j,k} \,. \nonumber
\end{align}
\noindent
These products define \emph{computational paths} through the network.

\paragraph{Connection to DLNs} Each term in the expression above is a product of weights, i.e., a monomial in the parameters, optionally scaled by an input or bias. This aligns exactly with the form of a DLN:
\[
    f_{\mathrm{DLN}}(\vartheta_1, \dots, \vartheta_n) = \prod_{i=1}^n \vartheta_i \, ,
\]
where $\vartheta$ are the parameters of a DLN, each representing an input or a bias term in the MLP. The terms for input contributions are DLNs scaled by by a constant $x_i$, and terms for bias contributions correspond to constant DLNs (no input dependence). 

This formulation shows that an MLP with identity activations and scalar output computes a sum over monomials in the parameters, each associated with a specific path. In practice, MLPs share weights across paths, but for theoretical analysis we abstract this structure as a sum of independently (disjointly) parameterized DLNs. We extend our analysis to ReLU in Appendix~\ref{app:rpe:dln-relu}, and relax the disjoint assumption in Appendix~\ref{app:rpe:dln-shared}. 

%

\subsection{Derivations for Subsection~\ref{sec:rpe:rot:dln2}}\label{app:rpe:dln2}

\subsubsection{Closed form expression for $R$}\label{app:rpe:dln2:R}
We begin by computing the eigenvectors of the $2 \times 2$ Hessian matrix:
\[
\mH = 
\begin{bmatrix}
z'' \vartheta_2^2 & z'' \vartheta_1 \vartheta_2 + z' \\
z'' \vartheta_1 \vartheta_2 + z' & z'' \vartheta_1^2
\end{bmatrix} \, ,
\]
which arises from the loss $\mathcal{L}(\vartheta_1, \vartheta_2) = z(\vartheta_1 \vartheta_2)$. Let $\vv_1 = \begin{bmatrix} v_{1,1} \\ v_{1,2} \end{bmatrix}$ denote the eigenvector corresponding to the largest eigenvalue $\lambda_1$.

To quantify this alignment, we compute the absolute ratio of the eigenvector components:
\[
R_{n=2} := \left| \frac{v_{1,1}}{v_{1,2}} \right| \, .
\]

Without loss of generality, we normalize the eigenvector direction as $\vv_1 = \begin{bmatrix} R_{n=2} \\ 1 \end{bmatrix}$, yielding:
\[
\mH \begin{bmatrix} R_{n=2} \\ 1 \end{bmatrix} = \lambda_1 \begin{bmatrix} R_{n=2} \\ 1 \end{bmatrix} \, .
\]
Expanding both sides gives:
\begin{align*}
(z'' \vartheta_2^2) R_{n=2} + (z'' \vartheta_1 \vartheta_2 + z') &= \lambda_1 R_{n=2}, \\
(z'' \vartheta_1 \vartheta_2 + z') R_{n=2} + (z'' \vartheta_1^2) &= \lambda_1 \, .
\end{align*}
Eliminating $\lambda_1$ and solving for $R_{n=2}$ yields a quadratic:
\[
z''(\vartheta_2^2 - \vartheta_1^2) R_{n=2}^2 - 2(z'' \vartheta_1 \vartheta_2 + z') R_{n=2} - z''(\vartheta_2^2 - \vartheta_1^2) = 0 \, .
\]

\noindent
Solving this gives:
\[
R_{n=2} = \beta + \sqrt{\beta^2 + 1},
\qquad 
\beta := \frac{z''(\vartheta_2^2 - \vartheta_1^2)}{2(z' + z'' \vartheta_1 \vartheta_2)} \, .
\]

\paragraph{Sign invariance}  
Since replacing $\vartheta_1^2 - \vartheta_2^2$ by its negative in the quadratic reciprocates the eigenvector slope, one has:
\[
R_{n=2}(\phi_1,\phi_2)\; R_{n=2}(-\phi_1,\phi_2) = 1 \, ,
\]
where $\phi_1 = \vartheta_2^2 - \vartheta_1^2$ and $\phi_2 = 2(z' + z''\Theta)/z''$. Thus only the absolute value of the ratio matters. Taking $\beta = \bigl|\tfrac{\phi_1}{\phi_2}\bigr|$ recovers the sign–invariant form:
\[
R_{n=2} = \beta + \sqrt{\beta^2 + 1}, \qquad 
\beta = \left| \frac{z''(\vartheta_2^2 - \vartheta_1^2)}{2(z' + z'' \vartheta_1 \vartheta_2)} \right| \, .
\]

\noindent
This is the expression used in Subsection~\ref{sec:rpe:rot:dln2}.

\subsubsection{Approximate expression for $\gamma_\beta$} \label{app:rpe:dln2:gamma}
To study how $\beta$ evolves under gradient descent, recall the definition for its relative change:
\[
    \gamma_\beta := \frac{\beta + \Delta \beta}{\beta} = \frac{\beta_{t+1}}{\beta_t} \,.
\]
We now approximate this expression by expanding the GD update for one step. The parameters evolve via:
\[
\vartheta_i^{(t+1)} = \vartheta_i^{(t)} - \eta \cdot \frac{\partial L}{\partial \vartheta_i}, \quad \text{for } i = 1, 2 \,.
\]
Applying these updates gives:
\[
\begin{aligned}
\vartheta_1^{(t+1)} &= \vartheta_1 - \eta z' \vartheta_2 \, ,\\
\vartheta_2^{(t+1)} &= \vartheta_2 - \eta z' \vartheta_1 \, .
\end{aligned}
\]
Next, we expand the numerator and denominator of $\beta$ after one GD step:
\[
    \beta_{t+1} = \left| \frac{(\vartheta_2^{(t+1)})^2 - (\vartheta_1^{(t+1)})^2}{2 (\Theta_{t+1})} \right| \,.
\]
Let $\Delta x$ denote the update $x_{t+1} = x_{t} + \Delta x $, and $\gamma_x$ denote the ratio $\frac{x_{t+1}}{x_t}=\frac{x_t+\Delta x}{x_t}$. We can write the updates as:
\begin{align*}
    \Delta \vartheta_{1} &= \eta \frac{dz}{d\vartheta_1} = \eta z' \vartheta_2; \text{ similarly } \Delta \vartheta_2 =  \eta z' \vartheta_1 \, ,\\
    \gamma_{\phi_1} &= 1-\eta^2z'^2 \, , \\
    \phi_2 &= 2\left(\frac{z'}{z''}+\Theta\right) =
    \begin{cases}
    2\Theta ,& \text{when } x \rightarrow 0\\
    c\Theta,              & \text{when } x \rightarrow \infty, \text{where $2 < c \leq 4$ is a constant}
    \end{cases} \\
    \intertext{Given the constant scaling to $\Theta$ in both limits, we approximate the ratio of change $\gamma_{\phi_2}$ with $\gamma_{r_\Theta}$:}
    \implies \gamma_\beta &= \left| \frac{\gamma_{\phi_1}}{\gamma_{\phi_2}} \right| \approx \left| \frac{\gamma_{\phi_1}}{\gamma_\Theta} \right| = \left| \frac{1-\eta^2z'^2}{1-\zeta \eta z'+\eta^2z'^2} \right| \, .
\end{align*}
where, similar to the main text, we temporarily define \( \zeta := \left( \frac{\vartheta_1}{\vartheta_2} + \frac{\vartheta_2}{\vartheta_1} \right) \). This definition of $\zeta$ will not be used beyond this line. This recovers the form used in Subsection~\ref{sec:rpe:rot:dln2}, and isolates the learning-rate dependence of curvature alignment. The numerator reflects a second-order stabilizing effect, while the denominator captures first-order coupling between parameters.

\subsection{Settings and Outline of Theoretical Extension}\label{sec:rpe:rot:dln-ext}

Our analysis in Subsection \ref{sec:rpe:rot:dln2} focused on the $n=2$ setting for clarity and tractability. However, our theoretical framework can be general, as it extends to many settings where under milder conditions. Here, we provide a brief sketch of technical strategies to show that the behaviors of RPE persist under a number of theoretical generalizations. The detailed derivations are in Appendices~\ref{app:rpe:dln-n}-\ref{app:rpe:dln-shared}. 

\begin{enumerate}
    \item \textbf{Extension to arbitrary depth (general $n$-parameter DLNs)}
    
    In Appendix~\ref{app:rpe:dln-n}, we analyze a depth-$n$ DLN and show that the leading eigenvector of the Hessian continues to align with sharper parameters. The derivation follows the same strategy as in the $2$-parameter case: we apply the characteristic equation and perform a constrained optimization over \( \sum_j v_j / \vartheta_j \), subject to a unit-norm constraint. This yields the closed-form solution \( v_j \propto 1/\vartheta_j \), which matches the eigenvector observed in practice when accounting for additional constraints. Crucially, we find that sharper parameters consistently receive greater curvature weight \emph{above the stability threshold}, while exhibiting the opposite behavior below it, revealing a sign change in the rotation direction depending on the stability regime.

    \item \textbf{Extension to ReLU activation and CE loss via two-step dynamics}
    
    Appendix~\ref{app:rpe:dln-relu} establishes that ReLU activation functions are equivalent to sum-of-product (SoP) models in local regions. In this limit, when ReLU masks are considered fixed, the network output decomposes into a sum over multiplicative paths, each corresponding to a DLN term. This justifies modelling the loss curvature of ReLU using the SoP formulation, which inherits the same Hessian structure and eigenvector behavior as in the DLN case. Thus, ReLU activations do not disrupt the rotation phenomenon in local regions of parameter space. 
    
    ReLU activation is also asymmetrical about the minima, so the derivation instead considers the $2$-step dynamics, where the existence of many previous results are grounded through the \emph{Intermediate Value Theorem}. This same two-step treatment also applies to CE loss, whose asymmetry around the optimum necessitates a similar analysis. We empirically justify using the $2$-step dynamics in Figure~\ref{fig:rpe:period2}. 

    \item \textbf{Extension to multiple DLNs (general $m$)}

    Appendix~\ref{app:rpe:dln-m} generalizes the analysis to networks whose output is a sum of $m$ independently parameterized DLNs. The Hessian takes on a block structure with both intra- and cross-DLN curvature terms. When one DLN dominates the output magnitude (e.g., \( \Theta_a^2 \gg \Theta_b^2 \) for all $b \neq a$), we again use constrained optimization to derive the leading eigenvector, which aligns with sharper parameters of the dominant block. The analysis shows that curvature shifts toward weaker parameters as the system approaches the stability boundary, but rotates back toward sharper parameters above it. A rotation ratio $R_n(i,j)$ quantifies this shift, and the Davis-Kahan theorem provides formal bounds on eigenvector deviation due to inter-block perturbations.

    \item \textbf{Extension to shared-parameter DLNs}
    Appendix~\ref{app:rpe:dln-shared} generalizes the arguments to the case where DLNs overlap in parameters, as occurs in real MLPs. Instead of block-structured Hessians, parameter sharing produces dense curvature couplings, requiring a reformulation of the eigenproblem in scaled coordinates $\omega_i = v_i \vartheta_i$. This leads to a symmetric rescaled Hessian $\mB$, whose top eigenvector enforces a global balance across all parameters rather than the blockwise constancy seen in the disjoint case. Anisotropy arises because $B_{kj} \sim 1/\vartheta_j^2$, implying that small-scale (sharper) parameters exert greater curvature influence. Defining a generalized rotation ratio $R^2(i,j)$, the analysis shows that as scale differences grow, the leading eigenvector systematically shifts weight toward smaller-magnitude parameters. Thus, RPE persists under dense parameter sharing, with curvature alignment continuing to favor sharper directions despite the loss of exact constancy constraints
\end{enumerate}

\noindent
These settings relax of the conditions used in the simple case (Subsection~\ref{sec:rpe:rot:dln2}). The behaviors of RPE, specifically \textbf{opposite rotation} and \textbf{monotonicity}, are re-derived under these settings, which we find to be general.

%
\subsection{Eigenvector Dynamics in General-$n$ DLNs}
\label{app:rpe:dln-n}

We extend the analysis of eigenvector rotation from the 2-parameter DLN case to general $n$-parameter DLNs. 

\subsubsection*{Setting up the derivation}

Consider a DLN with parameters \( \vartheta_1, \ldots, \vartheta_n \in \mathbb{R} \) and scalar output:
\[
    \Theta = \prod_{i=1}^n \vartheta_i \, .
\]
The loss is given by a scalar function of this output:
\[
\mathcal{L}(\vartheta_1, \ldots, \vartheta_n) = z(\Theta) \, ,
\]
where \( z: \mathbb{R} \to \mathbb{R}_{\geq 0} \) is convex and twice differentiable, and minimized at \( \Theta = \Theta^* \in \mathbb{R} \). Without loss of generality, we evaluate all derivatives at the current $\Theta$, not assuming \( \Theta^* = 0 \). That is, define:
\[
    z' := \left. \frac{d z}{d \Theta} \right|_{\Theta}, \qquad z'' := \left. \frac{d^2 z}{d \Theta^2} \right|_{\Theta} \, .
\]
This general formulation preserves correctness of the eigenstructure and curvature analysis for any scalar choice of $\Theta^*$. 

By the chain rule, the gradient of the loss with respect to each parameter is:
\[
    \frac{\partial \mathcal{L}}{\partial \vartheta_i} = z' \cdot \frac{\partial \Theta}{\partial \vartheta_i} = z' \cdot \frac{\Theta}{\vartheta_i} \,.
\]
Taking second derivatives, we obtain the full Hessian:
\[
\frac{\partial^2 \mathcal{L}}{\partial \vartheta_i \partial \vartheta_j}
= \frac{\partial}{\partial \vartheta_j} \left( z' \cdot \frac{\Theta}{\vartheta_i} \right)
= z'' \cdot \frac{\Theta}{\vartheta_i} \cdot \frac{\Theta}{\vartheta_j} + z' \cdot \frac{\partial}{\partial \vartheta_j} \left( \frac{\Theta}{\vartheta_i} \right)  \,.
\]
Observe that:
\[
\frac{\partial}{\partial \vartheta_j} \left( \frac{\Theta}{\vartheta_i} \right) = 
\begin{cases}
\frac{\Theta}{\vartheta_j \vartheta_i}, & i \neq j \\
- \frac{\Theta}{\vartheta_i^2}, & i = j
\end{cases}
\]
Putting this together, the Hessian \( \mH \in \mathbb{R}^{n \times n} \) has entries:
\[
\mH_{ij} = z'' \cdot \frac{\Theta^2}{\vartheta_i \vartheta_j} 
+ z' \cdot \left( \delta_{ij} \cdot \left( - \frac{\Theta}{\vartheta_i^2} \right) + (1 - \delta_{ij}) \cdot \frac{\Theta} {\vartheta_i \vartheta_j} \right)  \,,
\]
or more compactly,
\[
\boxed{
\mH = z'' \cdot \Theta^2 \cdot \left( \frac{1}{\vartheta_i \vartheta_j} \right)_{ij}
\ +\ z' \cdot \left[ \left( \frac{\Theta }{\vartheta_i \vartheta_j} \right)_{i \neq j}
\ -\ \mathrm{diag} \left( \frac{\Theta}{\vartheta_i^2} \right) \right] \, .
}
\]
\noindent
In other words, the Hessian consists of a \textbf{rank-1 matrix} \( \propto \left( \frac{1}{\vartheta_i} \right)\left( \frac{1}{\vartheta_j} \right)^T \) scaled by $z'' \Theta^2$, which reflects curvature from second-order changes in $\Theta$, and a \textbf{diagonal correction} \( \propto -\frac{\Theta}{\vartheta_i^2} \) from the first derivative term $z'$, which modulates sharpness locally.

\subsubsection*{The structure of the Hessian} 

We now analyze the structure of the Hessian $\mH$ to identify its principal eigendirections. Recall from earlier that:
\[
    H_{jk} = D \cdot \vartheta_j^{-1} \vartheta_k^{-1} - z'\Theta \cdot \vartheta_j^{-2} \cdot \delta_{jk}, \quad \text{where } D = z'' \Theta^2 + z'\Theta \,.
\]
To identify eigenvectors, we consider a unit-norm vector \( \vv_i = [v_{i,1}, v_{i,2}, \dots, v_{i,n}]^\top \), and impose the eigenvalue condition:
\[
    \mH \vv_i = \lambda_i \vv_i \,.
\]
Taking the $k$-th coordinate of both sides yields:
\[
\lambda_i v_{i,k} = \sum_{j=1}^n H_{kj} v_{i,j} = D \vartheta_k^{-1} \sum_{j=1}^n \frac{v_{i,j}}{\vartheta_j} - z'\Theta \cdot v_{i,k} \vartheta_k^{-2} \,.
\]
Rearranging:
\[
\left(\lambda_i + \frac{z'\Theta}{\vartheta_k^2} \right) v_{i,k} = D \cdot \frac{1}{\vartheta_k} \sum_{j=1}^n \frac{v_{i,j}}{\vartheta_j} \,.
\]
Define:
\[
    C_i := D \sum_{j=1}^n \frac{v_{i,j}}{\vartheta_j} \, ,
\]
so we can rewrite the eigenvalue condition as:
\begin{equation}\label{eq:eigenvalue-condition-gen-n}
\left( \lambda_i + \frac{z'\Theta}{\vartheta_k^2} \right) v_{i,k} = \frac{C_i}{\vartheta_k} \,.
\end{equation}
This expresses each component of the eigenvector $\vv_i$ in terms of a shared constant $C_i$ and the local curvature scaling with $\vartheta_k$. Note that the rank-1 structure of $\mH$ implies that all components are coupled through $C_i$. The constancy of $C$ has also been verified in a toy $3$D example, as shown below in Figure~\ref{fig:rpe:constancy}.

\begin{figure}[h]
\centering
\includegraphics[width=0.7\textwidth]{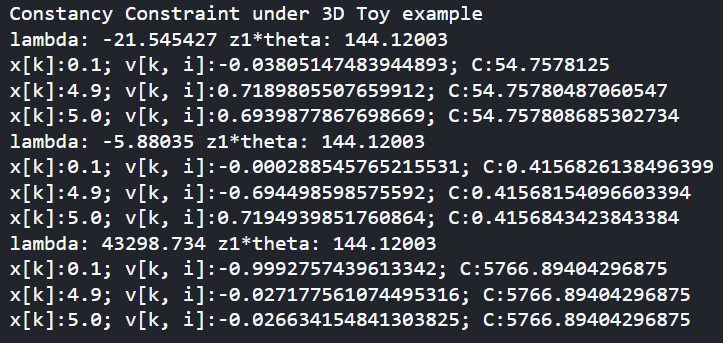}
\caption[The constancy constraint in a numerical simulation. ]{\textbf{The constancy constraint in a numerical simulation. } Numerical simulation results with $3$D DLN on quadratic loss. $C_i$s agree to $4$th decimal place. }
\label{fig:rpe:constancy}
\end{figure}

\subsubsection*{Maximizing the leading eigenvalue via $C_1$} 

We focus on the dominant eigenvalue $\lambda_1$ and corresponding eigenvector $\vv_1$. From the structure above, we observe that $\lambda_1$ increases monotonically with $C_1$. Therefore, to identify the sharpest eigenvector, we consider the following optimization:
\begin{align}
    \text{maximize} \quad & C_1 = D \sum_{j=1}^n \frac{v_{1,j}}{\vartheta_j} \label{eq:optim-c1} \\
    \text{subject to} \quad & \sum_{j=1}^n v_{1,j}^2 = 1 \tag*{(normalization constraint)}
\end{align}
We can absorb the constant $D$ and define the objective:
\[
    \max_{\vv_1} \sum_{j=1}^n \frac{v_{1,j}}{\vartheta_j} \quad \text{s.t. } \|\vv_1\|^2 = 1 \,.
\]
The solution to this is standard: the optimal $u^*_{1,j}$ is proportional to $\vartheta_j$. That is, there exists a constant $\psi$ such that:
\[
    u^*_{1,j} \vartheta_j = \psi = \frac{1}{\sqrt{\sum_k \vartheta_k^{-2}}} \,.
\]
This implies:
\[
    u^*_{1,j} = \frac{\psi}{\vartheta_j}, \quad \text{and } \sum_{j=1}^n \frac{u^*_{1,j}}{\vartheta_j} = \psi \sum_j \vartheta_j^{-2} = \psi^{-1} \,.
\]

\noindent
Under this solution, the eigenvector $\vv_1$ under this approximation points along a direction inversely scaled with parameter sharpness $\vartheta_j^{-1}$. Thus, directions aligned with smaller $\vartheta_j$ (i.e., sharper parameters) contribute more heavily to the curvature. We will later show that this structure approximately holds even under the full eigenvalue condition (as written in Equation~\ref{eq:eigenvalue-condition-gen-n}), making this solution a good initial estimate for the leading eigendirection.

\paragraph{Constancy constraint and deviation analysis}

While the approximate eigenvector $\vv_1^*$ derived in the previous section maximizes $C_1$ under a norm constraint, it does not fully satisfy the eigenvalue condition:
\[
    \left( \lambda_1 + \frac{z'\Theta}{\vartheta_k^2} \right) v_{1,k} = \frac{C_1}{\vartheta_k} \,.
\]
This condition must hold \emph{simultaneously for all $k$}, which imposes an additional constraint on our optimization:
\[
    \left( \lambda_1 + \frac{z'\Theta}{\vartheta_k^2} \right) v_{1,k} \vartheta_k = C_1, \quad \forall k \,.
\]
We refer to this as the \textbf{constancy constraint}, as it requires the scaled quantity on the left-hand side to remain constant across coordinates.

\paragraph{Approximate vs. exact solution} Let $u^*_{1,k}$ denote the optimizer from the previous section (without constancy constraints), and define:
\[
    \widehat{u}_{1,k} = u^*_{1,k} (1 + \epsilon_k) \,,
\]
where $\epsilon_k \in \mathbb{R}$ is a small deviation introduced to satisfy the full eigenvalue condition. We will now analyze the effect of these deviations.

\paragraph{Normalization revisited} Since the original solution $\vv_1^*$ is unit norm, the perturbed vector $\widehat{\vv}_1$ must also satisfy:
\[
    \sum_{k=1}^n \widehat{u}_{1,k}^2 = 1 \,.
\]
Substituting in the perturbed form:
\[
    \sum_{k=1}^n \left( u^*_{1,k} \right)^2 (1 + \epsilon_k)^2 = 1 \,.
\]
Expanding this gives:
\[
    \sum_k (u^{*}_{1,k})^2 + 2 \sum_k (u^{*}_{1,k})^2 \epsilon_k + \sum_k (u^{*}_{1,k})^2 \epsilon_k^2 = 1 \,.
\]
Let:
\[
    A := \sum_k (u^{*}_{1,k})^2 \epsilon_k, \qquad B := \sum_k (u^{*}_{1,k})^2 \epsilon_k^2 \,,
\]
so the normalization condition becomes:
\[
    1 + 2A + B = 1 \quad \Rightarrow \quad B = -2A \,.
\]

\paragraph{Bounding deviations} By the Cauchy-Schwarz inequality, we write:
\[
    A^2 \leq B \sum_k (u^{*}_{1,k})^2 = B \cdot 1 = B \,.
\]
Substituting $B = -2A$, we find:
\[
    A^2 \leq -2A \quad \Rightarrow \quad -2 \leq A \leq 0 \,.
\]
Hence, the total deviation is negative and bounded, with $A \to 0^-$ in well-conditioned cases. This validates that the approximate optimizer is close to a true eigenvector, especially when the constancy constraint is weakly violated.

\paragraph{Effect on $C_1$} Recalling from earlier that:
\[
    C_1 = D \sum_k \frac{\widehat{u}_{1,k}}{\vartheta_k} = D \sum_k \frac{u^*_{1,k}(1 + \epsilon_k)}{\vartheta_k} \,.
\]
We can write:
\[
    \sum_k \frac{u^*_{1,k}(1 + \epsilon_k)}{\vartheta_k} = \psi \sum_k \frac{1 + \epsilon_k}{\vartheta_k^2} = (1 + A) \cdot \psi^{-1} \,,
\]
where we used \( u^*_{1,k} = \psi / \vartheta_k \), and the fact that \( \sum_k \vartheta_k^{-2} = \psi^{-2} \). Thus, we can approximate:
\[
    C_1 \approx D (1 + A) \cdot \psi^{-1} \,.
\]

\noindent
The constancy constraint imposes small corrections $\epsilon_k$, which reduce $C_1$ slightly (since $A < 0$). Consequently, the leading eigenvalue $\lambda_1$ computed from the approximated $\vv_1^*$ slightly overestimates the true eigenvalue $\widehat{\lambda}_1$, but remains close. 

\subsubsection*{Quantifying Deviation via the Constancy Constraint}

To understand how the constancy constraint modifies the eigenvector structure, we apply it at two coordinates: one associated with the smallest parameter $\vartheta_1$ (typically sharpest), and one with the largest, $\vartheta_m \gg \vartheta_1$ (typically flattest). We write the constancy condition:
\[
    \left( \widehat{\lambda}_1 + \frac{z'\Theta}{\vartheta_k^2} \right) \widehat{u}_{1,k} \vartheta_k = \text{const}, \quad \forall k \,.
\]

\noindent
Substituting \( \widehat{u}_{1,k} = u^*_{1,k}(1 + \epsilon_k) = \frac{\psi}{\vartheta_k}(1 + \epsilon_k) \), we obtain:
\[
    \left( \widehat{\lambda}_1 + \frac{z'\Theta}{\vartheta_k^2} \right)(1 + \epsilon_k) \cdot \frac{\psi}{\vartheta_k} \cdot \vartheta_k = \left( \widehat{\lambda}_1 + \frac{z'\Theta}{\vartheta_k^2} \right)(1 + \epsilon_k)\psi = C_1 \, ,
\]
where \( C_1 = D(1 + A) \cdot \psi^{-1} \), as derived earlier.

Applying this equation to $k = 1$ and $k = m$ gives:
\[
    \left( \widehat{\lambda}_1 + \frac{z'\Theta}{\vartheta_1^2} \right)(1 + \epsilon_1)\psi = C_1 \, , \qquad
    \left( \widehat{\lambda}_1 + \frac{z'\Theta}{\vartheta_m^2} \right)(1 + \epsilon_m)\psi = C_1 \, .
\]
Since the right-hand sides are identical, we equate the two to get:
\[
    \left( \widehat{\lambda}_1 + \frac{z'\Theta}{\vartheta_1^2} \right)(1 + \epsilon_1) = \left( \widehat{\lambda}_1 + \frac{z'\Theta}{\vartheta_m^2} \right)(1 + \epsilon_m) \,.
\]
Rearranging:
\[
    \left( \widehat{\lambda}_1 + \frac{z'\Theta}{\vartheta_1^2} \right) = \left( \widehat{\lambda}_1 + \frac{z'\Theta}{\vartheta_m^2} \right)\cdot \frac{1 + \epsilon_m}{1 + \epsilon_1} \,.
\]

\paragraph{Ill-conditioning regime} In highly ill-conditioned DLNs, we assume that \( \vartheta_m^2 \gg \vartheta_1^2 \), so \( \frac{z'\Theta}{\vartheta_m^2} \ll \frac{z'\Theta}{\vartheta_1^2} \).  In this regime, the sharpest coordinate ($k=1$) carries the largest weight in the normalization sum, since $(u^*_{1,1})^2 \gg (u^*_{1,m})^2$. Any nonzero perturbation $\epsilon_1$ would therefore induce a disproportionately large violation of the unit norm condition. Furthermore, in the constancy equation, the factor $\widehat{\lambda}_1 + \tfrac{z'\Theta}{\vartheta_1^2}$ strongly amplifies $\epsilon_1$. For both reasons, the minimal--correction solution sets $\epsilon_1 \approx 0$, with the necessary adjustment absorbed instead in the flatter directions. 

Hence, we use $\frac{z'\Theta}{\vartheta^2_m}\approx 0$ and $1+\epsilon_1 \approx 1$ to get the ratio:
\[
\left( \widehat{\lambda}_1 + \frac{z'\Theta}{\vartheta_1^2} \right) \approx \widehat{\lambda}_1(1 + \epsilon_m) \,.
\]
This suggests:
\[
\epsilon_m \approx \frac{z'\Theta}{\vartheta_1^2 \widehat{\lambda}_1} \,.
\]

\noindent
We now compare deviation in $\lambda$ using prior results. Recall from earlier:
\[
    C_1 = D(1 + A)\psi^{-1} \approx \widehat{\lambda}_1(1 + \epsilon_m)\psi \,.
\]
Solving for $\widehat{\lambda}_1$:
\[
    \widehat{\lambda}_1(1 + \epsilon_m) = D(1 + A)\psi^{-2} \approx \lambda_1 \,.
\]
Therefore:
\[
\boxed{
    \widehat{\lambda}_1 = \lambda_1 (1 - \delta), \quad \text{where } \delta := \epsilon_m + o(\epsilon_m) > 0 \,.
} 
\]
Since $\epsilon_m > 0$, this confirms:
\[
    \lambda_1 > \widehat{\lambda}_1 > 0 \,,
\]
and by combining with our earlier bound, we get:
\[
    \epsilon_m \in \left(0,\ \frac{z'\Theta}{\vartheta_1^2 \widehat{\lambda}_1} \right) \,.
\]

\paragraph{Remark} We found previously that the constancy constraint slightly reduces the leading eigenvalue and perturbs the eigenvector. Here, we show that these deviations are controlled by the ill-conditioning (i.e., $\vartheta_m \gg \vartheta_1$) and the relative scale of $z'$ and $z''$. In practice, $\epsilon_k$ remains small and negative for sharp parameters, and positive but bounded for flat ones. 

\subsubsection*{Alignment Ratio and Sharpness Orientation}

We now derive the ratio of coordinates of the perturbed eigenvector $\widehat{\vv}_1$, focusing on the ratio:
\[
R_n \coloneqq \left| \frac{v_{1,1}}{v_{1,k}} \right|, \quad \text{for } k > 1 \,,
\]
where we use the subscript $n$ to denote the form of $R$ appropriate for general $n$. Using the constancy constraint:
\[
    \left( \widehat{\lambda}_1 + \frac{z'\Theta}{\vartheta_k^2} \right) \widehat{u}_{1,k} \vartheta_k = C_1 \,,
\]
we can write:
\[
    \widehat{u}_{1,k} = \frac{C_1}{\vartheta_k \left( \widehat{\lambda}_1 + \frac{z'\Theta}{\vartheta_k^2} \right)} \,.
\]
Similarly, for $j = 1$:
\[
    \widehat{u}_{1,1} = \frac{C_1}{\vartheta_1 \left( \widehat{\lambda}_1 + \frac{z'\Theta}{\vartheta_1^2} \right)} \,.
\]
Taking the ratio:
\[
    R_n = \left| \frac{\widehat{u}_{1,1}}{\widehat{u}_{1,k}} \right|
    = \frac{\left|\vartheta_k\right|}{\left|\vartheta_1\right|} \cdot
    \frac{\widehat{\lambda}_1 + \frac{z'\Theta}{\vartheta_k^2}}{\widehat{\lambda}_1 + \frac{z'\Theta}{\vartheta_1^2}} \,.
\]
Define:
\[
    f(\vartheta) := |\vartheta| \left( \widehat{\lambda}_1 + \frac{z'\Theta}{\vartheta^2} \right) \,,
\]
so the alignment ratio becomes:
\[
\boxed{
    R_n = \frac{f(\vartheta_k)}{f(\vartheta_1)} \, .
}
\]

\paragraph{Monotonicity} Consider the derivative of $f(\vartheta)$ for $\vartheta > 0$:
\[
    f'(\vartheta) = \widehat{\lambda}_1 - \frac{2z'\Theta}{\vartheta^3} \,.
\]
Define the critical value:
\[
    \vartheta_\mathrm{crit} := \left( \frac{2z'\Theta}{\widehat{\lambda}_1} \right)^{1/3} \,.
\]
Then, if \( \vartheta > \vartheta_\mathrm{crit} \), \( f'(\vartheta) > 0 \), so $f$ is increasing. Additionally, in ill-conditioned regimes (common in empirical DNNs), we often have \( \vartheta_k > \vartheta_\mathrm{crit} \ \forall k \), so $f(\vartheta)$ is \textbf{monotonically increasing} in $\vartheta$. This implies:
\[
    R_n = \frac{f(\vartheta_k)}{f(\vartheta_1)} \geq 1 \,,
\]
with equality only when \( \vartheta_k = \vartheta_1 \). Hence, the leading eigenvector tends to align more strongly with the \emph{sharpest} direction $\vartheta_1$, reinforcing the same structure of eigenvector rotation observed in the 2-parameter case in Section \ref{sec:rpe:rot:dln2}.

\paragraph{Alternative parametrization}
Let:
\[
    \vartheta_k^2 := r_k^2 \cdot \vartheta_1^2 \,,
\]
for some relative ratio $r_k \geq 1$. Then:
\[
    \Theta = \vartheta_1^n \prod_{k=1}^n r_k \,.
\]
and we get:
\[
    R_n = \frac{r_k}{1} \cdot \frac{\widehat{\lambda}_1 + \frac{z'\Theta}{r_k^2 \vartheta_1^2}}{\widehat{\lambda}_1 + \frac{z'\Theta}{\vartheta_1^2}} \,.
\]

\noindent
In this form, we again see that larger $r_k$ (i.e., flatter directions) increase the numerator, reinforcing the interpretation that:
\begin{enumerate}
    \item $R_n$ is \textbf{monotonically increasing in $r_k$}
    \item Stronger alignment to sharp directions (small $\vartheta_j$) occurs when parameters are highly anisotropic
    \item The function $f(\cdot)$ favors sharper directions in the leading curvature
\end{enumerate}

\noindent
In the general $n$ setting, the alignment ratio $R_n$ captures the relative weighting of eigenvector components under anisotropic parametrization. In strongly ill-conditioned models (typical in deep learning), this implies the leading eigendirection rotates toward unstable/sharp directions during stable phases, and may reorient during instability, which is a hallmark of the RPE mechanism.

\paragraph{Connection to 2-parameter case}
This analysis generalizes the rotation ratio $R_{n=2}$ introduced in the 2D setting. Recall:
\[
R_{n=2} = \left| \frac{v_{1, 1}}{v_{1, 2}} \right|
\]
In general-$n$, we interpret $R_n$ analogously: a large $R_n$ indicates strong alignment to the unstable/sharp parameter $\vartheta_1$, while a smaller value implies rotational movement away from it.


\subsection{Extension of DLN framework to ReLU activation}\label{app:rpe:dln-relu}

In this appendix, we extend our analysis of eigenvector rotation to networks with ReLU activations. Unlike the identity activation $\sigma(x) = x$ considered earlier, ReLU introduces two key challenges. First, the mapping $\sigma(x) = \max(0,x)$ is non-smooth at the origin, breaking the analytic symmetry assumptions underlying our previous arguments. Second, the loss landscape becomes asymmetric around the minimum, since perturbations that drive activations negative effectively flatten the curvature, while positive perturbations retain it. These features complicate the direct application of the smooth symmetric analysis developed above.

To handle this, we adopt a simple but principled strategy. We analyze the dynamics of gradient descent over \emph{two steps} rather than one. This period-2 process accounts for the asymmetry in the loss by treating the dynamics on both sides of the minimum jointly. Within this setup, we show that stable oscillations exist and that, under mild assumptions, the ReLU network behaves locally like a linear network with fixed activation patterns. In this local regime, the ReLU MLP reduces to a product-form structure analogous to the DLNs analyzed previously.

Our goal is to demonstrate that under these conditions, when the ReLU mask may remain constant, the phenomenon of RPE persists, and our theoretical RPE framework continues to apply.

\paragraph{Period-2 dynamics from loss asymmetry} 

ReLU networks introduce a notable complication absent in the previous setting: the loss landscape is no longer symmetric about the minimum. Specifically, due to the asymmetry of ReLU, the function $z(\Theta)$ induced by ReLU MLPs can exhibit different curvature and slope on either side of the minimum $\Theta^*$, even when the original loss is convex in the network output. This asymmetry breaks the assumptions typically used to analyze single-step GD updates in earlier sections.

To address this, we instead consider a two-step gradient descent process:
\[
    \Theta_{t+2} = \Theta_t - \eta z'(\Theta_t) - \eta z'(\Theta_{t+1}) \,,
\]
which allows us to treat the forward-and-return dynamics symmetrically, even when $z(\Theta)$ is not. The key insight is that even with nonsmooth or asymmetric losses, the \emph{sign structure} of updates is preserved across each half-cycle, once the stability threshold is established. This allows us to reason about equilibrium and stability using the \emph{composite update} over two steps. 

\paragraph{Existence of stable oscillations} 

We now demonstrate that under mild assumptions, a two-step GD process yields a dynamically stable periodic point. Let $z(\Theta)$ be a scalar loss function with the following properties (as before):
\begin{itemize}
    \item $z$ has a unique minimum at $\Theta = \Theta^*$
    \item $z$ is continuous and convex on $\Theta \neq \Theta^*$
    \item $z$ may be non-smooth or asymmetric at the minimum
\end{itemize}
These assumptions hold for many practical losses induced by ReLU networks near the linearization regime. Crucially, we make no assumption of symmetry.

Now, consider any initial $\Theta_t \neq \Theta^*$. Due to convexity, the sign of $z'(\Theta)$ matches the sign of $\Theta - \Theta^*$. Clearly, there exists large and small learning rates such that: 
\begin{itemize}
    \item There exists a \textbf{small learning rate} $\eta_s$, such that the update is conservative:
    \[
        \Theta_{t+2} = \Theta_t - \eta_s z'(\Theta_t) - \eta_s z'(\Theta_{t+1}) < \Theta_t \,.
    \]
    \item There exists a \textbf{large learning rate} $\eta_l$, such that the update overshoots:
    \[
        \Theta_{t+2} > \Theta_t \,.
    \]
\end{itemize}
The existence of these learning rates are non-controversial, consider $\eta_s = 0^+$ and $\eta_l = \infty$. By continuity of $z'$, the map $\Theta_{t+2}(\eta)$ varies smoothly with $\eta$, and we invoke the \emph{Intermediate Value Theorem} to conclude that there exists a value $\eta = \eta_\mathrm{eq}$ such that:
\[
    \Theta_{t+2} = \Theta_t \,,
\]
i.e., the trajectory returns to its starting point after two steps. This implies the existence of a stable \footnote{Here, we use `stable' to refer to a non-divergent period-$2$ cycle under fixed ReLU masks and learning rate. We do not claim convergence to this cycle.} \emph{period-$2$ cycle}. Near this equilibrium learning rate, the sign of pre-activations (and hence ReLU masks) can be expected to remain constant, which is a critical condition for the next step of our derivation.

\paragraph{ReLU MLPs as piecewise linear networks}

In the linear activation case ($\sigma = \mI$), the MLP reduces exactly to a composition of linear maps, and each computational path contributes a multiplicative term across depth. This is the foundation of our DLN abstraction. For ReLU activations, however, the function is no longer globally linear. Nevertheless, within a fixed region of input space (i.e. when the signs of all pre-activations remain unchanged), the ReLU MLP behaves exactly like a linear network with masked weights.

Let $\sigma(x) = \max(0, x)$, and consider an $n$-layer MLP with ReLU activations. For a fixed input and parameter configuration, each ReLU unit is either active or inactive. This induces a diagonal masking matrix $\mM_i \in \mathbb{R}^{h_i \times h_i}$ for each layer $i$, whose diagonal entries are either $1$ (active) or $0$ (inactive). These masks encode the local linearity of the ReLU network:
\[
    f(\vx) = \mW_n \mM_{n-1} \mW_{n-1} \mM_{n-2} \cdots \mM_1 \mW_1 \vx + \cdots \,,
\]
where biases are absorbed into the expression as affine terms. When all $\mM_i$ are fixed, i.e. when the sign pattern of the pre-activations remains unchanged over time, the ReLU MLP becomes \emph{effectively linear}. The network then behaves like a DLN under composition, with each layer contributing a linear multiplicative factor gated by the fixed mask.

\paragraph{RPE under fixed ReLU masks}

We now connect this linearized ReLU behavior to our earlier analysis of eigenvector rotation in DLNs. Suppose we are in a regime where the ReLU masks $\mM_i$ remain constant over time-this occurs, for example, when the learning rate is close to the equilibrium rate $\eta_\mathrm{eq}$ identified earlier. In this case, the ReLU MLP reduces to a deterministic linear function of its parameters, and each path through the network computes a product of (masked) weights.

This structure matches the form of our DLN model:
\[
    \Theta = \prod_{i=1}^n \vartheta_i 
\]
up to constant scaling factors from inputs and masks. Since the forward function is a sum over such products, and the loss depends only on the scalar output $\Theta$, the composite gradient dynamics match those analyzed in Sections~\ref{sec:rpe:rot:dln2} and~\ref{app:rpe:dln2}.

Consequently, all results regarding eigenvector rotation, including RPE, carry over to the ReLU case in the local region \emph{when the ReLU masks do not change}. This approximation becomes accurate when learning rates are chosen close to $\eta_\mathrm{eq}$, where pre-activations are near the fixed point and sign flips are unlikely. In this regime, ReLU networks locally inherit the curvature asymmetry and rotational behavior of DLNs.  

\paragraph{Quadratic loss and the stability threshold}

To ground the discussion in a concrete example, we now consider the special case of a quadratic loss: $z(\Theta) = \frac{1}{2} (\Theta - \Theta^*)^2$. This choice satisfies all assumptions from the previous sections and enables an exact expression for the equilibrium learning rate.

For this loss, the gradient and Hessian are:
\[
    z'(\Theta) = \Theta - \Theta^*, \qquad z''(\Theta) = 1 \,.
\]
As in earlier sections, we define the sharpness $S(\vartheta)$ as the maximum eigenvalue of the loss Hessian in the parameter space, evaluated at the current iterate. When the ReLU masks remain fixed, the network behaves linearly and $S(\vartheta)$ remains constant during the $2$-step cycle.

In this setting, we can solve explicitly for the learning rate $\eta_\mathrm{eq}$ that yields a period-$2$ trajectory:
\[
    \Theta_{t+2} = \Theta_t - \eta z'(\Theta_t) - \eta z'(\Theta_{t+1}) = \Theta_t \,.
\]
Solving this recurrence with $z'(\Theta) = \Theta - \Theta^*$ yields:
\[
    \eta_\mathrm{eq} = \frac{2}{S(\vartheta)} \,.
\]
This corresponds to the familiar stability threshold from the descent lemma for quadratic losses. While our earlier derivation of period-2 cycles made no assumption of differentiability at $\Theta^*$, this explicit expression provides intuition: when the learning rate approaches this critical threshold, the system enters a boundary regime in which oscillations neither grow nor decay.

\paragraph{Empirical illustration of periodic eigenvector rotations}

\begin{figure}[h] 
\includegraphics[width=\linewidth]{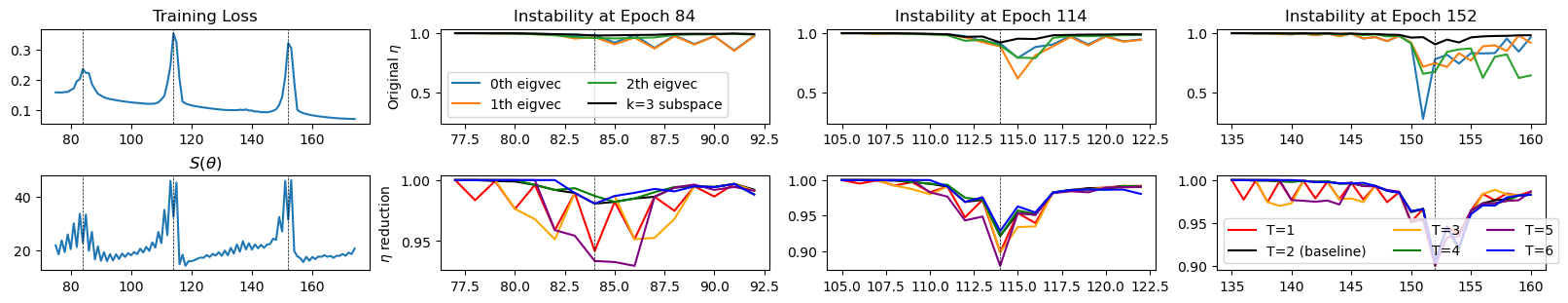}
\caption[$2$-step dynamics in an MLP.]{\textbf{$2$-step dynamics in an MLP.} We track the similarity of the sharpest Hessian eigenvectors across epochs through three instabilities. \textbf{Left:} $L(\vartheta)$ and $S(\vartheta)$. \textbf{Top:} similarities of the $k$-th eigenvectors (colored) and of subspaces formed by the top $3$ eigenvectors (black) during instabilities with period $T=2$. \textbf{Bottom:} similarity of subspaces formed by the top $3$ eigenvectors with varying period $T$s. During instabilities, the sharpest eigenvectors of the Hessian rotate away smoothly and monotonically (Top, same as Fig \ref{fig:rpe:instab_rot}. Smooth rotations only occur for even periods $T\%2=0$.}
\label{fig:rpe:period2}
\end{figure}

We now turn to empirical evidence for the presence of eigenvector rotations in ReLU MLPs trained near the stability threshold. In Figure~\ref{fig:rpe:period2}, we visualize the similarity of Hessian eigenvectors across epochs during training with ReLU activations. Smooth, monotonic rotations of the leading eigenvectors are observed during sharp changes in curvature, consistent with the predictions of RPE.

To probe the role of periodicity, we plot in Figure~\ref{fig:rpe:period2} the similarity of eigenvectors under different periodic sampling intervals. For periods $T = 2, 4, 6$, corresponding to even-step evaluations, the rotations are visible and smooth. In contrast, for odd periods $T = 3, 5$, the structure is lost. This supports our theoretical claim that eigenvector rotation emerges distinctly under period-2 dynamics - the same structure used in our extension to ReLU.

Together, these experiments suggest that the linearization approximation (via fixed ReLU masks) remains valid across a broad regime near the instability boundary. The range of learning rates for which RPE manifests is wider in practice than our strict theoretical assumptions might suggest.

\paragraph{Conclusion}

In this appendix, we extended the RPE framework to networks with ReLU activation functions. The key insight enabling this extension is that, under fixed ReLU masks, the network reduces to an effectively linear system-one that shares the multiplicative structure of DLNs. By analyzing the dynamics over a two-step period, we handled the asymmetry introduced by ReLU and demonstrated the existence of non-divergent periodic updates near an equilibrium learning rate.

While our theoretical claims are most rigorous under idealized conditions, specifically fixed activation masks and convex scalar losses, the empirical evidence suggests that the RPE effect persists across a much broader range of settings. In practice, we observe smooth eigenvector rotations over even-period cycles in ReLU MLPs trained near the instability threshold, consistent with our theory.

This provides evidence that the mechanism of RPE is not unique to identity-activated networks, but rather a consequence of the underlying multiplicative parametrization induced by depth. We expect similar behavior to hold for other piecewise linear activations (e.g., Leaky ReLU) and potentially broader classes of nonlinear objectives, especially in regimes where the network can be locally linearized over short time intervals.


\subsection{Extension to a Sum of DLNs (general $m$)} \label{app:rpe:dln-m}

In previous sections, we analyzed eigenvector rotation in a single DLN, where the scalar output takes the form $\Theta = \prod_{i=1}^n \vartheta_i$. However, this monomial structure is an idealization. In practical settings, such as fully connected ReLU networks, linearization reveals that outputs can be decomposed into sums of such monomials-each corresponding to an activation path through the network.

This motivates a more general framework in which the model output is a sum of depth-$n$ multiplicative paths:
\[
    f(x) = \sum_{j=1}^m \Theta_j, \qquad \text{where} \quad \Theta_j = \prod_{i=1}^{n_j} \vartheta_{i,j} \,,
\]
where the index $j \in \{1, 2, ... m\}$. This is the SoP formulation detailed in Appendix~\ref{app:rpe:sop}. 

To make the problem tractable and consistent with the earlier analysis, we consider the following in this section:
\begin{itemize}
    \item The input is scalar, so the model output is scalar as well
    \item Each term $\Theta_j$ is parameterized disjointly: no weight sharing across DLNs
    \item ReLU masks are fixed throughout training, so each path remains linear
\end{itemize}

This decomposition arises naturally under linearization of scalar-output ReLU MLPs. Our goal here is to extend the RPE framework to the general $m$-term setting, and in particular, to determine whether eigenvector rotation persists in this compositional structure. We first consider the two-term ($m=2$) case before generalizing to an arbitrary integer $m$.

\subsubsection*{Two-Term Additive DLN Model}

We begin with the simplest nontrivial case: a loss defined on two additive DLNs:
\[
    z(\Theta_a, \Theta_b) := (\Theta_a + \Theta_b)^q \,,
\]
where $q \in \mathbb{N}$ is even, and each $\Theta$ is a monomial in a disjoint set of parameters:
\[
    \Theta_a = \prod_{i=1}^{n_a} \vartheta_i, \qquad \Theta_b = \prod_{i=n_a+1}^{n_a+n_b} \vartheta_i \,,
\]
where we index the parameters with $i \in \{1, 2, ... n_a, n_a+1, ... n_a + n_b\}$. This setup corresponds to two independent depth-$m$ DLNs contributing to a shared scalar output.

We are interested in the curvature structure of the loss with respect to the full parameter vector $\vtheta = (\vartheta_1, \dots, \vartheta_n)$, and specifically in the leading eigenvector of the Hessian matrix:
\[
    \mH := \nabla^2_{\vtheta} z \,.
\]
We compute second derivatives of $z$ using the chain rule. Define:
\[
z'_a := \frac{\partial z}{\partial \Theta_a}, \quad z'_b := \frac{\partial z}{\partial \Theta_b}, \quad 
z''_a := \frac{\partial^2 z}{\partial \Theta_a^2}, \quad 
z''_b := \frac{\partial^2 z}{\partial \Theta_b^2}, \quad 
z''_{ab} := \frac{\partial^2 z}{\partial \Theta_a \partial \Theta_b} \,.
\]
The full Hessian $\mH \in \mathbb{R}^{n \times n}$ decomposes naturally into block form:
\begin{align}
    \mH &= \left[\begin{array}{c c c : c c c}
        z_a'' \Theta_a^2 \vartheta_1^{-2} & \cdots &  z_a'' \Theta_a^2 \vartheta_1^{-1}\vartheta_{n_a}^{-1} & z_{ab}'' \Theta_a \Theta_b \vartheta_1^{-1}\vartheta_{n_a+1}^{-1} & \cdots & z_{ab}'' \Theta_a \Theta_b \vartheta_1^{-1}\vartheta_{n_a+n_b}^{-1} \\
        \vdots & & \vdots & \vdots & & \vdots \\
        z_a'' \Theta_a^2 \vartheta_{n_a}^{-1}\vartheta_1^{-1} & \cdots &  z_a'' \Theta_a^2 \vartheta_{n_a}^{-2} & z_{ab}'' \Theta_a \Theta_b \vartheta_{n_a}^{-1}\vartheta_{n_a+1}^{-1} & \cdots & z_{ab}'' \Theta_a \Theta_b \vartheta_{n_a}^{-1}\vartheta_{n_a+n_b}^{-1} \\[4pt] \hdashline
        \\[-8pt]
        z_{ab}'' \Theta_b \Theta_a \vartheta_{n_a+1}^{-1} \vartheta_1^{-1} & \cdots &  z_{ab}'' \Theta_b \Theta_a \vartheta_{n_a+1}^{-1} \vartheta_{n_a}^{-1} & z_b'' \Theta_b^2 \vartheta_{n_a+1}^{-2} & \cdots & z_b'' \Theta_b^2 \vartheta_{n_a+1}^{-1} \vartheta_{n_a+n_b}^{-1} \\
        \vdots & & \vdots & \vdots & & \vdots \\
        z_{ab}'' \Theta_b \Theta_a \vartheta_{n_a+n_b}^{-1} \vartheta_1^{-1} & \cdots &  z_{ab}'' \Theta_b \Theta_a \vartheta_{n_a+n_b}^{-1} \vartheta_{n_a}^{-1} & z_b'' \Theta_b^2 \vartheta_{n_a+n_b}^{-1} \vartheta_{n_a+1}^{-1} & \cdots & z_b'' \Theta_b^2 \vartheta_{n_a+n_b}^{-2}
    \end{array}\right] \nonumber \\
    &= \begin{bmatrix}
        \mH(\Theta_a) & \mG(\Theta_a, \Theta_b) \\
        \mG(\Theta_b, \Theta_a) & \mH(\Theta_b)
    \end{bmatrix} \,, \label{eqn:Hess-block-form-clean}
\end{align}
where $\mH(\Theta_a) \in \mathbb{R}^{n_a \times n_a}$ encodes intra-group curvature within DLN-$a$; $\mH(\Theta_b) \in \mathbb{R}^{(n_b) \times (n_b)}$ does the same for DLN-$b$; and $\mG(\Theta_a, \Theta_b)$ contains second-order interactions between the two blocks.

Each block admits a simple parametric form due to the product structure of $\Theta_a$ and $\Theta_b$:
\begin{align*}
\frac{\partial^2 z}{\partial \vartheta_i \partial \vartheta_j} =
\begin{cases}
z''_a \cdot \frac{\Theta_a^2}{\vartheta_i \vartheta_j} + z'_a \cdot \frac{\Theta_a}{\vartheta_i^2} \delta_{ij}, & i,j \leq n_a, \\
z''_b \cdot \frac{\Theta_b^2}{\vartheta_i \vartheta_j} + z'_b \cdot \frac{\Theta_b}{\vartheta_i^2} \delta_{ij}, & i,j > n_a, \\
z''_{ab} \cdot \frac{\Theta_a \Theta_b}{\vartheta_i \vartheta_j}, & i \leq n_a,\, j > n_a \text{ or vice versa}.
\end{cases}
\end{align*}

\noindent
This decomposition reveals that the full curvature is driven by two competing terms: the curvature contributions from each \emph{individual} DLN and their \emph{cross-interactions}. Our analysis focuses on the scenario where one DLN dominates, e.g., $\Theta_a^2 \gg \Theta_b^2$, which simplifies the structure of the leading eigenvector and allows us to recover and extend the rotation behavior seen in the single-DLN case.

\subsubsection*{Constrained optimization for the leading eigenvector}

To study the top eigenvector of $\mH$, we begin from the eigenvalue equation:
\[
    \mH \vv_1 = \lambda_1 \vv_1 \,,
\]
where $\vv_1 = (v_{1,1}, \dots, v_{1,n})^\top$ is the leading eigenvector. 

From the block structure above, each coordinate $k$ in the eigenvalue equation can be expressed as:
\[
    \lambda_1 v_{1,k} = \sum_{j=1}^n H_{k,j} v_{1,j}, \quad \text{for all } k \,.
\]

To analyze this, we define the following constants that summarize local curvature and coupling terms:
\begin{align*}
    D_a := z''_a \Theta_a^2 + z'_a \Theta_a, \quad D_b := z''_b \Theta_b^2 + z'_b \Theta_b,  \quad D_{ab} := z''_{ab} \Theta_a \Theta_b \,.
\end{align*}

Now split the eigenvalue equation depending on whether $k \leq n_a$ (DLN-$a$) or $k > n_a$ (DLN-$b$). Then:
\[
\lambda_1 v_{1,k} = 
\begin{cases}
\displaystyle \sum_{j \leq n_a} D_a \cdot \frac{v_{1,j}}{\vartheta_j \vartheta_k}
+ \sum_{j > n_a} D_{ab} \cdot \frac{v_{1,j}}{\vartheta_j \vartheta_k}
- z'_a \cdot \frac{\Theta_a}{\vartheta_k^2} v_{1,k}, & k \leq n_a \\[1em]
\displaystyle \sum_{j > n_a} D_b \cdot \frac{v_{1,j}}{\vartheta_j \vartheta_k}
+ \sum_{j \leq n_a} D_{ab} \cdot \frac{v_{1,j}}{\vartheta_j \vartheta_k}
- z'_b \cdot \frac{\Theta_b}{\vartheta_k^2} v_{1,k}, & k > n_a
\end{cases}
\]

We now define a constant term $C_1$ summarizing the right-hand side of the eigenvector equation:
\[
C_1 := \sum_{j=1}^n E_j \cdot \frac{v_{1,j}}{\vartheta_j}, \quad \text{where } 
E_j := 
\begin{cases}
D_a, & j \leq n_a \\
D_{ab}, & j > n_a
\end{cases}
\quad \text{if } k \leq n_a \,.
\]
and similarly interchanging subscripts $a \leftrightarrow b$ if $k > n_a$.

This leads to the following optimization problem: find the unit vector $\vv_1$ that maximizes the right-hand side of the eigenvalue equation under a normalization constraint. That is,
\[
\text{maximize} \quad C_1 = \sum_{j=1}^n \frac{v_{1,j}}{\vartheta_j} \quad 
\text{subject to} \quad \sum_{j=1}^n v_{1,j}^2 = 1 \,.
\]
Using the method of Lagrange multipliers, the optimizer is:
\[
v_{1,j}^* \propto \frac{1}{\vartheta_j}, \quad \text{normalized as } \quad 
v_{1,j}^* = \frac{1}{\vartheta_j} \cdot \psi, \quad 
\psi := \left( \sum_{j=1}^n \vartheta_j^{-2} \right)^{-1/2} \,.
\]

\noindent
This eigenvector generalizes the single-DLN solution and captures the dominant direction of curvature when one DLN contributes most of the gradient magnitude.

\subsubsection*{Deviation and Rotation Ratio}

We now consider small deviations from the ideal eigenvector solution. Suppose:
\[
    v_{1,j} = (1 + \epsilon_j) v_{1,j}^*, \quad \text{with } |\epsilon_j| \ll 1 \,.
\]
Define a perturbation scalar:
\[
    A := \sum_{j=1}^n (v_{1,j}^*)^2 \epsilon_j \,.
\]
Substituting into the eigenvalue equation, and grouping common terms, we obtain the approximate consistency condition:
\[
\left( \lambda_1 + \frac{z'_a \cdot \Theta_a \cdot \delta(j \leq m) + z'_b \cdot \Theta_b \cdot \delta(j > m)}{\vartheta_j^2} \right)
(1 + \epsilon_j) v_{1,j}^* \vartheta_j 
\approx E(1 + A)\psi^{-1} 
\]
for each $j$, where $E$ is the local effective curvature constant as before.

To isolate the relative contributions of different parameters, we define a local functional:
\[
f_y(\vartheta) := |\vartheta| \lambda_1 + \frac{\partial z}{\partial y} \cdot \frac{y}{\vartheta^2}, \quad 
\text{for } y \in \{ \Theta_a, \Theta_b \} \,.
\]

Then for two coordinates $i$ and $j$ (possibly from different DLNs), the \emph{rotation ratio} $R$ is defined as:
\[
    R_{m=2}(i, j) := \frac{f_{\Theta_b}(\vartheta_j)}{f_{\Theta_a}(\vartheta_i)} \,,
\]
where we use the subscript $m=2$ for $R(i, j)$ to denote its form in this setting. 

This expression quantifies the relative curvature-induced rotation strength of $\vartheta_j$ (from a weaker DLN) compared to $\vartheta_i$ (from the dominant DLN). A key feature of $R_{m=2}(i,j)$ is that:
\[
    \frac{d R_{m=2}(i,j)}{d r_{ij}} > 0 \quad \text{where } \vartheta_j = r_{ij} \cdot \vartheta_i \,,
\]
which means that the ratio increases as $\vartheta_j$ grows relative to $\vartheta_i$.

Moreover, $\frac{d R_{m=2}(i,j)}{d \vartheta_j} = 0$ when holding the ratio $r_{ij}$ fixed, indicating that the absolute magnitude of $\vartheta_j$ is not the driving factor. Only its scale relative to other parameters matters.

This completes the perturbative picture for interactions between two DLNs when indices $i$, $j$ belong to separate DLNs, and we refer to the derivations in Appendix~\ref{app:rpe:dln-n} under general $n$, $m=1$, for interactions of intra-DLN parameters. These derivations reveal a consistent picture - even when the eigenvector deviates slightly from optimality, its shape continues to reflect local scale-sensitive curvature, consistent with the claims of RPE. 

\subsubsection*{Davis-Kahan as a Supporting Tool}

While our main analysis derives the structure of the leading eigenvector via a constrained optimization framework, it is important to understand why this approximation remains valid even in the presence of off-block (i.e., cross-DLN) curvature.

To that end, we use the \emph{Davis-Kahan} theorem as a supporting tool \parencite{davis1970rotation,yu2015usefulDK}. Suppose that one DLN, say DLN $a$, dominates the others in its contribution to the output, such that
\[
    \Theta_a^2 \gg \Theta_b^2 \qquad \text{for all } b \neq a \, .
\]
Then the block $\mH^{(a,a)}$ dominates the Hessian $\mH$, which can be written as
\[
\mH 
= \underbrace{\mH^{(a,a)} \oplus \mathbf{0}}_{\text{block-diagonal approximation}} 
+ \underbrace{\Delta}_{\text{cross-DLN perturbations}} \, ,
\]
where $\oplus$ denotes the direct sum embedding of $\mH^{(a,a)}$ into the full parameter space, and $\Delta$ collects all off-diagonal interaction blocks.

Let $\vv_0^{(a)}$ be the normalized eigenvector corresponding to the largest eigenvalue $\lambda_1$ of the isolated block $\mH^{(a,a)}$. Then the Davis-Kahan theorem provides a bound on the deviation between the leading eigenvector $\vv$ of the full Hessian $\mH$ and the block-wise idealization $\vv_0^{(a)}$:
\[
\sin \angle\!\bigl(\vv, \vv_0^{(a)}\bigr) 
\;\leq\; \frac{\| \Delta \vv_0^{(a)} \|_2}{\lambda_1 - \lambda_2} \, ,
\]
where $\lambda_2$ is the next-largest eigenvalue of $\mH$. The numerator quantifies the distortion induced by cross-block curvature, while the denominator captures the spectral gap.

Under the imbalance condition above, the off-block norms satisfy
\[
\| \Delta \vv_0^{(a)} \|_2 = \mathcal{O}\!\Bigl(\Theta_a \cdot \max_{b\neq a}\Theta_b\Bigr)
\qquad\text{whereas}\qquad
\lambda_1 = \Theta_a^2 \cdot \Omega(1) \, 
\]
where we use the asymptotic notation $\mathcal{O}(\cdot)$ to indicate an upper bound up to a constant factor, and $\Omega(\cdot)$ to indicate a lower bound up to a constant factor.  

Hence $\| \Delta \vv_0^{(a)} \|_2 \ll \lambda_1$, so the deviation angle is small, and the eigenvector of the full Hessian is well approximated by that of the dominant block.

Thus, the Davis-Kahan bound provides a formal justification for treating the constrained optimizer derived from the dominant block as a valid proxy for the leading eigenvector of the full Hessian.

\subsubsection*{generalization to $m$-Term Additive DLNs}

We now extend the rotation analysis from the $m=2$ case to the general case where the model output is a sum of general-$m$ disjoint DLNs:
\[
    f(x) = \sum_{a=1}^m \Theta_a, \qquad \text{with } \Theta_a = \prod_{i=1}^n \vartheta_{i,a} \,.
\]
Each $\Theta_a$ corresponds to an independently parameterized depth-$n_a \leq n$ DLN term. The scalar loss is taken to be $z(f(x)) = f(x)^q$, where $q$ is an even integer.

Our goal is to characterize the top eigenvector $\vv_1$ of the full Hessian $\mH \in \mathbb{R}^{mn \times mn}$ with respect to the full parameter vector $\vtheta = \{ \vartheta_{i,a} \}_{i=1,\dots,n}^{a=1,\dots,m}$.

\vspace{0.5em}
\noindent The derivation follows the $m=2$ case closely. We outline the following steps:

\begin{enumerate}
    \item \textbf{Hessian Structure.}  
    The Hessian $\mH$ inherits a block structure with intra- and inter-DLN curvature terms:
    \[
    \mH = \begin{bmatrix}
        \mH^{(1,1)} & \mG^{(1,2)} & \cdots & \mG^{(1,m)} \\
        \mG^{(2,1)} & \mH^{(2,2)} & \cdots & \mG^{(2,m)} \\
        \vdots & \vdots & \ddots & \vdots \\
        \mG^{(m,1)} & \mG^{(m,2)} & \cdots & \mH^{(m,m)}
    \end{bmatrix} \,.
    \]
    For $a \neq b$, the mixed blocks are:
    \[
    \mG^{(a,b)}_{ij} = \frac{\partial^2 z}{\partial \Theta_a \partial \Theta_b} \cdot \frac{\Theta_a \Theta_b}{\vartheta_{i,a} \vartheta_{j,b}} \,,
    \]
    while diagonal blocks include an additional first-derivative term:
    \[
    \mH^{(a,a)}_{ij} = \frac{\partial^2 z}{\partial \Theta_a^2} \cdot \frac{\Theta_a^2}{\vartheta_{i,a} \vartheta_{j,a}} + \delta_{ij} \cdot \frac{\partial z}{\partial \Theta_a} \cdot \frac{\Theta_a}{\vartheta_{i,a}^2} \,.
    \]

    \item \textbf{Characteristic Equation.}  
    Let $\vv_1$ denote the leading eigenvector and $\lambda_1$ the top eigenvalue of $\mH$. For each parameter $\vartheta_{k}$ in DLN $\Theta_a$, we write:
    \[
    \lambda_1 v_{1,k} = \sum_{j} H_{k,j} v_{1,j} \,,
    \]
    which we expand by grouping the sum over DLN blocks. We define constants:
    \[
    D_a := \frac{\partial^2 z}{\partial \Theta_a^2} \Theta_a^2 + \frac{\partial z}{\partial \Theta_a} \Theta_a, \qquad
    D_{ab} := \frac{\partial^2 z}{\partial \Theta_a \partial \Theta_b} \Theta_a \Theta_b \,,
    \]
    so that the characteristic equation simplifies into a weighted sum:
    \[
    \left( \lambda_1 + \frac{\partial z}{\partial \Theta_a} \cdot \frac{1}{\vartheta_{k}^2} \right) \cdot v_{1,k} \cdot \vartheta_k = 
    \left( D_a + \sum_{b \neq a} D_{ab} \right) \cdot \sum_j \frac{v_{1,j}}{\vartheta_j} \,.
    \]

    \item \textbf{Constrained optimization.}  
    To satisfy this uniformity condition across all coordinates $k$, we define:
    \[
    C := \sum_j \frac{v_{1,j}}{\vartheta_j} \,.
    \]
    We then solve the optimization:
    \[
    \text{maximize} \quad C = \sum_j \frac{v_{1,j}}{\vartheta_j} \qquad \text{subject to} \quad \sum_j v_{1,j}^2 = 1 \,.
    \]
    The optimal solution is:
    \[
    v_{1,j}^* = \frac{1}{\vartheta_j} \cdot \left( \sum_{k} \vartheta_k^{-2} \right)^{-\frac{1}{2}} \,.
    \]
    This generalizes the result from the 2-DLN case: the leading eigenvector of the full Hessian is aligned with inverse-scaled parameters.

    \item \textbf{Deviation Analysis and Rotation Ratio.}  
    Suppose the actual eigenvector deviates slightly from this ideal solution:
    \[
    v_{1,j} = (1 + \epsilon_j) \cdot v_{1,j}^* \,.
    \]
    Define $A := \sum_j (v_{1,j}^*)^2 \cdot \epsilon_j$ to express average deviation. Plugging into the characteristic equation, we obtain for each $j$:
    \[
    \left( \lambda_1 + \frac{\partial z}{\partial \Theta_a} \cdot \frac{1}{\vartheta_j^2} \right) (1 + \epsilon_j) v_{1,j}^* \vartheta_j = (1 + A) \cdot C_{\mathrm{eff}} \,.
    \]

    \item \textbf{Rotation Function and Monotonicity.}  
    Define the DLN-dependent rotation function:
    \[
    f_{\Theta_a}(\vartheta_j) := |\vartheta_j| \lambda_1 + \frac{\partial z}{\partial \Theta_a} \cdot \frac{1}{\vartheta_j^2} \,,
    \]
    and rotation ratio for any two coordinates $i,j$:
    \[
    R(i,j) := \frac{f_{\Theta_b}(\vartheta_j)}{f_{\Theta_a}(\vartheta_i)} \,.
    \]
    If $\vartheta_j = r_{ij} \cdot \vartheta_i$, we compute:
    \[
    \frac{d R(i,j)}{d r_{ij}} > 0, \qquad \frac{d R(i,j)}{d \vartheta_j} = 0 \,.
    \]
    These show that rotation strength increases with relative scale but is invariant to absolute scale.
\end{enumerate}

From the above, we see that the same monotonicity result holds as in the 2-DLN case. Regardless of DLN structure, the eigenvector rotation is driven by relative scale, curvature, and local optimization geometry. Hence, RPE extends robustly to general $m$.

The expressions for $f_{\Theta_a}$ and $R(i,j)$ depend only on local parameter scales and their DLN curvature constants. Hence, the rotation effect persists irrespective of whether $\vartheta_i$ and $\vartheta_j$ belong to the same DLN. The mechanism is symmetric across DLNs.
    
A similar strategy to justify that the constrained-optimization eigenvector remains close to the true eigenvector can be built using the Davis-Kahan \cite{davis1970rotation} theorem. 


\subsection{Extension to DLNs with shared parameters} \label{app:rpe:dln-shared}

In previous sections, we studied the rotational behavior of DLN eigenvectors under the assumption that parameters are disjoint across DLN terms. This led to block-structured Hessians and enabled a clean derivation of a constancy constraint on the leading Hessian eigenvector, from which RPE behavior followed. In this appendix, we generalize this analysis to the more realistic setting where parameters may be shared across multiple DLNs, as is typical in MLPs. In this case, the membership of each parameter $\vartheta$ to each DLN, which represents a \emph{computation path} of an MLP, may be dense. 

Let each DLN term be defined as a product over a subset of the parameters:
\[
    \Theta_a = \prod_{i \in \mathcal{I}_a} \vartheta_i, \quad a \in \{1, 2, \dots, m\} \,,
\]
where $\mathcal{I}_a \subseteq \{1, 2, \dots, n\}$ is the index set of parameters participating in $\Theta_a$. Importantly, we now allow the sets $\mathcal{I}_a$ to overlap arbitrarily, i.e., each parameter $\vartheta_i$ may appear in multiple DLNs.

We consider the model output as the sum of these DLN terms:
\[
    f(\vtheta) = \sum_{a=1}^m \Theta_a \,,
\]
and study a smooth scalar loss function $L(\vtheta) = z(\Theta_1, \dots, \Theta_m)$, where $z: \mathbb{R}^m \to \mathbb{R}$ is assumed to be differentiable to second order.

Our goal is to analyze the structure of the Hessian $\mH = \nabla^2 L(\vtheta)$ and understand the behavior of its top eigenvector $\vv_1$ under parameter scaling. In particular, we define the transformed eigenvector components:
\[
    \omega_i \coloneqq v_i \cdot \vartheta_i \,,
\]
as a natural generalization of the scaling-based alignment used in previous RPE derivations. Our previous definitions used the ratios between coordinates to measure rotation. Here, we will use this formulation of $\omega$ to define a more general rotation measure and show that RPE behavior persists under dense DLN overlap.

\subsubsection*{Eigenproblem in Scaled Coordinates}

We now analyze the top eigendirection of the Hessian in the shared parameter setting. Recall the scalar loss function $L(\vtheta) = z(\Theta_1, \dots, \Theta_m)$, where each $\Theta_a$ is a multiplicative DLN term. Unlike earlier sections, we allow the index sets $\mathcal{I}_a$ to overlap arbitrarily: parameters may participate in multiple DLNs, leading to a dense coupling in the Hessian.

We are interested in the leading eigenpair $(\lambda_1, \vv_1)$ of the Hessian:
\[
    \mH \vv_1 = \lambda_1 \vv_1 \,,
\]
and in particular, how the structure of $\vv_1$ depends on parameter scale.

Substituting $\omega$ into the eigenproblem, we have for each index $k$:
\[
    \lambda_1 \psi_k = \vartheta_k \cdot \sum_{j=1}^n H_{kj} \cdot v_j = \sum_{j=1}^n \left( \vartheta_k H_{kj} / \vartheta_j \right) \cdot \omega_j \,.
\]
This motivates defining a new matrix $\mB \in \mathbb{R}^{n \times n}$ with entries:
\[
    B_{kj} := \frac{\vartheta_k}{\vartheta_j} H_{kj} \,,
\]
so that the scaled eigenproblem becomes:
\[
    \mB \vomega = \lambda_1 \vomega \,.
\]
This is a standard linear eigenproblem for $\vomega$, but the matrix $\mB$ now encodes both the second-order curvature and the scaling relationships between parameters.

We note that the structure of $\mB$ has two key properties. Firstly, $\mB$ is symmetric because $\mH$ is symmetric and the scaling preserves symmetry: $B_{kj} = B_{jk}$. Additionally, the entries $B_{kj}$ inherit curvature coupling from all DLNs in which both $\vartheta_k$ and $\vartheta_j$ participate. These terms are weighted by the inverse scales $1 / (\vartheta_k \vartheta_j)$, due to the multiplicative structure of each DLN.

This eigenproblem provides a natural generalization of the constancy condition derived in the disjoint case. Instead of exact constancy of $\psi_i$ within disjoint blocks, we now obtain a global balance constraint across all $\psi_i$, implicitly defined by the eigensystem $\mB \vomega = \lambda_1 \vomega$. The solution to this system encodes how the top curvature direction distributes mass across parameters of varying scales, subject to the curvature structure of the network.

\subsubsection*{RPE via Parameter Scaling}

To understand how RPE manifests in the shared parameter setting, we now analyze how the top eigendirection $\vpsi$ of the scaled system:
\[
    \mB \vomega = \lambda_1 \vomega  
\]
responds to variation in the relative magnitudes of parameters $\vartheta_i$ and $\vartheta_j$. Specifically, we define a generalized rotation measure:
\[
    R^2(i,j) := \left( \frac{\omega_j}{\omega_i} \right)^2 = \left( \frac{v_j \cdot \vartheta_j}{v_i \cdot \vartheta_i} \right)^2 \,,
\]
which reflects the relative alignment of the top eigenvector $\vv_1$ along directions $i$ and $j$ under parameter scaling. While the absolute-value form was easier to work with in the disjoint setting, we use a quadratic form here to maintain positivity and clarity of analysis. 

From the structure of the Hessian and DLN terms, the entries $H_{kj}$ are governed by:
\[
    H_{kj} \sim \sum_{a,b} \frac{\partial^2 z}{\partial \Theta_a \partial \Theta_b} \cdot \frac{\Theta_a \Theta_b}{\vartheta_k \vartheta_j} \cdot \mathbb{I}[k \in \mathcal{I}_a, j \in \mathcal{I}_b]
    + \sum_a \frac{\partial z}{\partial \Theta_a} \cdot \frac{\Theta_a}{\vartheta_k \vartheta_j} \cdot \mathbb{I}[k, j \in \mathcal{I}_a] \,.
\]
Thus, $H_{kj}$ decays approximately as $1 / (\vartheta_k \vartheta_j)$.

Substituting into the definition of $\mB$, we obtain:
\[
    B_{kj} = \frac{\vartheta_k}{\vartheta_j} H_{kj} \sim \frac{1}{\vartheta_j^2}
    \cdot \overbrace{\vartheta_k \vartheta_j H_{kj}}^{\text{curvature weight}}  \,.
\]
That is, the magnitude of $B_{kj}$ is inversely related to $\vartheta_j^2$, modulated by the contribution of $\vartheta_j$ to DLN curvature. Consequently, parameters with small magnitude $\vartheta_j$ contribute more heavily to the total influence of $\omega_j$ in the eigensystem.

Consider now a fixed index $i$ and vary $\vartheta_j = r_{ij} \cdot \vartheta_i$, with $r_{ij} > 1$. To isolate the effect of relative scale, we assume that the curvature terms in $H_{kj}$ remain fixed, i.e., the structure of DLNs and their gradients are held constant. As $r_{ij} \to \infty$, $\vartheta_j^2 \gg \vartheta_i^2$ and thus $B_{kj}$ becomes negligible for all $k$. To maintain the eigenbalance for the eigenequation $\mB \vomega = \lambda_1 \vomega$, the solution must reduce $\omega_j$ relative to $\omega_i$, implying:
\[
    \frac{d R^2(i,j)}{d r_{ij}} > 0 \,.
\]
This recovers the same qualitative signature of RPE found in the disjoint DLN setting: as parameter scales diverge, the top eigendirection shifts toward the direction of smaller-scale (i.e., higher-curvature) parameters.

In this section, we extended our derivation to the shared parameter setting. While the constancy constraint $\psi_j = \text{const}$ no longer holds exactly under parameter sharing, the eigenproblem $\mB \vomega = \lambda_1 \vomega$ imposes a global balance across scaled coordinates. The structure of $\mB$ ensures that this balance remains sensitive to parameter scale, favoring alignment with small-scale (sharp) parameters. 


\subsection{Monotonicity of RPE} \label{app:rpe:fmnist-mono}

\begin{figure}[h]
\centering
\includegraphics[width=0.9\linewidth]{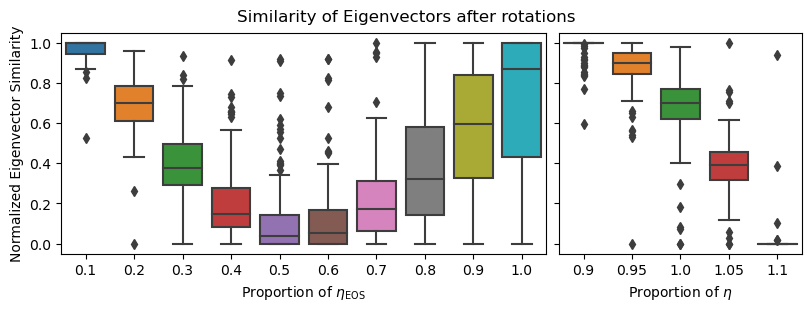}
\caption[Monotonicity of eigenvector rotations in MLPs.]{\textbf{Monotonicity of eigenvector rotations in MLPs.} We sample trajectories using varied learning rates and plot the similarities between leading eigenvectors before and after updates. The results support the phased behavior of $\gamma_\beta$ and the monotonicity of RPE. \textbf{Left:} Learning rates as a proportion of the threshold $\eta_\mathrm{EOS}$. \textbf{Right:} Learning rates as a proportion of the current learning rate.}
\label{fig:rpe:mono-sim}
\end{figure}

In Subsection~\ref{sec:rpe:rot}, we introduced the concept of RPE. Here, we further substantiate our theoretical claims by intervening just prior to the peak of instability events during training. Specifically, we train $15$ MLPs on fMNIST with the same hyperparameters except for seed of initialization, and we identify $T$ as the epoch indices where training loss reaches a local maximum, interpreted as a peak of instability. Across the $15$ seeds, we identify a total of $107$ such peaks. For each instability, we reinitialize models at epoch $T-4$ and continue training for two additional update steps. We then measure the cosine similarity of the leading Hessian eigenvectors before and after these updates.

These eigenvector similarities are normalized to the interval $[0, 1]$ and presented in Figure \ref{fig:rpe:mono-sim}. On the left panel, we observe an inverted-U pattern across learning rates scaled relative to $\eta_\mathrm{EOS}$. This pattern is consistent with the theoretical phases of $\gamma_\beta$ described in Figure \ref{fig:rpe:cartoon}. The right panel shows that when learning rates are scaled relative to the current learning rate, we observe a nearly linear trend: the lowest tested learning rate ($0.9\eta$) preserves eigenvector orientation the most, while the highest ($1.1\eta$) leads to the largest deviation. This empirical behavior supports the monotonicity property of RPE, highlighting that increasing learning rates within the instability regime amplify RPE.

\subsection{A t-SNE plot of the instability in Figure~\ref{fig:rpe:mono_rot}. }

\begin{figure}[h]
\centering
\includegraphics[width=0.8\textwidth]{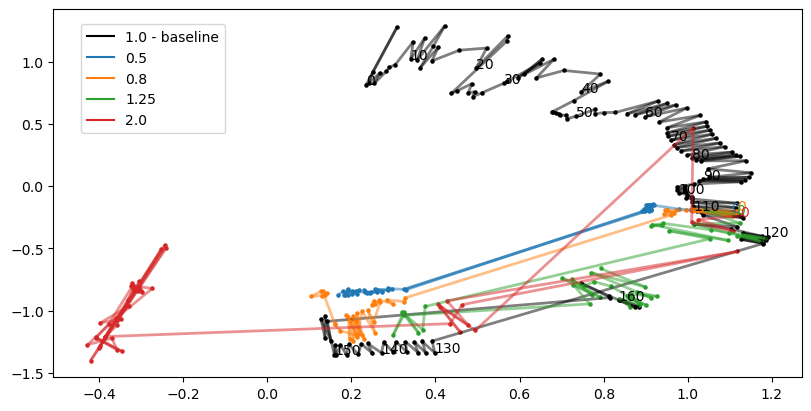}
\caption[RPE, Monotonicity in MLPs, t-SNE.]{\textbf{RPE, Monotonicity in MLPs, t-SNE.} We intervene \emph{post-hoc} on learning-rates right before instability is resolved, at scales $\eta_\mathrm{int} \in [0.5\eta, 0.8\eta, \eta, 1.25\eta, 2.0\eta]$, where $\eta_\mathrm{int} = \eta$ is the intervention-free baseline. We use the same color legend as Figure~\ref{fig:rpe:mono_rot}. }
\label{fig:rpe:rot_tsne}
\end{figure}

\newpage
\section{\label{app:flat} Proofs, derivations, and additional experiments for Section~\ref{ch:flatness} - flatness bias in GD}

In this appendix chapter, we supplement our dynamical model for $X_t = \log U_t$. Appendix~\ref{sec:flat:dynamics} describes the choices of conditions more clearly. Appendix~\ref{app:flat:fmnist-ut} supports our characterization of $U_t$ empirically. 

Throughout this appendix, we use Condition~\ref{cond:flat:standing}, which guarantees a strictly increasing $\alpha : \mathbb{R} \rightarrow \mathbb{R}^+$. We provide an outline of proofs in Appendix~\ref{app:flat:proof-outlines}, with detailed proofs in Appendices~\ref{app:flat:pm-contract-proof}-\ref{app:flat:median-drift-unimodal-proof}. 

Finally, Appendix~\ref{sec:flat:xt-sim} provides a numerical simulation for $X_t$. 

\subsection{Evolution of Curvature Under Training Instabilities} \label{sec:flat:dynamics}

Here, we develop the properties and conditions for a dynamical process model for GD updates in instabilities. 

\subsubsection*{Setup}

RPE prescribes rotations that scale strictly with the learning rate: beyond the stability limit, larger instability surpluses lead to larger eigenvector rotations. 

Directly modelling the discrete-time evolution of $\delta_t$, the instability surplus, is challenging, as the model of $\lambda$ typically studied in the \emph{descent lemma} models reality poorly. In addition to the weakness of \lammax, we find that updates in $\delta_t$ that depend \emph{only} on its past values to be inadequate to capture the full behavior. We find instead the leading eigenvalue also depends heavily on the current distance of parameters from the local minima, which changes $\lambda$ through the effect of higher derivatives. For example, one can consider the initial iterations from whence the GD training trajectories enter instability, where $\delta_t$s are likely to show a consecutive sequence of positive updates as the distance from minima grows rapidly as the model climbs up the loss surface. This dependence on distance, which cannot be easily measured, makes it extremely challenging to formulate a memory-less update rule for $\delta_t$. 

In our attempt to overcome these challenges, we instead conduct our analysis on the HOCMs of the loss landscape. We introduce $U_t$, a positive scalar that captures the non-quadratic content of the loss around the minima, in the directions of leading eigenvectors. The general properties of $U_t$ are:
\begin{itemize}
    \item Larger $U_t$ corresponds to larger HOCMs in the loss landscape
    \item $U_t$ influences $\lambda_{1,t}$ as HOCMs contributes to the leading eigenvalue
    \item Conversely, $\delta_t$ affects $U_t$ through eigenvector rotations. Its magnitude grows with $\delta_t$ and injects variability into the HOCMs
\end{itemize}

\noindent
While our derivations depend on the general properties of $U_t$, a possible concrete definition for $U_t$ is an \emph{even-order envelope} around the top-curvature direction. Let $\vv_{1,t}$ denote the leading eigenvector of the Hessian at $\vtheta_t$, and let $\mathcal C_t$ be a small cone of directions around $\vv_{1,t}$ (e.g. an angular cap). For a small radius $\rho_0>0$, define:
\begin{equation}\label{eqn:flat:Ut-even-env-definition}
    U_t  := 
    \mathbb{E}_{\vv\in \mathcal C_t} 
    \sup_{0<\rho\leq \rho_0} 
    \frac{\bigl| 
    \mathcal{L}(\vtheta_t+\rho \vv)+\mathcal{L}(\vtheta_t-\rho \vv)-2\mathcal{L}(\vtheta_t) - \tfrac12 \lambda_{1,t} \rho^2
        \bigr| }
      {\rho^{4}} \,.
\end{equation}
This quantity measures the \emph{even, non-quadratic remainder} of the loss after removing the local quadratic term. It is strictly positive and grows when any even-order curvature content (quartic, sextic, \dots) increases. In the limit $\rho\to0$ and for $v=v_{1,t}$, a Taylor expansion yields:
\begin{equation*}
\mathcal{L}(\vtheta_t+\rho v)+\mathcal{L}(\vtheta_t-\rho v)-2\mathcal{L}(\vtheta_t)
 = \lambda_{1,t} \rho^2 + \tfrac{1}{12} D^4\mathcal{L}(\vtheta_t)[v,v,v,v] \rho^4 + O(\rho^6) \,,
\end{equation*}
so Equation~\ref{eqn:flat:Ut-even-env-definition} reduces (up to the constant $1/12$) to the \emph{directional fourth derivative} when quartic terms dominate. This connects $U_t$ to the empirical fourth-derivative proxy; by taking the supremum over $\rho$ and averaging over a cone $\mathcal C_t$, the definition also captures higher even orders when they are material.

\medskip
\noindent\textbf{Empirical proxy.}
When computing high-order derivatives is costly, a finite-difference surrogate along $v_{1,t}$ is:
\begin{equation}\label{eqn:flat:Ut-fd}
    \widetilde U_t^{(h)}
     := 
    \frac{\lambda_1(\vtheta_t+h \vv_{1,t}) - 2 \lambda_1(\vtheta_t) + \lambda_1(\vtheta_t-h \vv_{1,t})}{h^2},
    \qquad h>0 \, \, \text{small} \,, 
\end{equation}
which equals the quartic term up to $\mathcal{O}(h^2)$ and is sensitive to higher even orders as $h$ varies.

\medskip
\noindent \textbf{Minimal definition.} We stress that the specific realization of $U_t$ is \emph{illustrative}; all subsequent analysis only relies on the interaction and update equations below (Equation~\ref{eqn:flat:Ut-dynamics}), which treat $U_t$ as a positive scalar whose logarithm evolves multiplicatively and which modulates typical sharpness:
\begin{equation}\label{eqn:flat:Ut-dynamics}
\log U_{t+1}  =  \log U_t  +  \xi_t  f_{\log U}(\delta_t),
\qquad
\delta_t  =  \eta \lambda_{1,t} - C_t
 =  \eta f_\lambda(U_t) - C_t + \eta \zeta_t
\end{equation}

\noindent
In particular, $f_{\log U}(\cdot)$ increasing in $\delta_t$ captures that stronger rotations alters non-quadratic structure proportionally, as is the case with its effects on curvature, while the non-decreasing function $f_\lambda$ expresses that larger HOCMs bias the \emph{typical} curvature upward. Our results depend only on these properties of $U_t$, not on choosing any particular expression. 

Our goal, in introducing $U_t$ and its setup, is to model a bidirectional coupling between a curvature quantity and rotations in a form that is both analytically tractable and faithful to the instability dynamics revealed by our analysis and the empirical support of RPE. In the remainder of this section, we formalise and justify the interaction between $U_t$ and $\delta_t$, leading to a dynamical system whose long-term behavior reveals a flatness bias for GD training during instabilities.

\subsubsection*{Instability surplus}
We now formalize the central quantity that mediates eigenvector rotations from RPE and curvature dynamics. Motivated by the stability analysis of the \emph{descent lemma} and \emph{Edge of Stability (EoS)} \cite{nesterov2004convex, cohen2022eos}, we define the \emph{instability surplus} at iteration $t$ as follows. 
\begin{definition} \label{def:delta_t}
The instability surplus is: 
\[
    \delta_t := \eta \lambda_{1,t} - C_{\mathrm{eos}, t} \,,
\]
where $\lambda_{1,t}$ is the largest eigenvalue of the Hessian at iteration $t$, $\eta$ is the learning rate, and $C_{\mathrm{eos}, t} > 0$ is a constant determined by the local loss landscape such that $\eta_\mathrm{eos}\lambda_{1,t} = C_{\mathrm{eos}, t}$ corresponds to the stability boundary. 
\end{definition}

Note that $\eta$ does not use the time-index $t$ as we consider training trajectories with constant learning rates. 

In the case of quadratic loss, this threshold takes the familiar value $C_{\mathrm{eos}, t} \equiv 2$, since stability is lost when $\eta\lambda_{1,t} > 2$. In more general settings, $C_{\mathrm{eos}, t}$ may depend on local anisotropy, non-quadratic curvature, or other structural properties of the loss. Contributions from positive higher-order terms of the Taylor expansion leads to $C_{\mathrm{eos}} < 2$. 

Our analysis will focus on the \emph{instability regime} where $\delta_t > 0$. In this regime, RPE theory predicts that eigenvector rotations occur and their magnitude grows strictly with $\delta_t$. Conversely, $\delta_t \leq 0$ corresponds to the stable regime in which eigenvector rotate stably toward sharper curvature directions - these rotations are barely observed empirically; we will treat $\delta_t \leq 0$ as an exit condition from the system under consideration.  

While the immediate role of the instability surplus $\delta_t$ is to determine the rotation of eigenvectors, the rotations also act as a driver for the dynamics of $U_t$, which we introduce below.

\subsubsection*{Higher-Order Curvature Moments}

We now introduce $U_t \in \mathbb{R}_{>0}$, a positive scalar that describes the deviation of the loss landscape from quadratic expansions among the directions of leading eigenvectors. In Section~\ref{ch:phases}, we briefly discussed the `Hessian variance', whose increase in magnitude signals a deterioration in the local `quadraticness' of the loss landscape, so it measured \emph{curvature nonlinearity}. We see $U_t$ and HOCMs as a measure of the local non-linearity of curvature. 

A material realization of $U_t$ is in Equation~\ref{eqn:flat:Ut-even-env-definition}, and as an example, it can be intuitively and analogously compared to the fourth derivative of the loss with respect to parameters. While we will not compute such derivatives explicitly in the analysis, this analogy conveys the intended role of $U_t$ through its properties: it tracks the evolving capacity of the landscape to produce sharp and convex changes in curvature. 

Positivity of $U_t$ is a property we expect around locally convex minima in parameter space, which is grounded in empirical evidence (where upwards convex curvatures are observed around local minima) and theoretical studies in the literature, where convexity is frequently assumed \cite{nesterov2004convex, boyd2004convex, hardt2016train, arora2019fine}. Consequently, we encode this positivity in the discrete-time updates of $U_t$ naturally by working in log-space and instead consider the evolution of $\log U_t$. This choice also means that multiplicative effects on $U_t$ are additive in $\log U_t$, simplifying subsequent analysis. 

Empirical evidence (see Appendix~\ref{app:flat:fmnist-ut}) supports that HOCMs fluctuate in sign from step to step without exhibiting a consistent upward or downward drift when the system remains in instability. This condition also underlines an epistemically agnostic position to the drift of $U_t$, since we do not directly observe $U_t$ via experimentation. To capture the sign symmetry, we introduce a sign variable $\xi_t$, taking values in $\{-1, 1\}$ with equal probability (a Rademacher random variable), independently at each time step. This variable determines whether $U_t$ increases or decreases at a given step.  

Formally, we consider the update rule:
\[
    \log U_{t+1} = \log U_t + \xi_t   f_{\log U}(\delta_t) \,,
\]
where $f_{\log U} : \mathbb{R}_{>0} \to \mathbb{R}_{>0}$ is a deterministic function describing the magnitude of log-space changes in $U_t$ as a function of the instability surplus $\delta_t$. The restriction $f_{\log U}(\cdot) > 0$ ensures that $\delta_t$ controls only the \emph{size} of updates, while their \emph{direction} is governed by the symmetric noise $\xi_t$. We note that we have chosen an additive noise in log-space with the above formulation, which we justify in the following subsection. 

\subsubsection*{Coupling of surplus and Higher-Order Curvature Moments}

The dynamics between $\delta_t$ and $U_t$ is based on a coupling. The dependency of $U_t$ on $\delta_t$ encodes our RPE claims, which were made and justified in Section~\ref{ch:rpe}. The converse link, from $U_t$ to $\delta_t$, reflects our definition of $U_t$ as the HOCMs, detailed in the previous subsection. Here, we expound and explain the two links separately. 

First, the RPE results in Section~\ref{ch:rpe} suggest that the magnitude of curvature fluctuations (strictly) increases with the instability surplus. We expect this relationship to extend to HOCMs:  
\[
\delta_t^{(1)} < \delta_t^{(2)}
\quad \Longrightarrow \quad
f_{\log U}(\delta_t^{(1)}) < f_{\log U}(\delta_t^{(2)})
\]
for any positive $\delta_t^{(1)}, \delta_t^{(2)}$. This strict increase formalizes the idea that stronger instability amplifies the size of $U_t$ updates.  

We choose a multiplicative form of noise for $U_t$. Numerical experiments in the RPE setting revealed that, \emph{ceteris paribus}, varying the instability surplus via modifications to the learning rate led to changes in the leading eigenvalue of the Hessian proportional to the magnitude of the instability surplus (for details, see Subsection~\ref{sec:rpe:exp} and Appendix~\ref{app:rpe:fmnist-mono}). This is consistent with the role of rotations as a multiplier of change in curvature, which we measured directly. Since we draw analogy between $U_t$ to curvature-like quantities (e.g., a curvature of curvature, which is the fourth derivative of the loss), we find it reasonable to posit that rotations have a multiplicative effect on $U_t$. Multiplicative noise in $U_t$ implies additive noise in the $\log U$ updates. 

Second, we model the leading eigenvalue $\lambda_{1,t}$ as a noisy observation of a deterministic function of $U_t$:
\[
    \lambda_{1,t} = f_\lambda(U_t) + \zeta_t \,,
\]
where $\mathbb{E}[\zeta_t \mid U_t] = 0$ and $f_\lambda : \mathbb{R}_{>0} \to \mathbb{R}_{>0}$ satisfies the strictly increasing property \emph{in expectation}. For $U_a < U_b$, we have:
\[
    \mathbb{E}[\lambda_{1,t} \mid U_t = U_a]  \leq  \mathbb{E}[\lambda_{1,t} \mid U_t = U_b] \,.
\]
This reflects the interpretation of $U_t$ as the HOCMs of the loss landscape in a convex setting: on average, larger $U_t$ corresponds to sharper local curvature and hence a larger expected $\lambda_{1,t}$. $U_t$ therefore, captures persistent, higher-order structure in the loss landscape that is not reflected in $\delta_t$ alone. 

Conversely, the instability surplus $\delta_t$ is the difference between the scaled leading curvature $ \eta f_\lambda(U_t) $ (plus zero-mean noise $\eta\zeta_t$) and the local stability threshold $C_t$ (recall $C_t=2$ in a quadratic setting). The function $f_\lambda$ need not be linear or bounded; only the increasing property in expectation is required in our analysis.  

Combining these relationships yields the coupled system:
\[
    \log U_{t+1} = \log U_t + \xi_t   f_{\log U}(\delta_t) \quad
    \delta_t = \eta \lambda_{1,t} - C_t = \eta f_\lambda(U_t) - C_t + \eta \zeta_t \,.
\]
The first equation states that $U_t$ evolves as a multiplicative random walk, in which the step size depends on the instability surplus. The second equation closes the loop by expressing the surplus as the difference between the learning-rate scaled sharpness determined by $U_t$ (plus noise) and the stability threshold $C_t$.  

This coupling forms a feedback system: $\delta_t$ controls the fluctuation magnitude of $U_t$, while $U_t$ in turn governs $\lambda_{1,t}$ and thus $\delta_t$. The instability regime $\delta_t > 0$ sustains this feedback, whereas crossing into $\delta_t \leq 0$ halts it, marking an exit into stability.  

\begin{tcolorbox}[title=Example for Intuition ,colback=white,colframe=black]
Consider the illustrative choices:
\[
    f_{\log U}(\delta) = k \delta, \quad f_\lambda(U) = a U^{b}
\]
with scalar constants $a>0$, $b>0$, $k>0$.  

From the coupled system:
\begin{itemize}
    \item \textbf{Step-size scaling:} $f_{\log U}(\delta_t) = k \delta_t$ shows that doubling the instability surplus doubles the magnitude of multiplicative changes in $U_t$.  
    \item \textbf{Sharpness-curvature link:} $f_\lambda(U) = a U^b$ is strictly increasing, so a decrease in $U_t$ produces a decrease in the expected sharpness $\lambda_{1,t}$, which reduces the surplus $\delta_t$.
\end{itemize}

This example is solely for intuition. The subsequent proofs do not assume these specific forms, and only the increasing properties of $f_{\log U}$ and $f_\lambda$ stated in \textbf{C3} and \textbf{C4} of Condition~\ref{cond:flat:standing} are used.
\end{tcolorbox}

\subsubsection*{Initial conditions}

We initialize the system in the instability regime. Specifically, we consider $t=0$ to be a time such that:
\[
    \delta_0 > 0
\]
so that the initial instability surplus is strictly positive. This choice ensures that the feedback loop between $U_t$ and $\delta_t$ described above is active from the outset. 

Once $\delta_t$ becomes non-positive, the system enters the stable regime. In our framework, such a transition is treated as an exit condition: the fluctuation dynamics of $U_t$ cease to be driven by instability, and the subsequent evolution of $U_t$ falls outside the scope of the present analysis.

\subsection{Properties of High-order Curvature} \label{app:flat:fmnist-ut}

\begin{figure}[h]
\centering
\includegraphics[width=0.9\linewidth]{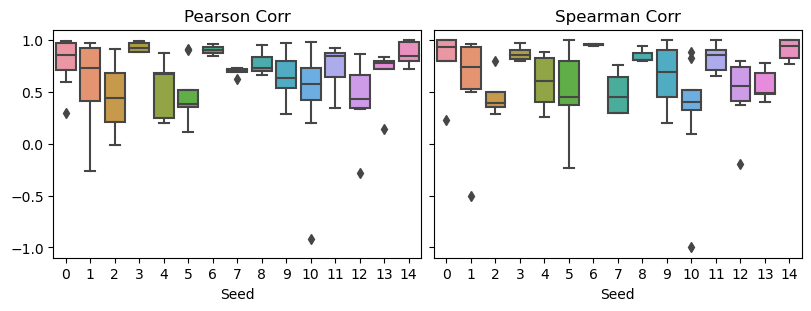}
\caption[Correlation between $|\gamma_{U_t}|$ and $\lambda_{1,t}$.]{\textbf{Correlation between $|\gamma_{U_t}|$ and $\lambda_{1,t}$.} High linear and rank correlations observed across seeds.}
\label{fig:flat:ut_corr}
\end{figure}

In Subsection~\ref{sec:flat:dynamics}, we introduced $U_t$, a quantity designed to capture the effect of higher-order curvature moments (HOCMs) across time. To empirically evaluate this quantity, we use the fourth-order directional derivative along the gradient direction as a proxy for $U_t$.

We train $15$ MLPs on fMNIST and identify $107$ instances of instability, defined as points where the leading Hessian eigenvalue $\lambda_{1,t}$ exceeds 20 (given our fixed learning rate $\eta = 0.1$). For each such event, we compute the change in higher-order curvature using the following metric:
\[
    |\gamma_{U_t}| \coloneqq \max\left(\frac{U_t}{U_{t+1}}, \frac{U_{t+1}}{U_t}\right) \, .
\]

We then analyze the correlation between this metric and $\lambda_{1,t}$. Results in Figure \ref{fig:flat:ut_corr} show strong positive correlations, both Pearson and Spearman, suggesting that higher values of $\lambda_{1,t}$ are associated with greater variability in $\log(U_t)$. These results lend empirical support to \textbf{C3} of Condition~\ref{cond:flat:standing}, which posits that the variability in $U_t$ increases with the sharpness of the loss landscape. While the assumption requires a strictly increasing relationship, the observed strong linear and rank correlations are consistent with a monotone relationship; while \textbf{C3} of Condition~\ref{cond:flat:standing} requires strict increase, the empirical evidence supports this as a useful approximation.

\begin{figure*}[h]
    \centering
    \begin{tabular}[b]{c|c|c|c|c}
    & \multicolumn{2}{c|}{\textbf{Pearson}} & \multicolumn{2}{c}{\textbf{Spearman}}\\
    \hline
    Index & statistic & p-value & statistic & p-value\\
    \hline
    1 & 0.4902 & 8.48E-04 & 0.5193 & 3.59E-04 \\
    2 & 0.8368 & 3.23E-11 & 0.6840 & 1.58E-06 \\
    3 & 0.5957 & 8.24E-04 & 0.5348 & 3.37E-03 \\
    4 & 0.6984 & 3.58E-05 & 0.6218 & 4.12E-04 \\
    5 & 0.5028 & 2.44E-03 & 0.7317 & 8.80E-07 \\
    6 & 0.2995 & 2.36E-02 & 0.3157 & 1.67E-02 \\
    7 & 0.6317 & 1.15E-02 & 0.6429 & 9.74E-03 \\
    8 & 0.6743 & 1.15E-04 & 0.6099 & 7.31E-04 \\
    9 & 0.7917 & 1.90E-05 & 0.7987 & 1.42E-05 \\
    10 & 0.6291 & 1.37E-05 & 0.6270 & 1.50E-05 \\
    11 & 0.5716 & 6.59E-07 & 0.5314 & 5.25E-06 \\
    12 & 0.5117 & 5.33E-04 & 0.6999 & 2.50E-07 \\
    13 & 0.4572 & 4.45E-03 & 0.4720 & 3.18E-03 \\
    14 & 0.8128 & 6.26E-11 & 0.5856 & 4.62E-05 \\
    15 & 0.7632 & 2.52E-08 & 0.6700 & 4.21E-06 \\
    \hline
    \textbf{All} & 0.5905 & 1.50E-53 & 0.5485 & 5.42E-45 \\
    \hline
\end{tabular}
\captionof{table}[Aggregate correlation statistics between $|\gamma_{U_t}|$ and $\lambda_{1,t}$.]{\textbf{Aggregate correlation statistics between $|\gamma_{U_t}|$ and $\lambda_{1,t}$.} Strong correlations across seeds affirm \textbf{C3} of Condition~\ref{cond:flat:standing}.}
\label{fig:flat:ut_corr_table}
\end{figure*}

Further evidence is presented in Table \ref{fig:flat:ut_corr_table}, which aggregates correlation statistics across all $15$ seeds. The consistently low p-values strongly reject the null hypothesis of no correlation, reinforcing the robustness of this relationship despite potential noise in high-order empirical measurements across different instances of instability. 

\begin{figure}[h]
\centering
\includegraphics[width=0.6\linewidth]{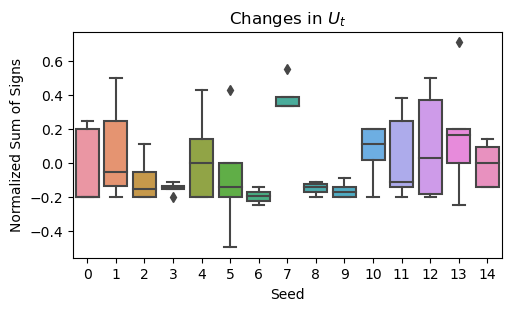}
\caption[Sign of $\gamma_{U_t}$.]{\textbf{Sign of $\gamma_{U_t}$.} Sum of signs varies across instability episodes around $0$.}
\label{fig:flat:ut_sign}
\end{figure}

Additionally, in \textbf{C3} of Condition~\ref{cond:flat:standing}, we introduced $\xi_t$, a discrete stochastic variable uniformly sampled from $\{-1, 1\}$. To assess the empirical consistency of this model, we examine the sign of $\gamma_{U_t}$, assigning a value of $+1$ when $U_{t+1} > U_t$, $-1$ when $U_{t+1} < U_t$, and omitting values when $U_{t+1} = U_t$ due to floating-point tolerance. Figure \ref{fig:flat:ut_sign} presents the cumulative sign values across continuous instability phases for each seed. The balanced distribution of signs provides support for the hypothesis that $\xi_t$ behaves as a uniform discrete random variable in $\{-1, 1\}$ in practice.

\subsection{Proof Outlines}\label{app:flat:proof-outlines}

For the reader’s convenience, we provide here a high-level roadmap of the proofs collected in the appendix. The detailed arguments are given in the following Appendices~\ref{app:flat:pm-contract-proof}-\ref{app:flat:median-drift-unimodal-proof}, but the essential logical flow is as follows. Note, we mention specific terms but leave the formal definition to the relevant proofs.

\medskip

\paragraph{Theorem~\ref{thm:flat:point-contraction} (Point–Mass Contraction)}  
We begin with the simplest initial condition, where $X_0$ is a point mass. The key observation is that when $\alpha$ is strictly increasing, cancellations of a $+$ step with a $-$ step are biased to the left:
\begin{itemize}
    \item \emph{Adjacent cancellations are negative:} the two-step patterns $(+,-)$ and $(-,+)$ always move leftward, since the second step overshoots due to strict increase of $\alpha$.
    \item \emph{Zero-sum strings:} any zero-sum sign–string can be decomposed by recursive constructors (`Overset' and `Add'), starting from the basic two-step cancellations.
    \item \emph{Inductive negativity:} both constructors preserve negativity, so every zero-sum string yields a strict left move; negative-sum strings decompose into zero-sum substrings and unmatched minuses, which push further left.
\end{itemize}
Thus, every nonpositive-sum sign–string is strictly negative. Specializing to $t=2$ shows three of four branches lie left of the starting point, so $F_2(u-\varepsilon)\geq 3/4$ for small $\varepsilon>0$. With global bounds $\alpha\geq c_0>0$, $\alpha'\geq c_1>0$, one obtains a quantitative gap $\gtrsim (t/2)c_0 c_1$.

\medskip

\paragraph{Corollary~\ref{cor:flat:mass-conditions-median} (Mass-conditions for median contraction)}  
At two steps, partitioning the line into four regions $A,B,C,D$ around a median $m_0$ shows that in each case at least half the mass lies below $m_0$ after two updates. This guarantees the median does not increase.

\medskip

\paragraph{Theorem~\ref{thm:flat:quantile-drift} (Upper–envelope decrease)}  
Define the per-ancestor two-step median $M_2(x)$ and its upper envelope $Z_t:=\esssup m_t(x)$. Since each ancestor median decreases by at least $c_0 c_1$, the envelope drifts left by a uniform gap every two steps. All medians are dominated by this envelope, so they too must decrease.

\medskip

\paragraph{Lemma~\ref{lem:flat:edge-strict-descent} (Strict monotonicity at the edges)}  
Assuming unimodality of $f_0$ and Lipschitz control $0 < \alpha'\leq L_\alpha<1$, each branch map $T_s$ is increasing. Studying the pre-image on a sliding window $[x,x+w]$ shows that branch contributions are monotone outside a central band: nondecreasing on the far left, nonincreasing on the far right. Beyond these limits the smoothed density is therefore monotone, and strict in the nondegenerate case.

\medskip

\paragraph{Lemma~\ref{lem:flat:clt-descent} (CLT-bounded unimodality in the middle)}  
Typical sign–strings behave like a random walk, staying within $\kappa\sqrt{t}$ excursions. A discrete Grönwall inequality controls deviations of $T_s$ from the starting mode, giving $|T_s(m^*)-m^*|\lesssim \kappa\sqrt{t}$. Thus typical branches stay within a central CLT band, while atypical branches have exponentially small weight. Inside this band, Lemma~\ref{lem:flat:edge-strict-descent}’s edge monotonicity extends with high probability, ensuring unimodality.

\medskip

\paragraph{Proposition~\ref{prop:flat:median-dominate-mode} (Median dominates the mode)}  
Conditioning on two-step prefixes, the \emph{straddle lemma} positions the ancestor mode between ordered preimages of the descendant mode. The \emph{cross-balance lemma} then shows $F_t(\text{mode})\geq 1/2$. Hence, every median lies at or to the right of the mode: $m^*\leq m_t$.

\medskip

\paragraph{Theorem~\ref{thm:flat:median-drift-unimodal} (Median drift under unimodality and $1$-Lipschitz $\alpha$)}  
Combining Lemmas~\ref{lem:flat:edge-strict-descent}-\ref{lem:flat:clt-descent} (high-probability unimodality), Proposition~\ref{prop:flat:median-dominate-mode} (mode-median ordering), and local two-step contraction bounds shows that medians decrease monotonically at even times:
\[
    m_{t+2} < m_t, \qquad t\ \text{even} \,.
\]
Under nondegeneracy ($\alpha(m_t)>0$, $\alpha'(u)>0$ nearby), this yields a quantitative strict drift:
\[
     m_{t+2} \leq m_t - p_t \alpha(m_t)\,\underline{\alpha}'(m_t) \,.
\]
Thus medians move steadily leftward over time.

\subsection{Proof of Point-Mass Contraction, Theorem~\ref{thm:flat:point-contraction}} \label{app:flat:pm-contract-proof}
\subsubsection*{Properties of Updates}

We study the pathwise behavior of the stochastic process:
\[
X_{t+1} = X_t + \xi_t  \alpha(X_t), \quad t\geq 0
\]
with initial state $X_0=u \in \mathbb{R}$, increments $\xi_t\in\{-1,+1\}$ independent and equally likely, and $\alpha:\mathbb{R}\to \mathbb{R}_+$ finite and \emph{strictly} increasing. For each $t\geq 1$ we write:
\[
S_t \coloneqq \sum_{i=0}^{t-1} \xi_i
\]
for the cumulative sign–sum and:
\[
F_t(x) \coloneqq \mathbb{P}(X_t \leq x)
\]
for the distribution function of $X_t$.

\paragraph{Auxiliary notation}
For a finite sign–string $S=(s_0,\dots,s_{m-1})\in\{-1,+1\}^m$ and a starting point $x\in\mathbb{R}$ we define the trajectory $(x_k)_{k=0}^m$ by:
\[
x_{k+1} = x_k + s_k  \alpha(x_k), \quad x_0=x
\]
and denote the terminal state:
\[
\Phi(x;S) \coloneqq x_m
\]
We say $S$ is \emph{negative} at $x$ if $\Phi(x;S) < x$. By convention, we write $+$ and $-$ for $+1$ and $-1$.

\paragraph{Motivation}
To control the distribution $F_t$, it suffices to understand how sign–strings $S$ affect $\Phi(x;S)$ relative to $x$. A general sign–string can be decomposed into patterns whose effects are easier to analyze. We show, in order:
\begin{enumerate}
    \item Two–step adjacent pairs are strictly negative 
    \item Zero–sum substrings can be generated from simple operators (Add and Overset), each strictly negative
    \item Negative–sum strings can be decomposed into zero–sum substrings interleaved with isolated $-1$ steps, ensuring strict negativity overall
\end{enumerate}
Note that we do not assume but carefully derive the observation cancellations between paired $+-$ or $-+$ signs will lead to left drift. This structural decomposition allows us to deduce that, when $t$ is even, at least half the mass of $X_t$ lies strictly below $u$, establishing contraction in distribution.

\paragraph{Adjacent cancellations}

A key building block for analyzing arbitrary sign–strings is the behavior of adjacent two–step patterns. In particular, we study the substrings:
\[
(+1,-1), \quad (-1,+1)
\]
Each represents a cancellation of one forward and one backward move, but because $\alpha$ is increasing, the cancellation is not exact: the net effect is strictly negative.

\begin{lemma}[Two–step adjacent cancellations are negative]\label{lem:flat:adjacent-two}
Fix $x\in\mathbb{R}$ and consider the sign–strings $S_+=(+1,-1)$ and $S_-=(-1,+1)$. Then since $\alpha$ is \emph{strictly} increasing:
\[
\Phi(x;S_+) - x = \alpha(x) - \alpha\bigl(x+\alpha(x)\bigr) < 0
\]
\[
\Phi(x;S_-) - x = -\alpha(x) + \alpha\bigl(x-\alpha(x)\bigr) < 0
\]
\end{lemma}

\begin{proof}
We check the two cases separately.

\textbf{Case 1: $S_+=(+1,-1)$.} Let $x_1 = x+\alpha(x)$. The second update subtracts $\alpha(x_1)$, so the final state is:
\[
\Phi(x;S_+) = x_1 - \alpha(x_1) = x+\alpha(x) - \alpha(x_1)
\]
Hence:
\[
\Phi(x;S_+) - x = \alpha(x) - \alpha(x_1)
\]
Since $x_1 > x$ and $\alpha$ is strictly increasing, we have $\alpha(x_1)>\alpha(x)$, so the difference is strictly negative.

\textbf{Case 2: $S_-=(-1,+1)$.} Let $x_1 = x-\alpha(x)$. The second update adds $\alpha(x_1)$, so the final state is:
\[
\Phi(x;S_-) = x_1 + \alpha(x_1) = x-\alpha(x) + \alpha(x_1)
\]
Hence:
\[
\Phi(x;S_-) - x = -\alpha(x) + \alpha(x_1)
\]
Since $x_1 < x$ and $\alpha$ is strictly increasing, we have $\alpha(x_1)<\alpha(x)$, so again the difference is strictly negative.
\end{proof}

\noindent
Intuitively, in $(+,-)$ the process first moves right to a larger state where $\alpha$ is larger, so the backward step overshoots the forward step. In $(-,+)$ the process, first moves left to a smaller state where $\alpha$ is smaller, so the forward step is insufficient to recover the initial drop. In both cases the two–step string results in a strictly negative net displacement.

\subsubsection*{Primitive zero-sum sign strings}

We next analyze sign–strings whose total sum is zero. These represent trajectories with as many $+1$ as $-1$ updates. One might expect such strings to return to their starting point, but the strictly increasing property of $\alpha$ ensures that they actually end strictly below. To formalize this, we introduce primitive strings and two constructive operations. 

\begin{definition}[Primitive sign–strings]\label{def:flat:primitive}
Let $S=(s_0,\dots,s_{m-1})$ be a sign–string of length $m$. We say $S$ is \emph{primitive} if no proper nonempty prefix has total sum zero, that is:
\[
\sum_{i=0}^j s_i \neq 0 \quad \text{for all } 0<j<m
\]
Equivalently, a primitive zero–sum string returns to level zero only at its final step.
\end{definition}

\subsubsection*{Two constructive operations - Add and Overset}
For a sign–string $\theta$ we define:
\[
\mathrm{Add}(S,\theta) \coloneqq S\cdot\theta \quad \text{(concatenation)}
\]
\[
\mathrm{Overset}_+(\theta) \coloneqq (+1)\cdot\theta\cdot(-1), 
\quad
\mathrm{Overset}_-(\theta) \coloneqq (-1)\cdot\theta\cdot(+1)
\]
That is, Overset wraps a string between opposite signs.

\begin{lemma}[Generation of zero–sum strings]\label{lem:flat:gen-zerosum}
Every zero–sum string $S$ can be obtained from the empty string by finitely many applications of $\mathrm{Overset}_\pm$ and $\mathrm{Add}$.
\end{lemma}

\begin{proof}
We argue by induction on the length $|S|$. 

\noindent
\textbf{Base case:} the empty string is trivial.  

\noindent
\textbf{Inductive step.} Let $S$ be zero–sum and nonempty. Let $j$ be the smallest index such that $\sum_{i=0}^j s_i=0$. The prefix $S^{(1)}=(s_0,\dots,s_j)$ is a primitive zero–sum string, and the suffix $S^{(2)}=(s_{j+1},\dots,s_{m-1})$ is zero–sum. Then:
\[
S = \mathrm{Add}(S^{(1)},S^{(2)})
\]

\noindent
Since $S^{(1)}$ is primitive, it begins and ends with opposite signs, so:
\[
S^{(1)} = \mathrm{Overset}_\pm(\sigma)
\]
for some shorter zero–sum string $\sigma$ obtained by deleting the first and last signs. By induction, both $\sigma$ and $S^{(2)}$ can be generated, hence so can $S$.
\end{proof}

Zero–sum strings can be cut at the first return to zero. Each primitive block is just a smaller zero–sum string wrapped between opposite signs, and these primitive blocks concatenate to form all zero–sum strings.

\begin{lemma}[Each incremental Overset is negative]\label{lem:flat:overset-neg}
Let $\sigma$ be any zero–sum sign-string with $\Phi\bigl(x, \sigma \bigr) = y$, and suppose $\sigma$ is negative, i.e. $y < x, \forall x$. Then for all $x\in\mathbb{R}$:
\[
    \Phi\bigl(x;\mathrm{Overset}_\pm(\sigma)\bigr) < x
\]
\end{lemma}

\begin{proof}
\textbf{Case $\mathrm{Overset}_+(\sigma)$.} 
First update: $y_+ = x+\alpha(x)$.  
Run $\sigma$ from $y_+$, yielding $z=\Phi(y_+;\sigma)$. By construction, $\sigma$ is negative, so $z<y_+$. 
Final update: $z-\alpha(z)$.  

If $z<x$, then $z-\alpha(z)<x$ trivially.  
If $z\geq x$, then $\alpha(z)\geq \alpha(x)$ and recall $z < y_+ = x+\alpha(x)$. Thus:
\[
z-\alpha(z) < y_+ - \alpha(z) = (x+\alpha(x)) - \alpha(z) \leq (x+\alpha(x))-\alpha(x) = x
\]

\textbf{Case $\mathrm{Overset}_-(\sigma)$.} We can use analogous reasoning: the initial move is leftward to $y_- = x-\alpha(x)$, running $\sigma$ yields $z<y_-$, and the final $+$ step gives $z+\alpha(z)<x$. 
\end{proof}

The Overset operator starts with a move away from $x$, inserts a zero–sum block (already negative), and ends with the opposite move. The initial shift ensures that the opposite move overshoots, forcing the terminal state strictly below $x$. 

\begin{lemma}[Each incremental Add is negative]\label{lem:flat:add-neg}
Let $\sigma$ be any zero–sum string with $\Phi\bigl(x, \sigma \bigr) = y$, and suppose $\sigma$ is negative, i.e. $y < x$. Then, for any zero-sum pair $\tau \in [(+-), (-+)]$, we have:
\[
\Phi \bigl(x; \mathrm{Add}(\sigma, \tau) \bigr) = \Phi(x;\sigma\cdot\tau) < x
\]
for all $x$. The same holds for a finite concatenation of zero–sum strings.
\end{lemma}

\begin{proof}
Write $y=\Phi(x;\sigma)$ and $z=\Phi(y;\tau)$. By Lemma~\ref{lem:flat:overset-neg}, $y<x$ and $z<y$. Thus $z<x$, proving strict negativity. 
\end{proof}

\begin{lemma}[All zero–sum strings are negative]\label{lem:flat:zerosum-neg}
For any zero–sum string $S$ and $x\in\mathbb{R}$, we have:
\[
\Phi(x;S) < x
\]
\end{lemma}

\begin{proof}
By Lemma~\ref{lem:flat:gen-zerosum}, $S$ can be generated using Overset and Add from a null sign-string. By Lemma~\ref{lem:flat:adjacent-two}, the first $\pm$ pair, which is generated from either of $\mathrm{Add}$ and $\mathrm{Overset}$, is negative. After the initial step, we continue by induction through repeated application of Lemmas~\ref{lem:flat:overset-neg} and \ref{lem:flat:add-neg}. Since each input sign-string is negative, the outputs from each incremental operation remains negative. Therefore $\Phi(x;S)<x$. 
\end{proof}

Zero–sum strings contain equal numbers of $+1$ and $-1$ steps, but cancellations are always biased to the left because $\alpha$ is increasing. Whether constructed by nesting (Overset) or by concatenation (Add), the final outcome is strictly below the starting point.

\subsubsection*{Negative–sum strings}

We now analyze sign–strings with net negative sum. The key idea is to factor any such string into zero–sum substrings interleaved with isolated $-1$ updates, then apply the zero–sum negativity and the strictly increasing property of $\alpha$. 

\begin{lemma}[First–passage decomposition]\label{lem:flat:record-low}
Let $S=(s_0,\dots,s_{m-1})$ be a sign–string with net sum $\sum_{i=0}^{m-1}s_i=-N<0$. Define partial sums $H_{-1}\coloneqq 0$ and $H_j\coloneqq \sum_{i=0}^{j}s_i$ for $0\leq j\leq m-1$. For $k=1,\dots,N$ define the first–passage indices as:
\[
T_k \coloneqq \min\{ j\geq 0 : H_j=-k\}
\]
Then $s_{T_k}=-1$ and $H_{T_k-1}=-(k-1)$ for each $k$. With $T_0\coloneqq -1$ and $T_{N+1}\coloneqq m$. Furthermore, the substrings:
\[
\theta_k \coloneqq (s_{T_k+1},\dots,s_{T_{k+1}-1}) \quad k=0,1,\dots,N
\]
are zero–sum (possibly empty), and we have the concatenation:
\[
S = \theta_0 \cdot (-1) \cdot \theta_1 \cdot (-1) \cdots \theta_{N-1} \cdot (-1) \cdot \theta_N
\]
\end{lemma}

\begin{proof}
By minimality of $T_k$, $H_{T_k-1}=-(k-1)$ and $s_{T_k}=-1$. For $k=0$, we get:
\[
\sum \theta_0 = H_{T_1-1}-H_{-1} = 0
\]
For $1\leq k\leq N-1$, we have:
\[
\sum \theta_k = H_{T_{k+1}-1}-H_{T_k} = (-k)-(-k) = 0
\]
For $k=N$, we can write:
\[
\sum \theta_N = H_{m-1}-H_{T_N} = (-N)-(-N) = 0
\]
Concatenation of the pieces yields the desired factorization.
\end{proof}

This proof cuts each sign-string at the first-passage of each record low of the running sum. Between two successive record lows, the running sum stays above the lower record and returns to it just before the next $-1$, so the substring in between must have zero total. This produces a unique left-to-right interleaving of zero–sum substrings and isolated $-1$’s.

\begin{lemma}[Negative–sum sign-strings are negative]\label{lem:flat:neg-sum}
Let our process have starting level $x$, and $S$ have net sum $-N<0$ and the decomposition of Lemma~\ref{lem:flat:record-low}:
\[
S=\theta_0 \cdot (-1) \cdot \theta_1 \cdot (-1) \cdots \theta_{N-1} \cdot (-1) \cdot \theta_N
\]
Define intermediate states as:
\[
x_0 \coloneqq x,\quad y_k \coloneqq \Phi\bigl(x_k;\theta_k\bigr),\quad x_{k+1} \coloneqq y_k - \alpha\bigl(y_k\bigr) \quad (k=0,1,\dots,N-1)
\]
and the terminal state $x^* \coloneqq \Phi\bigl(x_N;\theta_N\bigr)$. Then:
\[
x^* < x
\]
\end{lemma}

\begin{proof}
For each $k=0,1,\dots,N-1$ we compare successive states.

\noindent
\textbf{Zero–sum stage.} Since $\theta_k$ is zero–sum, Lemma~\ref{lem:flat:zerosum-neg} yields:
\[
y_k \leq x_k, \quad \text{with } y_k<x_k \text{ if } \theta_k \text{ is nonempty}
\]

\noindent
\textbf{Isolated $-1$ stage.} The update $x_{k+1}=y_k-\alpha\bigl(y_k\bigr)$ gives:
\[
x_{k+1} < y_k
\]

Chaining these inequalities over $k=0,1,\dots,N-1$ we obtain $x_N < x^{(0)}=x$. Finally, the terminal zero–sum substring satisfies:
\[
x^* = \Phi\bigl(x_N;\theta_N\bigr) \leq x_N
\]
with strict inequality if $\theta_N$ is nonempty. Combining yields $x^* < x$.
\end{proof}

The decomposition isolates the irreducible negative thrust into explicit $-1$ steps, with neutral oscillations quarantined into zero–sum substrings. Each zero–sum substring is strictly negative, and each isolated $-1$ step is nonincreasing (strictly decreasing when $\alpha>0$ at that state). Sequentially composing these effects forces a strict drop unless every zero–sum piece is empty and every $-1$ happens at a state where $\alpha=0$.

This approach avoids deleting arbitrary adjacent pairs. By cutting at \emph{first–passage} times to new record lows of the partial sum, the factorization into zero–sum substrings and isolated $-1$’s is canonical, and the sequential inequalities are evaluated at the realized entry states $x^{(k)}$, $y^{(k)}$, avoiding any unstated commutation of updates.

\subsubsection*{Density contraction}

We now translate the pathwise results into quantitative distributional statements about $F_t(x)=\mathbb{P}(X_t<x)$. 

From Lemma~\ref{lem:flat:zerosum-neg} and Lemma~\ref{lem:flat:neg-sum}, any realized sign–string $S$ with $\sum s_i\leq 0$ yields $\Phi(u;S)<u$. By symmetry of $S_t=\sum_{i=0}^{t-1}\xi_i$:
\[
\mathbb{P}(S_t<0)=\mathbb{P}(S_t>0), \quad \mathbb{P}(S_t=0)>0  \text{if $t$ is even}
\]
Hence:
\[
\mathbb{P}(X_t<u) \geq  \mathbb{P}(S_t<0)+\mathbb{P}(S_t=0)
\]
and therefore:
\[
F_t(u) \geq  \tfrac12  \text{for all $t$, with } F_t(u)>\tfrac12  \text{whenever $t$ is even}
\]

Intuitively, every pathway that is not strictly dominated by $+$ updates finishes strictly left of $u$. At even horizons there is extra mass from $S_t=0$, making the inequality strict. 

\begin{proposition}[Two–step quantitative contraction]\label{prop:flat:two-step}
Consider $X_{t+1}=X_t+\xi_t\alpha(X_t)$ with $\xi_t\in\{-1,+1\}$ i.i.d.\ symmetric and $\alpha$ strictly increasing. Fix a starting point $X_0=u$. Then there exists $\varepsilon_*>0$ such that for all $\varepsilon\in(0,\varepsilon_*)$,
\[
    F_2(u-\varepsilon) \;\ge\; \tfrac34 \;>\; \tfrac12 \,.
\]
In particular, three of the four two–step sign patterns land strictly below $u-\varepsilon$.
\end{proposition}

\begin{proof}
Enumerate the four patterns and compute $X_2$ exactly:
\[
(++):\quad X_2=u+\alpha(u)+\alpha\bigl(u+\alpha(u)\bigr)
\]
\[
(+-):\quad X_2=u+\alpha(u)-\alpha\bigl(u+\alpha(u)\bigr)
\]
\[
(-+):\quad X_2=u-\alpha(u)+\alpha\bigl(u-\alpha(u)\bigr)
\]
\[
(--):\quad X_2=u-\alpha(u)-\alpha\bigl(u-\alpha(u)\bigr)
\]
Define the positive gaps as:
\[
\delta_1 \coloneqq \alpha\bigl(u+\alpha(u)\bigr)-\alpha(u) > 0
\]
\[
\delta_2 \coloneqq \alpha(u)-\alpha\bigl(u-\alpha(u)\bigr) > 0
\]
\[
\delta_3 \coloneqq \alpha(u)+\alpha\bigl(u-\alpha(u)\bigr) > 0
\]
For any $\varepsilon\in\bigl(0,\varepsilon_*\bigr)$ with $\varepsilon_*\coloneqq \min\{\delta_1,\delta_2,\delta_3\}$ we have:
\[
(+-): X_2=u-\delta_1<u-\varepsilon,\quad (-+): X_2=u-\delta_2<u-\varepsilon,\quad (--): X_2=u-\delta_3<u-\varepsilon
\]
Hence $F_2(u-\varepsilon)\geq 3/4$.
\end{proof}

Intuitively, in $(+-)$ the right move is undone from a point where $\alpha$ is larger, producing a deficit of size $\delta_1$. In $(-+)$ the left move is only partially recovered from a point where $\alpha$ is smaller, giving deficit $\delta_2$. The double left $--$ accumulates a larger drop $\delta_3$. Only $(++)$ is to the right of $u$, so three out of four outcomes fall below $u-\varepsilon$ for small $\varepsilon$.

\paragraph{Uniform shift to $u-\varepsilon$ at general even times}

Pathwise strict negativity at $u$ does not by itself yield a uniform shift to $u-\varepsilon$ for fixed $\varepsilon>0$: the negative drift on $S_t=0$ can be arbitrarily small if $\alpha$ flattens along the visited states. A quantitative modulus on the local growth of $\alpha$ provides a sufficient condition.

\begin{proposition}[Uniform gap with global lower bounds]\label{prop:flat:uniform-gap-global}
Let $t\in 2\mathbb{N}$. From Condition~\ref{cond:flat:standing}, for constants $c_0$, $c_1$, $\alpha$ satisfies:
\[
    \alpha'(x) \geq c_1 > 0 \quad\text{and}\quad \alpha(x) \geq c_0 > 0 \quad\text{for all } x\in\mathbb{R} 
\]
Then, for any trajectory:
\[
    S_t = 0
    \Longrightarrow
    X_t \leq u - \frac{t}{2}c_1c_0,
\]
\[
    S_t < 0 \Longrightarrow X_t \leq u - \frac{t}{2}c_1c_0 - c_0\cdot |S_t| \quad \text{(at least $c_0$ extra drop per unmatched $-$)}
\]
Consequently, for any $\varepsilon \in \bigl(0,\tfrac{t}{2}c_1c_0\bigr)$:
\[
    F_t(u-\varepsilon) \geq \mathbb{P}\bigl(S_t \leq 0 \bigr) = \tfrac12 + \tfrac12\mathbb{P}(S_t = 0) > \tfrac12 
\]
\end{proposition}

\begin{proof}
Pair the $t$ signs greedily into adjacent opposite pairs. When $S_t<0$ there are exactly $|S_t|$ unmatched ``$-$'' signs.

For any adjacent pair starting at state $x$, the Mean Value Theorem with $\alpha' \geq c_1$ gives, for $h>0$:
\[
    \alpha(x+h) - \alpha(x) \geq c_1 h, \quad \alpha(x) - \alpha(x-h) \geq c_1 h 
\]

\noindent
\textbf{Case $(+,-)$:} The increment equals:
\[
    \alpha(x) - \alpha(x+\alpha(x)) \leq -c_1\alpha(x) \leq -c_1c_0
\]

\noindent
\textbf{Case $(-,+)$:} Let $y = x - \alpha(x)$. The increment is:
\[
    -\alpha(x) + \alpha(y) = -\big(\alpha(x) - \alpha(y)\big) \leq -c_1(x - y) = -c_1\alpha(x) \leq -c_1c_0 
\]
Summing over the $t/2$ pairs yields:
\[
    X_t - u \leq -\frac{t}{2}c_1c_0 \quad    \text{when } S_t = 0 
\]
When $S_t < 0$, after pairing, there remain $|S_t|$ unmatched ``$-$'' signs. Each unmatched step at state $z$ contributes at most:
\[
    -\alpha(z) \leq -c_0 
\]
Thus:
\[
    X_t \leq u - \frac{t}{2}c_1c_0 - c_0 |S_t| 
\]
which proves the claimed uniform bound. Finally, since $t$ is even, $\mathbb{P}(S_t \leq 0) = \tfrac12 + \tfrac12 \mathbb{P}(S_t=0) > \tfrac12$, giving the lower bound for $F_t(u-\varepsilon)$.
\end{proof}

In this proof, the lower bounds on the derivative $\alpha' \geq c_1 > 0$ ensures that each adjacent cancellation produces a strictly negative decrement proportional to the step size, and the positive floor $\alpha \geq c_0 > 0$ prevents steps from degenerating. On $S_t=0$, there are exactly $t/2$ such cancellations, so the cumulative drop is at least $(t/2) c_0 c_1$. Any negative imbalance ($S_t<0$) only increases the downward drift by at least $c_0$ per unmatched $-$.

The combination of a positive slope $c_1>0$ and a positive floor $c_0>0$ is a strong sufficient condition for a uniform contraction gap. It makes the drift estimate completely path-independent and eliminates the need for confinement to a local window. Without such global regularity, one still has contraction in distribution (Theorem~\ref{thm:flat:point-contraction}), but only the one-sided bound $F_t(u)>\tfrac12$ can be guaranteed in general, along with the sharp two-step estimate of Proposition~\ref{prop:flat:two-step}. 

Finally, for even $t$, one always has $F_t(u)>\tfrac12$. Under the quantitative modulus of Proposition~\ref{prop:flat:uniform-gap-global}, this strengthens to:
\[
F_t(u-\varepsilon)>\tfrac12
\quad\text{for any}\quad
\varepsilon\in\bigl(0,\tfrac{t}{2} c_0 c_1\bigr)
\]




\subsection{Proof of Mass-Conditions for Median Contraction, Corollary~\ref{cor:flat:mass-conditions-median}}
\label{app:flat:mass-conditions-median-proof} 

We restate the updates: 
\[ 
    X_{t+1}=X_t+\xi_t \alpha(X_t),\quad \xi_t\in\{-1,+1\} \text{i.i.d. symmetric} 
\] 

\subsubsection*{Main Proof}

\begin{proof}
From Condition~\ref{cond:flat:standing}, $\alpha$ is finite and strictly increasing. Let $m_0$ be a median of $X_0$. Define the one–sided values: 
\[ 
    a^-:=\lim_{x\uparrow m_0}\alpha(x),\quad a^+:=\lim_{x\downarrow m_0}\alpha(x) \,,
\] 
which exist by the strict increasing property and satisfies $a^- < a^+$. Use the standard shorthand: 
\[ 
    T_+(x):=x+\alpha(x),\quad T_-(x):=x-\alpha(x) 
\] 
We analyze the two–step update $X_2=T_{\xi_1}\circ T_{\xi_0}(X_0)$ and bound $\Pr(X_2\leq m_0\mid X_0)$ on four regions: 
\[ 
\begin{aligned} 
A&:=(-\infty,m_0-2a^-] \quad &B&:=(m_0-2a^-, m_0] \\
C&:=(m_0, m_0+2a^+] \quad &D&:=(m_0+2a^+, \infty) 
\end{aligned} 
\] 

\paragraph{Case A: $X_0\leq m_0-2a^-$} 
Since $X_0\leq m_0$, we have $\alpha(X_0)\leq a^-$. Thus: 
\[ 
T_+(X_0)=X_0+\alpha(X_0) \leq  m_0-2a^- + a^- = m_0-a^- \leq  m_0 
\] 
At $T_+(X_0)\leq m_0$ the step size remains $\leq a^-$, so a second right step adds at most $a^-$. Hence: 
\[ 
    T_+\big(T_+(X_0)\big) \leq  (m_0-a^-)+a^- = m_0 
\] 
Moreover $T_-(x)\leq T_+(x)$ for all $x$, and both $T_\pm$ are increasing therefore for any two–step pattern: 
\[ 
    T_-\circ T_- \leq  T_-\circ T_+ \leq  T_+\circ T_+ \quad T_+\circ T_- \leq  T_+\circ T_+ 
\] 
so if $T_+\circ T_+(X_0)\leq m_0$ then \emph{all} patterns end $\leq m_0$. Thus: 
\[ 
    \Pr(X_2\leq m_0\mid X_0\in A)=1 
\] 

\paragraph{Case B: $m_0-2a^-<X_0\leq m_0$} 
If the first step is $-$, then $T_-(X_0)=X_0-\alpha(X_0)<X_0\leq m_0$, hence already below $m_0$ and the second step cannot cross above $m_0$. If the first step is $+$, then: 
\[ 
    T_-\big(T_+(X_0)\big) = X_0+\alpha(X_0)-\alpha\big(X_0+\alpha(X_0)\big)  \leq  X_0 \leq  m_0 
\] 
because $\alpha$ is increasing, so $\alpha(X_0+\alpha(X_0))\geq \alpha(X_0)$. Therefore every two–step pattern except $++$ lands $\leq m_0$, and: 
\[ 
    \Pr(X_2\leq m_0\mid X_0\in B) \geq  \tfrac34 
\] 

\paragraph{Case C: $m_0< X_0\leq m_0+2a^+$} 
Here $\alpha(X_0)\geq a^+$. Two left steps suffice: 
\[ 
    T_-\big(T_-(X_0)\big) = X_0-\alpha(X_0)-\alpha\big(X_0-\alpha(X_0)\big)  \leq  X_0-a^+-a^+ \leq  m_0 
\] 
Hence: 
\[ 
    \Pr(X_2\leq m_0\mid X_0\in C) \geq  \Pr(\xi_0=\xi_1=-1) = \tfrac14 
\] 

\paragraph{Case D: $X_0> m_0+2a^+$} The trivial lower bound $\Pr(X_2\leq m_0\mid X_0\in D)\geq 0$ suffices. 

\medskip 
\paragraph{Aggregation} Averaging over the partition gives: 
\[
  F_2(m_0)=\Pr(X_2\le m_0) 
  \;\ge\;
  \Pr(X_0\in A)\cdot 1
  + \Pr(X_0\in B)\cdot \tfrac34
  + \Pr(X_0\in C)\cdot \tfrac14
  + \Pr(X_0\in D)\cdot 0 \,.
\]
\end{proof}


\subsection{Proof of Upper-Quantile Contraction, Theorem~\ref{thm:flat:quantile-drift}} \label{app:flat:quantile-drift-proof}

We restate the stochastic updates (as in Equation~\ref{eqn:flat:X-updates}):
\[
    X_{t+1}=X_t+\xi_t \alpha(X_t),\quad \xi_t\in\{-1,+1\} \text{i.i.d. symmetric}    
\]
Use the standard shorthand: 
\[ 
    T_+(x):=x+\alpha(x),\quad T_-(x):=x-\alpha(x) 
\] 
and write the two-step branch maps:
\[
\begin{aligned}
T_{++}(x)&=x+\alpha(x)+\alpha(x+\alpha(x))\\
T_{+-}(x)&=x+\alpha(x)-\alpha(x+\alpha(x))\\
T_{-+}(x)&=x-\alpha(x)+\alpha(x-\alpha(x))\\
T_{--}(x)&=x-\alpha(x)-\alpha(x-\alpha(x))
\end{aligned}
\]
Using the definition of $K$, we write the \emph{two-step kernel}:
\[
    K_2(x, A):=\Pr\!\big(T_{\xi_2\xi_1}(x)\in A\big)
     = \tfrac14\sum_{a,b\in\{+,-\}}\mathbf 1\!\big\{T_{ab}(x)\in A\big\}
\]

\subsubsection*{Main Proof}
\begin{proof}
\textit{Pointwise two-step contraction.}
Fix $x\in\mathbb R$. By the mean value theorem and $\alpha'(u)\geq \alpha'_{\min}$ on intervals of length $\alpha(x)$, we have:
\[
\alpha(x+\alpha(x))-\alpha(x) \geq  \alpha'_{\min} \alpha(x),\quad
\alpha(x)-\alpha(x-\alpha(x)) \geq  \alpha'_{\min} \alpha(x)
\]
Hence:
\[
T_{+-}(x) \leq  x-\alpha'_{\min} \alpha(x) \quad T_{-+}(x) \leq  x-\alpha'_{\min} \alpha(x)
\]
and, using $\alpha\geq \alpha_{\min}$, we get:
\[
T_{--}(x) \leq  x-\alpha(x)-\alpha(x-\alpha(x)) \leq  x-2\alpha_{\min} \leq  x-\alpha'_{\min} \alpha_{\min}
\]
Since the four branches have equal probability $1/4$:
\[
\Pr\big\{X_{t+2}\leq x-\alpha'_{\min} \alpha(x) \big| X_t=x\big\}  \geq  \Pr\{+-\}+\Pr\{-+\} = \tfrac12
\]
Therefore a median of the two-step law from $x$ satisfies:
\[
M_{t+2}(x) \leq  x-\alpha'_{\min} \alpha(x)
\]
Applying this with $x$ replaced by $M_t(x)$ and using $\alpha\geq \alpha_{\min}$ gives:
\begin{equation}\label{eqn:flat:per-ancestor-step-t}
M_{t+2}(x) \leq  M_t(x) - \alpha'_{\min} \alpha\big(M_t(x)\big)  \leq  M_t(x) - \alpha'_{\min} \alpha_{\min}  = M_t(x) - c
\end{equation}

\noindent
Now we show our three main claims in Theorem~\ref{thm:flat:quantile-drift}. 
\begin{enumerate}
    \item \textbf{Strict decay:} Taking the essential supremum over $x$ in Equation~\ref{eqn:flat:per-ancestor-step-t} yields $Z_{t+2}\leq Z_t-c$. Inductively, $Z_t\leq Z_0 - (t/2)c \leq b - (t/2)c$. 
    \item \textbf{Envelope dominates the median: }
    By definition of $M_t(x)$, $K_2^{* (t/2)}\big(x,(-\infty,M_t(x)]\big)\geq \tfrac12$ for each $x$. 
    For the $t$-step mixture $\mu_t(\cdot) \coloneqq \int K_2^{* (t/2)}(x,\cdot) \mu_0(dx)$ we have:
    \[
    F_t(Z_t) =\int K_2^{* (t/2)}\big(x,(-\infty,Z_t]\big) \mu_0(dx) \geq \int \tfrac12 \mu_0(dx)
     = \tfrac12
    \]
    so any even-time median $m_t$ satisfies $m_t\leq Z_t$.
    \item \textbf{Finite crossing and perpetual decay: }
    Let $N=\lceil(b-m_0)/c\rceil$. Then $Z_{2N}\leq b-Nc\leq m_0$, so $F_{2N}(m_0)\geq \tfrac12$.
    For even $t\geq 2N$, repeated application of (i) gives
    $Z_t\leq Z_{2N} - \frac{t-2N}{2} c \leq m_0 - \frac{t-2N}{2} c$, proving the claim.
    
\end{enumerate}
\end{proof}


\subsection{Proof of Strict Edge Descent, Lemma~\ref{lem:flat:edge-strict-descent}} \label{app:flat:strict-descent-proof}

We restate the stochastic updates (as in Equation~\ref{eqn:flat:X-updates}):
\[
    X_{t+1}=X_t+\xi_t \alpha(X_t),\quad \xi_t\in\{-1,+1\} \; \text{i.i.d. symmetric}    
\]

Our results in this proof appendix consider the setting of Lemma~\ref{lem:flat:edge-strict-descent}, and we omit explicit consideration in our subsequent statements. 

\subsubsection*{Auxiliary Results}
We first collect several intermediate lemmas used in the proof. These lemmas are also used in the proof of Lemma~\ref{lem:flat:clt-descent}, found in Appendix~\ref{app:flat:clt-descent-proof}. 

\begin{lemma}[Order preservation and preimages]\label{lem:flat:monotone-branches}

Consider Condition~\ref{cond:flat:alpha-regularity} with $0<\alpha'(x)\leq L_\alpha<1$ on the visited range. Then $T_\pm(x)=x\pm\alpha(x)$ are $C^1$ and strictly increasing, with:
\[
T_\pm'(x)=1\pm \alpha'(x)\in [1-L_\alpha,1+L_\alpha]
\]
Consequently, for any $t\geq 1$ and sign string $s\in\{+,-\}^t$, the composition $T_s:=T_{s_t}\circ\cdots\circ T_{s_1}$ is strictly increasing.
For $w>0$ and $x\in\mathbb{R}$ the pre-image:
\[
I_s(x):=T_s^{-1}([x,x+w])
\]
is a single interval $[a_s(x),b_s(x)]$ with $a_s,b_s$ increasing in $x$.
\end{lemma}

\begin{proof}
By Condition~\ref{cond:flat:alpha-regularity}, $0 < \alpha'\leq L_\alpha<1$, hence $T_\pm'(x)=1\pm\alpha'(x)>0$, so $T_\pm$ are strictly increasing; compositions of strictly increasing maps are strictly increasing. If $T_s$ is strictly increasing, $T_s^{-1}$ exists and is strictly increasing. 

Since $0 \leq \alpha'(x) \leq L_\alpha < 1$, we have the linear bounds:
\[
  (1-L_\alpha)x - C \leq T_\pm(x) \leq (1+L_\alpha)x + C
\]
for some constant $C > 0$, so that $\lim_{x \to \pm\infty} T_\pm(x) = \pm\infty$. 

Thus each branch map $T_s$ is a strictly increasing bijection $\mathbb{R} \to \mathbb{R}$, and the inverse $T_s^{-1}$ is well-defined for all real values. 

The inverse image of a compact interval is a compact interval, and the endpoints $a_s(x)=T_s^{-1}(x)$, $b_s(x)=T_s^{-1}(x+w)$ are increasing in $x$ by monotonicity of $T_s^{-1}$. 
\end{proof}

\begin{lemma}[Monotone integral comparison]\label{lem:flat:mono-int}
Let $f$ be nonincreasing on $[m,\infty)$. If $J_1=[u_1,v_1]$ and $J_2=[u_2,v_2]$ are intervals with
$u_2\geq u_1$, $v_2\geq v_1$, and $|J_2|=|J_1|$, then:
\[
\int_{J_2} f(w) dw  \leq  \int_{J_1} f(w) dw \,,
\]
where $w$ is a dummy variable for integration. 
If $f$ has no plateau on any interval of positive length, the inequality is strict whenever $J_2\neq J_1$.
A symmetric statement holds on $(-\infty,m]$ for $f$ nondecreasing.
\end{lemma}

\begin{proof}
Write:
\[
v_i=u_i+\ell
\]
with common length $\ell>0$. 

For $t\mapsto \Phi(t):=\int_{t}^{t+\ell} f(u) du$ we have:
\[
\Phi'(t)=f(t+\ell)-f(t)\leq 0
\]
by the monotonicity of $f$, hence $\Phi$ is nonincreasing, which gives the claim. 

Strictness follows as $\Phi'$ is $<0$ wherever $f$ is not constant on $(t,t+\ell)$.
\end{proof}

\subsubsection*{Main Proof}
With the auxiliary lemmas in place, we proceed to the proof of Lemma~\ref{lem:flat:edge-strict-descent}.
\begin{proof}

Assume Conditions~\ref{cond:flat:unimodal} and \ref{cond:flat:alpha-regularity} with $0<\alpha'(x)\leq L_\alpha<1$ on the visited range; fix $t\geq 1$ and $w>0$. Let $m^*$ be a mode of $f_0$, so that $f_0$ is nonincreasing on $[m^*,\infty)$ and nondecreasing on $(-\infty,m^*]$.

\medskip
\noindent
\textbf{Right edge:}

Set $y_{+}(t):=T_{+\cdots+}(m^*)$. For any branch $s$, by Lemma~\ref{lem:flat:monotone-branches} each $T_s$ is strictly increasing and satisfies:
\[
T_s(m^*)\leq T_{+\cdots+}(m^*)=y_{+}(t)
\]
because $T_-(x)\leq T_+(x)$ pointwise and increasing compositions preserve order.
Fix $x\geq y_{+}(t)$. Then, for every branch $s$, we have:
\[
I_s(x)=T_s^{-1}([x,x+w])=[a_s(x),b_s(x)]\subset [m^*,\infty)
\]
since $a_s(x)\geq T_s^{-1}(y_{+}(t))\geq m^*$ by monotonicity of $T_s^{-1}$. Moreover $a_s(\cdot),b_s(\cdot)$ are increasing by Lemma~\ref{lem:flat:monotone-branches}. Because $f_0$ is nonincreasing on $[m^*,\infty)$, Lemma~\ref{lem:flat:mono-int} yields:
\[
x\mapsto \int_{I_s(x)} f_0(u)du \quad \text{is nonincreasing on $[y_{+}(t),\infty)$ for each $s$} \,,
\]
where $u$ is a dummy variable for integration. There is strict decrease wherever $I_s$ slides through a set on which $f_0$ is not a.e. constant. Averaging over branches:
\[
    \hat f_t^{(w)}(x)=\frac{1}{w 2^t}\sum_{s}\int_{I_s(x)} f_0(u)du
\]
is nonincreasing on $[y_{+}(t),\infty)$. If $f_0$ has no plateau on any interval of positive length, the set of $x$ where all branch contributions are non-strict has measure zero, so the decrease is strict for a.e. $x$.

The last step is strict for a.e. $x$ under the no-plateau assumption. Recall that if $f$ is monotone and not constant on any interval of positive length, then for the sliding integral:
\[
  \Phi(t) := \int_t^{t+\ell} f(u)du
\]
we have that $\Phi(t)$ is strictly decreasing for almost every $t$. Indeed, if $\Phi$ failed to decrease strictly on a set of positive measure, then $f$ would be essentially constant on a positive-length interval, contradicting the assumption of \emph{no positive-length plateaus}. Applying this reasoning to each branch interval $I_s(x)$ (which slides strictly right as $x$ increases) and then averaging over the finite set of branches yields the desired strict descent almost everywhere.

\medskip
\noindent
\textbf{Left edge:}

Define $y_{-}(t):=T_{-\cdots-}(m^*)$. A symmetric argument shows that for $x\leq y_{-}(t)-w$ all branch pre-images lie in $(-\infty,m^*]$ and $\int_{I_s(x)} f_0$ is nondecreasing in $x$ by Lemma~\ref{lem:flat:mono-int} applied to a nondecreasing function. Averaging gives that $\hat f_t^{(w)}$ is nondecreasing on $(-\infty,y_{-}(t)-w]$. 

\medskip
\noindent
\textbf{Width bound:}

Let $\Delta_k := y^{(k)}_{+} - y^{(k)}_{-}$. 
From the update rules, we get:
\[
y_{k+1}^{\max} = y^{(k)}_{+} + \alpha\bigl(y^{(k)}_{+}\bigr), 
\quad 
y_{k+1}^{\min} = y^{(k)}_{-} - \alpha\bigl(y^{(k)}_{-}\bigr)
\]
we obtain:
\[
\Delta_{k+1} = y_{k+1}^{\max} - y_{k+1}^{\min} = \bigl(y^{(k)}_{+} - y^{(k)}_{-}\bigr)   + \alpha\bigl(y^{(k)}_{+}\bigr) + \alpha\bigl(y^{(k)}_{-}\bigr) = \Delta_k  + \alpha\bigl(y^{(k)}_{+}\bigr) + \alpha\bigl(y^{(k)}_{-}\bigr)
\]
Since $\alpha$ is positive, we can bound:
\[
\Delta_{k+1} \leq \Delta_k + 2 \sup_{x \in [y^{(k)}_{-},y^{(k)}_{+}]} \alpha(x)
\]
Summing for $k = 0, 1, \ldots, t-1$ (with $\Delta_0 = 0$) yields:
\[
y_{+}(t) - y_{-}(t) = \Delta_t \leq 2 \sum_{k=0}^{t-1} \sup_{x \in [y^{(k)}_{-},y^{(k)}_{+}]} \alpha(x) \leq 2 t  \alpha_{\max} 
\]
where:
\[
  \alpha_{\max} := \sup_{x \in \mathbb{R}} \alpha(x)
\]
is the global supremum of $\alpha$. This completes the proof.
\end{proof}


\subsection{Proof of CLT-Controlled Descent, Lemma~\ref{lem:flat:clt-descent}} \label{app:flat:clt-descent-proof}

We restate the stochastic updates (as in Equation~\ref{eqn:flat:X-updates}):
\[
    X_{t+1}=X_t+\xi_t \alpha(X_t),\quad \xi_t\in\{-1,+1\} \; \text{i.i.d. symmetric}    \,.
\]

\noindent
Assume Conditions~\ref{cond:flat:unimodal} and \ref{cond:flat:alpha-regularity}. Fix $t\geq 1$, $w>0$, $\kappa \geq 1$. Let $m^*$ be the mode of $f_0$ and let $\mathcal N$ be a neighborhood of $m^*$ such that:
\begin{enumerate}
    \item $\alpha$ is bounded on $\mathcal N$ by $\alpha_{\rm loc}$
    \item $\mathcal N$ is large enough that $m^*\pm \alpha_{\rm loc} \kappa \sqrt{t}\in \mathcal N$. One can take $\mathcal N=[m^*-R,m^*+R]$ with any $R\geq \alpha_{\rm loc} \kappa \sqrt{t}$.
\end{enumerate}

\noindent
Additionally, define the CLT half-width:
\[
    W_t(\kappa):=\frac{2}{1-L_\alpha}  \alpha_{\text{loc}}  \kappa \sqrt{t} \,,
\]
and the CLT band:
\[
  B_t(\kappa) := \bigl[m^* - W_t(\kappa) - w, m^* + W_t(\kappa)\bigr] \,.
\]
Our results in this proof appendix consider the setting of Lemma~\ref{lem:flat:clt-descent}, and we omit explicit consideration in our subsequent statements. 

\subsubsection*{Auxiliary Results}
Here, we derive a discrete version of Gronwall's inequality \cite{Gronwall1919}, which is used to bound recursive processes similar in structure to our sign-strings. We then prove an intermediate lemma to bound movements for `typical' sign-strings near the peak, where the previous shape Lemma~\ref{lem:flat:edge-strict-descent} does not apply. 

The main proof also uses Lemma~\ref{lem:flat:monotone-branches} and Lemma~\ref{lem:flat:mono-int} stated and proven in Appendix~\ref{app:flat:strict-descent-proof}. 

\begin{lemma}[Discrete Grönwall's inequality]
\label{lem:flat:discrete-gronwall}
Let $(p_k)_{k\geq 0}$ be a nonnegative sequence with $p_0=0$. 
Suppose there exist $L\in[0,1)$ and a nondecreasing sequence $(A_k)_{k\geq 1}$ such that:
\begin{equation}\label{eqn:flat:DG-hyp}
p_k \leq A_k + L \sum_{j=0}^{k-1} p_j,
\quad k\geq 1 \,.
\end{equation}
Then, for every $k\geq 1$, we have:
\begin{equation}\label{eqn:flat:DG-conc}
    p_k \leq \frac{A_k}{1-L} \,.
\end{equation}
Moreover, the finer bound 
\[
    p_k \leq A_k\frac{1-L^k}{1-L}
\]
also holds.
\end{lemma}

\begin{proof}
Fix $k\geq 1$ and set $A:=A_k$. Since $(A_j)$ is nondecreasing, we have $A_j\leq A$ for all $0\leq j\leq k$. 

Define a comparison sequence $(q_j)_{j=0}^k$ by $q_0=0$ and:
\[
    q_j = A + L \sum_{i=0}^{j-1} q_i \quad (1\leq j\leq k)
\]
By monotonicity of the recursion, Equation~\ref{eqn:flat:DG-hyp} implies $p_j \leq q_j$ for all $j\leq k$. 

We now solve the $q_j$ recursion explicitly by induction. If
$q_1=A$, and if $q_i = A\sum_{\ell=0}^{i-1} L^\ell$ for $i<j$, then:
\[
    q_j = A + L\sum_{i=1}^{j-1} q_i = A + A L \sum_{i=1}^{j-1}\sum_{\ell=0}^{i-1} L^\ell = A \sum_{\ell=0}^{j-1} L^\ell = A\frac{1-L^j}{1-L} \,.
\]
Hence $p_k \leq q_k = A\frac{1-L^k}{1-L} \leq \frac{A}{1-L}$, which gives Equation~\ref{eqn:flat:DG-conc}.
\end{proof}

\begin{corollary}[Localization bound]
\label{cor:flat:localisation-clt}
Let $d_k:=x_k-m^*$ and $u_k:=\max_{0\leq j\leq k}|d_j|$. 
Then, for each sign string $s$:
\[
    u_k \leq \frac{\alpha(m^*)}{1-L_\alpha} M_k(s) \leq \frac{\alpha_{\rm loc}}{1-L_\alpha} M_k(s) \quad k\geq 1
\]
where $M_k(s):=\max_{1\leq j\leq k} |S_j(s)|$ and $\alpha_{\rm loc}:=\sup_{x\in\mathcal N}\alpha(x)$.
\end{corollary}

\begin{proof}
From:
\[
    d_k=\sum_{j=1}^k \varepsilon_j \alpha(x_{j-1}) = \alpha(m^*) S_k(s)+\sum_{j=1}^k \varepsilon_j(\alpha(x_{j-1})-\alpha(m^*))
\]
and $|\alpha(x_{j-1})-\alpha(m^*)|\leq L|d_{j-1}|$ we obtain:
\[
    |d_k| \leq \alpha(m^*)M_k(s)+L\sum_{j=1}^{k-1}|d_j| \leq \alpha(m^*)M_k(s)+L\sum_{j=0}^{k-1}u_j \,.
\]
Taking maxima gives:
\[
    u_k \leq A_k + L\sum_{j=0}^{k-1}u_j \,,
\]
with the nondecreasing term $A_k:=\alpha(m^*)M_k(s)$. 

Apply Lemma~\ref{lem:flat:discrete-gronwall} to get the desired statement.
\end{proof}

\subsubsection*{Main Proof}
\begin{proof}
Set $Y:=\{s:\,M_t(s)\le \kappa\sqrt t\}$ and $\phi_s(x):=\tfrac1w\int_{I_s(x)} f_0$ with $I_s(x)=T_s^{-1}([x,x+w])$.

\medskip
\noindent
\textbf{Step 1 - Localization of typical branches.}
For $x_{k+1}=x_k+\varepsilon_k\alpha(x_k)$ with $x_0=m^*$, write $d_k:=x_k-m^*$. Then:
\[
    d_k=\alpha(m^*)S_k(s)+\sum_{j=1}^k \varepsilon_j\big(\alpha(x_{j-1})-\alpha(m^*)\big)\,,
\]
and $|\alpha(x_{j-1})-\alpha(m^*)|\le L_\alpha |d_{j-1}|$ gives:
\(
    |d_k|\le \alpha(m^*)M_k(s)+L_\alpha\sum_{j=1}^{k-1}|d_j|\,.
\)
With $u_k:=\max_{0\le j\le k}|d_j|$ and Cor.~\ref{cor:flat:localisation-clt}, we have:
\[
    u_k \le \frac{\alpha(m^*)}{1-L_\alpha}M_k(s)\le \frac{\alpha_{\rm loc}}{1-L_\alpha}M_k(s)\,.
\]
Hence for $s\in Y$:
\[
    |T_s(m^*)-m^*|=|d_t|\le \frac{\alpha_{\rm loc}}{1-L_\alpha}\kappa\sqrt t \le W_t(\kappa)\,,
\]
so typical branches remain within $B_t(\kappa)$.

\medskip
\noindent
\textbf{Step 2 - Monotonicity of typical contributions outside the band.}
Fix $x\ge m^*+W_t(\kappa)$. For $s\in Y$, the bound above yields $T_s(m^*)\le m^*+W_t(\kappa)\le x$. Since $T_s$ is strictly increasing (Lemma~\ref{lem:flat:monotone-branches}), $I_s(x)\subset [m^*,\infty)$; by Lemma~\ref{lem:flat:mono-int}, $x\mapsto\int_{I_s(x)} f_0$ is nonincreasing there. Thus $\phi_s$ is nonincreasing on the right of the band. A symmetric argument gives nondecreasing behavior on the left.

\medskip
\noindent
\textbf{Step 3 - Exponentially small contamination.}
Decompose:
\[
    \hat f_t^{(w)}(x)=\sum_{s\in Y}2^{-t}\phi_s(x)\;+\;\sum_{s\in Y^c}2^{-t}\phi_s(x)\,.
\]
For $x_1<x_2$ with $\{x_1,x_2\}\cap B_t(\kappa)=\varnothing$, Step~2 implies:
\(
    \sum_{s\in Y}2^{-t}\big(\phi_s(x_2)-\phi_s(x_1)\big)\le 0
\)
on the right (and $\ge 0$ on the left). Since $0\le \phi_s\le 1/w$:
\[
\hat f_t^{(w)}(x_2)-\hat f_t^{(w)}(x_1)
\le \frac{1}{w}\sum_{s\in Y^c}2^{-t}\,.
\]
By the maximal deviation bound for simple symmetric random walks \cite{durrett2019pte}, we have:
\(
\sum_{s\in Y^c}2^{-t}
=\Pr(\max_{k\le t}|S_k|>\kappa\sqrt t)\le 4e^{-\kappa^2/2}\,.
\)
Therefore:
\[
\hat f_t^{(w)}(x_2)-\hat f_t^{(w)}(x_1)
\le \frac{4}{w}e^{-\kappa^2/2}
\]
on the right (and the reversed inequality on the left), proving the claim.
\end{proof}

\subsection{Proof of Median Dominating Mode, Proposition~\ref{prop:flat:median-dominate-mode}}
\label{app:flat:median-dominate-mode-proof}

We work with the update rule (Equation~\ref{eqn:flat:X-updates})
\[
    X_{t+1}=X_t+\xi_t\alpha(X_t),\quad \xi_t\in\{-1,+1\} \; \text{i.i.d. symmetric} \,.
\]
Assume Conditions~\ref{cond:flat:unimodal} and \ref{cond:flat:alpha-regularity}: $X_0$ has an absolutely continuous unimodal density, and $\alpha\in C^1$ on the visited range with $0\leq \alpha'(x)\leq L_\alpha<1$.
Use the standard shorthand:
\[
    T_+(x):=x+\alpha(x),\quad T_-(x):=x-\alpha(x)
\]
and write the two-step branch maps:
\[
\begin{aligned}
T_{++}(x)&=x+\alpha(x)+\alpha(x+\alpha(x))\\
T_{+-}(x)&=x+\alpha(x)-\alpha(x+\alpha(x))\\
T_{-+}(x)&=x-\alpha(x)+\alpha(x-\alpha(x))\\
T_{--}(x)&=x-\alpha(x)-\alpha(x-\alpha(x))
\end{aligned}
\]
Using the definition of $K$, we write the \emph{two-step kernel}:
\[
    K_2(x, A):=\Pr\!\big(T_{\xi_2\xi_1}(x)\in A\big)
     = \tfrac14\sum_{a,b\in\{+,-\}}\mathbf 1\!\big\{T_{ab}(x)\in A\big\} \,.
\]
Because $0\leq \alpha'<1$, each $T_\pm$ and hence each $T_{ab}$ is strictly increasing; consequently every inverse $T_{ab}^{-1}$ is well-defined, strictly increasing, and $C^1$.

Our results in this proof appendix consider the setting of Proposition~\ref{prop:flat:median-dominate-mode}, and we omit explicit consideration in our subsequent statements. 

\subsubsection*{Auxiliary Results}
Here, we collect intermediate lemmas used in the main proof. We highlight two important intermediate results:

The \emph{straddle} Lemma~\ref{lem:flat:two-step-straddle} positions the `ancestor' mode relative to the four preimage thresholds $a,b,c,d$. 

The \emph{cross-pair balance} Lemma~\ref{lem:flat:cross-balance} shows that $F_{t\mid\hat s}(m_t^*)\geq \tfrac12$, which is the key mode–median link (after averaging over $\hat s$).

\begin{lemma}[Two-step median contraction at a point]\label{lem:flat:two-step-pmc}
For every $x\in\R$:
\[
    K_2\!\big(x,(-\infty,x]\big) \geq \tfrac34
\]
with strict inequality unless $\alpha$ vanishes at $x$, $x\pm\alpha(x)$, and $x\pm\alpha(x)\pm\alpha(x\pm\alpha(x))$.
\end{lemma}

\begin{proof}
Treat $x\in \mathbb{R}$ individually as point-masses, this follows from Theorem~\ref{thm:flat:point-contraction}.
\end{proof}

\begin{lemma}[Monotone two-step kernel]\label{lem:flat:two-step-monotone}
If $x\leq y$ then $K_2(x,\cdot)\leq_{\mathrm{st}}K_2(y,\cdot)$. Equivalently, for every $u\in\R$ the map:
\[
h_u(x):=K_2\!\big(x,(-\infty,u]\big)
\]
is nondecreasing in $x$.
\end{lemma}

\begin{proof}
Fix $u$ and $a,b\in\{+,-\}$. Since $T_{ab}$ is strictly increasing, the indicator $\mathbf 1\{T_{ab}(x)\leq u\}$ is nondecreasing in $x$. Averaging over $(a,b)$ yields that $h_u$ is nondecreasing, i.e. $K_2(x,\cdot)\leq_{\mathrm{st}}K_2(y,\cdot)$ for $x\leq y$.
\end{proof}

\begin{lemma}[Two-step CDF decomposition]\label{lem:flat:two-step-cdf}
Fix even $t$. Condition on a sign prefix $\hat s$ of length $t-2$ and let $Y_{\hat s}:=T_{\hat s}(X_0)$ have density $g_{\hat s}$ and CDF $G_{\hat s}$. Then for every $x$:
\begin{equation}\label{eqn:flat:two-step-cdf}
F_{t\mid \hat s}(x) = \tfrac14\sum_{a,b\in\{+,-\}} G_{\hat s}\!\big(T_{ab}^{-1}(x)\big)
\end{equation}
\end{lemma}

\begin{proof}
Condition on $Y_{\hat s}=y$. Because $T_{ab}$ is increasing, we have:
\[
\Pr(T_{ab}(y)\leq x)=\Pr(y\leq T_{ab}^{-1}(x))=G_{\hat s}(T_{ab}^{-1}(x))
\]
Average over the four equally likely branches to obtain Equation~\ref{eqn:flat:two-step-cdf}.
\end{proof}

\begin{lemma}[Two-step straddle of the ancestor mode]\label{lem:flat:two-step-straddle}
Let $Y_{\hat s}:=T_{\hat s}(X_0)$ have density $g_{\hat s}$ and CDF $G_{\hat s}$. Let $m_t^*$ be a (global) mode of $f_t$ and let $m_{\hat s}^*$ be a mode of $g_{\hat s}$.
Define the four preimages of $m_t^*$ by:
\[
a:=T_{++}^{-1}(m_t^*),\quad
b:=T_{+-}^{-1}(m_t^*),\quad
c:=T_{-+}^{-1}(m_t^*),\quad
d:=T_{--}^{-1}(m_t^*)
\]
Then:
\[
a < m_{\hat s}^* < d \quad\text{and}\quad \min\{b,c\} \leq m_{\hat s}^* \leq \max\{b,c\}
\]
\end{lemma}

\begin{proof}
Because $T_{++}(y)>y$ and $T_{--}(y)<y$ for all $y$, we have $a<m_t^*<d$. If $m_{\hat s}^*\leq a$, then every $T_{ab}^{-1}(m_t^*)\geq a\geq m_{\hat s}^*$; since $g_{\hat s}$ is unimodal and nonincreasing on $[m_{\hat s}^*,\infty)$, for sufficiently small $\varepsilon>0$:
\[
g_{\hat s}\!\big(T_{ab}^{-1}(m_t^*-\varepsilon)\big) \geq g_{\hat s}\!\big(T_{ab}^{-1}(m_t^*)\big) \quad\text{for all }ab
\]
and with $(T_{ab}^{-1})'(x)>0$ this implies $f_{t\mid \hat s}(m_t^*-\varepsilon)\geq f_{t\mid \hat s}(m_t^*)$, contradicting that $m_t^*$ is a mode. Hence $m_{\hat s}^*>a$. A symmetric argument excludes $m_{\hat s}^*\geq d$. Thus $a<m_{\hat s}^*<d$.

For the cross pair, assume w.l.o.g. $b\leq c$. If $m_{\hat s}^*<b$, then by continuity $T_{+-}^{-1}(x)\in[m_{\hat s}^*,b]$ for $x$ near $m_t^*$ on the right; since $g_{\hat s}$ is nondecreasing on $(-\infty,m_{\hat s}^*]$, the $+-$ branch contribution is nondecreasing to the right of $m_t^*$, while the remaining branches cannot all decrease there (otherwise their sum would decrease, contradicting the mode). Hence $m_{\hat s}^*\geq b$. 

A symmetric argument excludes $m_{\hat s}^*>c$. Combining the results gives $b\leq m_{\hat s}^*\leq c$.
\end{proof}

\begin{lemma}[Cross-pair balance]\label{lem:flat:cross-balance}
With $a,b,c,d$ as in Lemma~\ref{lem:flat:two-step-straddle}, let $\ell:=\min\{b,c\}$ and $r:=\max\{b,c\}$. Then for $x^*:=m_t^*$ (the mode):
\[
F_{t\mid\hat s}(x^*) \geq \tfrac12
\]
\end{lemma}

\begin{proof}
Define the branch-count step function:
\[
q(y):=\tfrac14\sum_{a,b\in\{+,-\}}\mathbf 1\!\big\{T_{ab}(y)\leq x^*\big\} =\tfrac14\sum_{a,b}\mathbf 1\!\big\{y\leq T_{ab}^{-1}(x^*)\big\}
\]
Since each $T_{ab}$ is increasing, $q$ is right-continuous and nonincreasing, with jumps of size $1/4$ at the ordered thresholds:
\[
a < \ell \leq r < d
\]
Thus:
\[
q(y)=\begin{cases}
1, & y\leq a\\[2pt]
3/4, & a<y\leq \ell\\[2pt]
1/2, & \ell<y\leq r\\[2pt]
1/4, & r<y\leq d\\[2pt]
0, & y>d
\end{cases}
\]
By Lemma~\ref{lem:flat:two-step-cdf}:
\[
F_{t\mid\hat s}(x^*)=\int_{\R} q(y)g_{\hat s}(y)dy
\]
Write the difference from $\tfrac12$ as:
\[
\int_{\R}\!\Big(q-\tfrac12\Big)g_{\hat s}(u)du =\tfrac12\!\int_{(-\infty,a]} g_{\hat s}(u)du +\tfrac14\!\int_{(a,\ell]} g_{\hat s}(u)du -\tfrac14\!\int_{(r,d]} g_{\hat s}(u)du -\tfrac12\!\int_{(d,\infty)} g_{\hat s}(u)du \,,
\]
where $u$ is a dummy variable for integration. By Lemma~\ref{lem:flat:two-step-straddle}, $m_{\hat s}^*\in[\ell,r]$. Unimodality of $g_{\hat s}$ then gives:
\[
\int_{(-\infty,a]} g_{\hat s}(u)du \geq \int_{(d,\infty)} g_{\hat s}(u)du \quad\text{and}\quad \int_{(a,\ell]} g_{\hat s}(u)du \geq \int_{(r,d]} g_{\hat s}(u)du
\]
because each point to the right of $r$ (resp. in $(r,d]$) can be paired with a point to the left of $\ell$ (resp. in $(a,\ell]$) no farther from $m_{\hat s}^*$, where $g_{\hat s}$ is at least as large.

Combining the two inequalities with the coefficients $1/2$ and $1/4$ shows $\int(q-\tfrac12)g_{\hat s}\geq 0$, i.e. $F_{t\mid\hat s}(x^*)\geq\tfrac12$.
\end{proof}

\subsubsection*{Main Proof}

\begin{proof}
Fix even $t\geq2$ and let $m_t^*$ be a mode of $f_t$. Condition on any prefix $\hat s$ of length $t-2$ and let $a,b,c,d$ be as in Lemma~\ref{lem:flat:two-step-straddle}, that is:
\[
a=T_{++}^{-1}(m_t^*),\quad b=T_{+-}^{-1}(m_t^*),\quad
c=T_{-+}^{-1}(m_t^*),\quad d=T_{--}^{-1}(m_t^*)
\]
By Lemma~\ref{lem:flat:cross-balance} we have $F_{t\mid\hat s}(m_t^*)\geq \tfrac12$. Averaging over $\hat s$ yields $F_t(m_t^*)\geq \tfrac12$, hence every median $m_t$ satisfies $m_t\geq m_t^*$. 

If equality holds, then, by the strict parts of the lemmas above, $f_t$ is flat at the median; otherwise the inequality is strict.
\end{proof}

\subsection{Proof of Median Drift under Unimodal and $\alpha$ $1$-Lipschitz Conditions (Theorem~\ref{thm:flat:median-drift-unimodal})}
\label{app:flat:median-drift-unimodal-proof}

We recall the stochastic recursion:
\[
    X_{t+1} = X_t + \xi_t \alpha(X_t), 
    \quad \xi_t \in \{-1, +1\} \; \text{i.i.d., symmetric.}
\]
Assume that $0 \le \alpha'(x) \le L_\alpha < 1$ on the relevant range and that the distribution of $X_t$ is unimodal.  

Each two-step update can be written in terms of four branch maps:
\[
\begin{aligned}
T_{--}(x) &= x - \alpha(x) - \alpha(x - \alpha(x)), \\
T_{-+}(x) &= x - \alpha(x) + \alpha(x - \alpha(x)), \\
T_{+-}(x) &= x + \alpha(x) - \alpha(x + \alpha(x)), \\
T_{++}(x) &= x + \alpha(x) + \alpha(x + \alpha(x)).
\end{aligned}
\]
All $T_s$ are strictly increasing.

For any random variable $Y$ with CDF $F_Y$ and any increasing map $T$, the transformed variable $T(Y)$ satisfies:
\[
    F_{T(Y)}(y) = F_Y(T^{-1}(y)) \,.
\]
At time $t$, we write $F_t := F_{X_t}$ and $f_t := f_{X_t}$ for the CDF and PDF of $X_t$, and:
\[
    F_{s,t}(y) := F_{T_s(X_t)}(y)
\]
for the CDF of the transformed variable under branch $s$.

\subsubsection*{Auxiliary Lemmas}

\begin{lemma}[Median under increasing transforms]
\label{lem:flat:median-pushforward}
Let $m$ be a median of $X$. If $T:\R\to\R$ is strictly increasing, then $T(m)$ is a median of $T(X)$.
\end{lemma}

\begin{proof}
Since $T$ is increasing, $\{T(X)\le T(m)\} = \{X\le m\}$, hence
$F_{T(X)}(T(m)) = F_X(m) \ge \tfrac12$.
Similarly, $\{T(X)\ge T(m)\} = \{X\ge m\}$, giving the complementary condition.
\end{proof}

\begin{lemma}[Three-branch dominance at the median]
\label{lem:flat:three-branch-dominance}
Suppose $0 < \alpha'(x)\le L_\alpha<1$, and let $f$ be a unimodal density with median $m$. Then
\[
T_{--}(m) < T_{-+}(m) < m,
\qquad
T_{+-}(m) < m.
\]
Let:
\[
z_* := \max\{T_{-+}(m),\,T_{+-}(m)\} < m \,.
\]
Then for each $s \in \{--,-+,+-\}$:
\[
    F_{T_s(X)}(z_*) > \tfrac12 \,.
\]
Moreover, if the mode of $f$ lies at or to the left of $m$, then:
\[
    \frac{1}{4}\sum_{s\in\{--,-+,+-,++\}} F_{T_s(X)}(z_*) > \tfrac12\,.
\]
\end{lemma}

\begin{proof}
Monotonicity of $\alpha$ implies
\[
T_{-+}(m)=m-\alpha(m)+\alpha(m-\alpha(m))\le m,
\quad
T_{+-}(m)=m+\alpha(m)-\alpha(m+\alpha(m))\le m \,,
\]
and clearly $T_{--}(m)\le T_{-+}(m)$.
By Lemma~\ref{lem:flat:median-pushforward}, each $T_s(m)$ is a median of $T_s(X)$.
Thus, $F_{T_s(X)}(z_*)\ge \tfrac12$ for all $s\in\{--,-+,+-\}$ since $z_* \ge T_s(m)$.

For the mixture inequality, note that:
\[
    F_{t+2}(z) = \frac14 \sum_{s\in\{--,-+,+-,++\}} F_{s,t}(z)\,.
\]
At $z=z_*$, the three non-\texttt{++} branches contribute at least $3/8$ in total.
For a unimodal $f$ with mode on or to the left of $m$, standard left--right pairing
(matching each $x>m$ with a point $x'\le m$)
shows that the average contribution over all four branches exceeds $1/2$,
since mass to the left dominates when $z_*<m$.
\end{proof}

\subsubsection*{Main Proof}

\begin{proof}
Let $m_t$ be a median of $X_t$ and define:
\[
    z_* := \max\{T_{-+}(m_t),\,T_{+-}(m_t)\} < m_t\,.
\]
At even times $t$, the law of $X_t$ is unimodal with high probability with mode $m_t^* \le m_t$ (see Lemmas~\ref{lem:flat:edge-strict-descent}, \ref{lem:flat:clt-descent}).
Applying Lemma~\ref{lem:flat:three-branch-dominance} at time $t$ gives:
\[
    F_{t+2}(z_*) 
    = \frac14\sum_{s\in\{--,-+,+-,++\}} F_{s,t}(z_*) 
    \ge \frac12\,.
\]
Hence any median $m_{t+2}$ of $X_{t+2}$ satisfies:
\[
    m_{t+2} \le z_* < m_t \,,
\]
proving part (a) of Theorem~\ref{thm:flat:median-drift-unimodal}.

\medskip

For part (b), fix $r_t>0$ and set:
\[
A_t := \{|X_t - m_t| \le r_t\}, 
\qquad 
p_t := \Pr(A_t),
\qquad
\underline{\alpha}'(m_t) := \inf_{u \in [m_t - R,\, m_t + R]} \alpha'(u) > 0 \,,
\]
for some $R > \alpha(m_t)$ chosen so that:
\[
[x-\alpha(x),\,x+\alpha(x)] \subseteq [m_t-R,\,m_t+R] 
\quad \forall x \in A_t \,.
\]
For $x\in A_t$, a second-order estimate of the two-step map gives:
\[
M_2(x) \le x - \alpha(x)\!\!\inf_{u\in[x-\alpha(x),x+\alpha(x)]}\!\!\alpha'(u)
   \le m_t - \alpha(m_t)\underline{\alpha}'(m_t) + (x - m_t) + o_{r_t}(1)\,,
\]
where $o_{r_t}(1)\to 0$ as $r_t\to 0$ by continuity of $\alpha$.
Define the threshold:
\[
z_b := m_t - p_t\,\alpha(m_t)\underline{\alpha}'(m_t) + o_{r_t}(1)\,.
\]
Evaluating $F_{t+2}(z_b)$ and using the same three-branch dominance argument yields
$F_{t+2}(z_b) \ge \tfrac12$.
Therefore,
\[
m_{t+2} \le z_b 
= m_t - p_t\,\alpha(m_t)\underline{\alpha}'(m_t) + o_{r_t}(1) \,.
\]
Letting $r_t \to 0$ completes the proof of Theorem~\ref{thm:flat:median-drift-unimodal}, part (b).
\end{proof}

\subsection{Numerical Simulation of the Dynamics of $X_t$} \label{sec:flat:xt-sim}

\begin{figure}[h]
  \centering
  \includegraphics[width=0.48\textwidth]{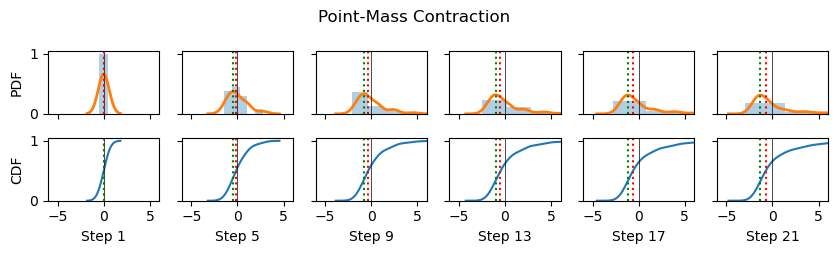}
  \hfill
  \includegraphics[width=0.48\textwidth]{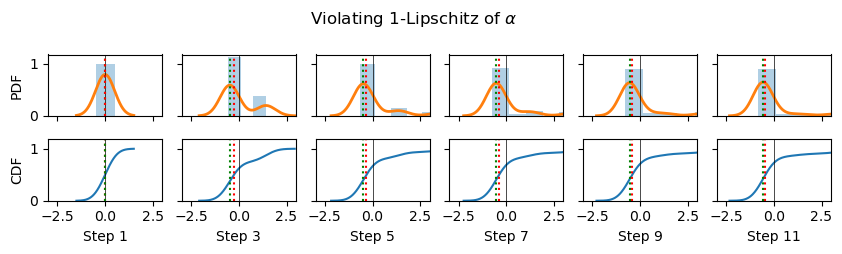}
  \hfill
  \includegraphics[width=0.48\textwidth]{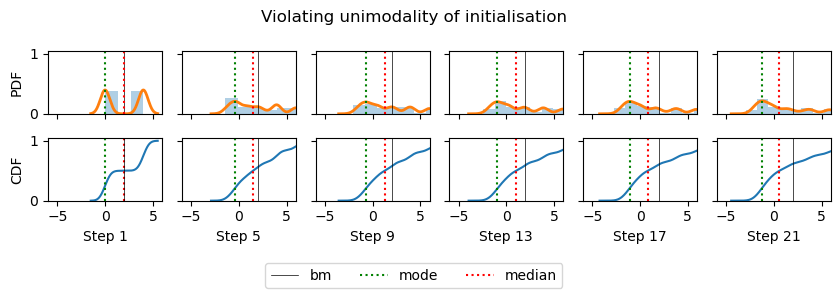}
  
  \caption[Numerical simulation of $X_t$ on a toy example.]{\textbf{Numerical simulation of $X_t$ on a toy example.} \textbf{Top:} Point-Mass Contraction; \textbf{Middle:} Violating $1$-Lipschitz of $\alpha$; \textbf{Bottom:} Violating unimodality of $X_0$; 
  We use $\alpha(x) = \frac{1}{1+e^(-\beta x)}$, where $\beta = 0.01$ for the examples of PMC (top) and violating unimodality (bottom), and $\beta=5$ for the example violating $1$-Lipschitz of $\alpha$. We use a point-mass initialization at $0$; a unimodal initialization at $0$, and a bi-modal initialization at $0, 4$ respectively for the three panes. }
  \label{fig:flat:xt-simulation}
\end{figure}

We complement the theoretical analysis with simulations of a stylized one-dimensional model, illustrating the qualitative behaviors predicted by our results and assessing their dependence on assumptions such as unimodality and the $1$-Lipschitz property of $\alpha$. Figure~\ref{fig:flat:xt-simulation} shows representative trajectories of the empirical density of $X_t$ over time (top: PDFs; bottom: CDFs), estimated via Gaussian kernel density estimation with bandwidth increasing smoothly with~$\alpha$. Results are robust to standard bandwidth rules and to histogram estimators with moderate bin counts. 

Starting from an initial distribution $X_0$ (specified per panel), we iterate the update Equation~\ref{eqn:flat:X-updates} for a fixed number of steps, visualizing every $g$ iterations. At each sampled time~$t$, we record:
\begin{enumerate}
    \item the median $m_t$ (the $0.5$ quantile of the estimated CDF),
    \item the mode $m^*_t$ (the maximizer of the estimated PDF),
    \item the edge spread $Q_{0.95}(t)-Q_{0.05}(t)$.
\end{enumerate}
These statistics test the qualitative predictions of Theorem~\ref{thm:flat:median-drift-unimodal} and the median-mode dominance property (Appendix~\ref{app:flat:median-dominate-mode-proof}).
\paragraph{Case I: Point-mass contraction}
With unimodal $X_0$ and $\alpha$ satisfying Conditions~\ref{cond:flat:standing}, \ref{cond:flat:unimodal}, and \ref{cond:flat:alpha-regularity}, the dynamics follow theoretical expectations:
\begin{itemize}
    \item \textbf{Unimodality persists:} the KDE remains unimodal, consistent with Lemmas~\ref{lem:flat:edge-strict-descent}-\ref{lem:flat:clt-descent}.
    \item \textbf{Median drift:} $m_{t+2}\le m_t$ holds, with strict descent when $\alpha(m_t)>0$, matching Theorem~\ref{thm:flat:median-drift-unimodal}(b).
    \item \textbf{Median--mode relation:} the mode remains to the left of the median ($m^*_t<m_t$), yielding a right-skewed shape.
\end{itemize}
These confirm the predicted persistence of unimodality and right-skew bias.

\paragraph{Case II: Beyond the $1$-Lipschitz regime}
When $\alpha$ violates the $1$-Lipschitz condition, median and mode continue to drift leftward and maintain $m^*_t<m_t$, indicating a robust skew pattern. However, the KDE often becomes multimodal, consistent with the breakdown of the mixing-by-increasing-kernels argument when $\alpha$ varies too sharply.

\paragraph{Case III: Non-unimodal initialization}
For bimodal $X_0$, the median and mode dynamics remain qualitatively consistent with theory---showing leftward drift and median dominance---but the distribution may retain multiple peaks. Thus, unimodality is not ensured without suitable initial or regularity conditions.

\paragraph{Summary}
Across all cases, the simulations corroborate the theoretical characterisations: even-time median descent, median dominance over the mode, and a persistent right-skew signature, while clarifying the limits of unimodality under relaxed assumptions.

\newpage
\section{Using random matrix theory to extend RPE to stochastic minibatches}\label{sec:sgd:fb-sgd-sim}

In this section, we extend the theoretical framework in Section~\ref{ch:flatness} to the non-deterministic setting SGD. Our goal is to understand how curvature noise, introduced by minibatch sampling, modifies the eigenvector dynamics identified in the full-batch deterministic setting. We build a random matrix model for the minibatch Hessian and analyze the resulting eigenvector overlap using tools from spiked matrix theory. This allows us to derive conditions under which RPE continues to hold, and to quantify the degradation introduced by stochasticity. 

\subsection*{Minibatch Hessian as a noisy perturbation}

Let $\mH_{\text{emp}} \in \mathbb{R}^{P \times P}$ denote the empirical Hessian, computed over the full dataset of size $N$. During SGD, we observe a noisy minibatch Hessian, which we model:
\[
    \mH_{\mathrm{mb}} = \frac{1}{B} \sum_{i \in \mathcal{B}_t} \nabla^2 \ell(x_i, \vtheta) \,,
\]
where $\mathcal{B}$ is a minibatch of size $B << N$. In expectation, $\mH_{\mathrm{mb}}$ is unbiased:
\[
\mathbb{E}_{\mathcal{B}}[\mH_{\mathrm{mb}}] = \mH_{\text{emp}}
\]
but its spectrum fluctuates due to finite-sample noise. We follow the decomposition of SGD (see Equation.~\ref{eqn:prelim:sgd-decomp}) and adopt a spiked Wigner model for the Hessian:
\[
    \mH_{\mathrm{mb}} = \mH_{\text{emp}} + \frac{1}{\sqrt{B}} \mXi
\]
where $\mXi$ is now a symmetric noise matrix drawn from a zero-mean (generalized) Wigner ensemble with variance $\sigma^2$ per entry. The zero-mean property follows from the unbiasedness in expectation, while the $1/\sqrt{B}$ scaling reflects the variance reduction with batch size.

\subsection*{Eigenvector overlap from spiked random matrices}

Let $\lambda_1 := \lambda_1(\mH_{\text{emp}})$ denote the top eigenvalue of the empirical Hessian, and let $\vv_1^{\text{emp}}$ and $\vv_1^{\mathrm{mb}}$ be the corresponding eigenvectors associated with the leading eigenvalues of $\mH_{\text{emp}}$ and $\mH_{\mathrm{mb}}$, respectively. We define the squared eigenvector alignment $\gamma$:
\[
    \gamma \;=\; 
    \frac{|\langle \vv_1^{\mathrm{mb}}, \vv_1^{\text{emp}} \rangle|^2}
    {\|\vv_1^{\mathrm{mb}}\|^2 \, \|\vv_1^{\text{emp}}\|^2} \,.
\]
Using results from spiked RMT \cite{granziol2020lr, benaych2010rmt}, we have:
\begin{equation}
\gamma =
\begin{cases}
1 - \dfrac{\sigma^2}{B \lambda_1^2}, & \text{if } \lambda_1 > \dfrac{2\sigma}{\sqrt{B}} \\
0, & \text{otherwise}
\end{cases}
\label{eq:eigv-overlap}
\end{equation}
This expression shows that eigenvector alignment is preserved when the signal eigenvalue $\lambda_1$ of $\mH_{\text{emp}}$ exceeds the stochastic noise floor, which scales as $\sigma / \sqrt{B}$. Below this threshold, the leading eigenvector of the minibatch Hessian becomes statistically indistinguishable from noise.
\begin{tcolorbox}[title=Non-negligible Deviation in Practice.,colback=white,colframe=black]

When $\lambda_1 > 2\sigma/\sqrt{B}$, the correction term $\sigma^2 / (B \lambda_1^2)$ may still be appreciable in practical deep learning settings. For example, a conservative set of values might be:
\[
\lambda_1 \sim 10, \quad \sigma^2 \sim 0.01, \quad B = 128 \quad \Rightarrow \quad \frac{\sigma^2}{B \lambda_1^2} \sim 10^{-2}
\]
$\lambda_1 \sim 10$ is absolutely the lower end of our empirical models, which are simple by design. We expect larger $\lambda_1$s for more complex models, especially transformers \cite{ormaniec2025hessiantransformer}. $\sigma^2 \sim 0.01$ uses reference computations from \textcite{granziol2020lr}. Thus, eigenvector misalignment is empirically upper bounded to $1\%$ in conservative estimates. This indicates that  the direction of maximum curvature remains detectable despite constant small distortions, and this is especially true for directions with sharper curvature profiles (large $\lambda_1$).
\end{tcolorbox}

\subsection*{Extending RPE theory to SGD}

To extend RPE to the stochastic setting, we only require that the leading eigendirection of the minibatch Hessian remains close to that of the empirical Hessian. From the spiked model, the squared overlap between $\vv^{\mathrm{mb}}_1$ and $\vv^{\text{emp}}_1$ is given by Equation~\ref{eq:eigv-overlap}, implying that when $\lambda_1 > 2\sigma/\sqrt{B}$ the minibatch leading eigenvector $\vv^\mathrm{mb}_1$ lies within angle $\delta$ of the `full' leading eigenvector, $\vv^\mathrm{emp}_1$, with $\cos^2\delta = \gamma$.

Linearizing the gradient about $\vtheta^*$ and separating signal from minibatch noise gives:
\[
    \vg_{\mathrm{mb}}(\vtheta_t) \;=\; \mH_{\mathrm{mb}}\,\vr_t \;+\; \vxi_t
    \;=\; \mH_{\text{emp}}\,\vr_t \;+\; \tfrac{1}{\sqrt{B}}\,\mXi_t\,\vr_t \;+\; \vxi_t \,,
\]
where $\vr_t \coloneqq \vtheta_t - \vtheta^*$ and $\vxi_t$ denotes zero-mean additive gradient noise whose variance scales as $1/B$. The SGD update then reads:
\begin{equation}
    \vr_{t+1} \;=\; \big(\mI - \eta \mH_{\text{emp}}\big)\vr_t
    \;-\; \tfrac{\eta}{\sqrt{B}}\,\mXi_t\,\vr_t \;-\; \eta\,\vxi_t \,.
\label{eq:rpe:lin-sgd}
\end{equation}

\noindent
Projecting Equation~\ref{eq:rpe:lin-sgd} onto the top eigenspace of $\mH_{\text{emp}}$ shows that the stochastic perturbations do not change the identity of the dominant direction, but merely attenuate it by the overlap factor $\sqrt{\gamma}$. In practice, as shown earlier, $\gamma$ is typically close to one (empirically $\gtrsim 0.99$), so the principal direction of curvature remains stable under minibatching.

Thus, RPE analyzes derived in the deterministic setting extend directly to SGD: the same rotational mechanism applies once $\eta\lambda_1 > 2$, with the caveat that observed dynamics along $\vv^{\mathrm{mb}}_1$ are attenuated in expectation by $\sqrt{\gamma}$. The bounded misalignment guaranteed by the spiked model therefore justifies applying RPE in the stochastic regime without structural modification.

\section{Learning-rate scaling for experiments in Figure~\ref{fig:sgd:fbst-prog-flat}}
\label{app:sgd:scaling}

For the experiments shown in Fig.~\ref{fig:sgd:fbst-prog-flat}, we refer to the empirical setup in Subsec.~\ref{sec:flat:cv-exp}, where stable convergence was achieved at $\eta_\mathrm{low} = 0.021$.  
The learning rates used in the initial SGD regime are numerically smaller than those in GD, but they correspond to comparable effective step sizes under the linear scaling rule~\cite{goyal2017accurate}.  
This rule predicts $\eta \propto B$, where $B$ is the minibatch size, and the relationship flattens as $B$ approaches the full dataset size~\cite{granziol2020flatness}.  
For CIFAR10, with dataset sizes between $5{,}000$ and $50{,}000$, this implies that $\eta_\mathrm{SGD}=0.002$ (with $B=32$) is roughly equivalent in stability scale to $\eta_\mathrm{GD}\approx0.31$.  
Both are thus substantially larger than $\eta_\mathrm{low}=0.021$, ensuring the initial SGD phase lies within the large–learning-rate regime relevant to instability and flattening.

\newpage
\section{Adaptive optimization as preconditioned gradient descent}\label{app:adaptive:precond-gd}

From \textcite{kingma2017adam}, the update for Adam is:
\[
    \vtheta_{t+1} = \vtheta_t - \eta \cdot \frac{\vm_t}{\sqrt{\vq_t} + \epsilon \mI}
\]
where $\vm_t$ and $\vq_t$ are exponential moving averages of the gradient and squared gradient \footnote{We use the notation $\vq$ to denote the $2$nd-order moment estimate, which is traditionally denoted by $\vv$ in the literature, to avoid confusion with the global notation $\vv$ for eigenvectors.}, respectively, and $\epsilon$ is a small constant for numerical stability in division. The updates to $\vm_t$ and $\vq_t$ are controlled by hyperparameters $\beta_1$ and $\beta_2$, respectively, which commonly are set to $\beta_1=0.9$ and $\beta_2=0.99, 0.999$ for Adam. Setting the hyperparameter $b_1=0$, which means $\vm_t=\vg_t$, recovers RMSprop, which can be seen as a special case of Adam. Define:
\[
    \mQ_t = \mathrm{diag} \big((\vv_t+\epsilon)^{-1/2}\big)
\]
where $\mathrm{diag}()$ constructs a diagonal square matrix from the vector inputs. Then the update may be written:
\[
    \vtheta_{t+1} = \vtheta_t - \eta \mQ_t \vm_t
\]
i.e.\ gradient descent (GD) with momentum and a diagonal, time-varying preconditioner. Equivalently, Adam is the steepest descent in the metric $\langle \va,\vb\rangle_{\mQ_t^{-1}}=\va^\top \mQ_t^{-1} \vb$ for arbitrary vectors $\va$, $\vb$. 

Adaptive optimizers, such as RMSprop and Adam, suppress training instabilities along sharp eigen-directions of the Hessian. Let $\mH_t$ denote the empirical Hessian at time $t$. Under adaptive optimizers, the effective Hessian is:
\[
    \mH_{\text{eff}} = \mQ_t^{1/2} \mH_t \mQ_t^{1/2}
\]
Consequently, training instability arises when:
\[
    \eta \lambda_{\max}(\mH_{\text{eff}}) > 2
\]
where $\lambda_{\max}$ computes the largest eigenvalue of a square matrix. 

\begin{figure}[h]
    \centering
    \includegraphics[width=1.0\linewidth]{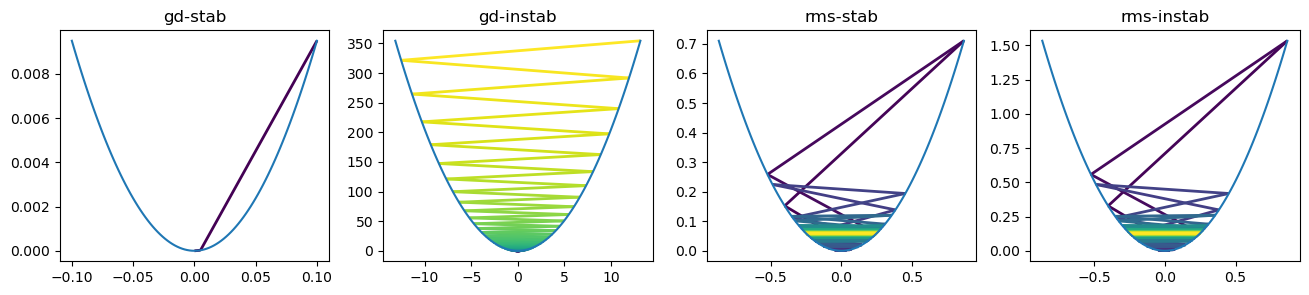}
    \caption[Stable and Unstable training under GD and Adaptive optimizers.]{\textbf{Stable and Unstable training under GD and Adaptive optimizers.} We train a $1$D toy example with quadratic loss under GD and RMSprop, which we use to represent behavior for adaptive optimization, under two different curvature for stable and unstable training. The colormap \emph{viridis} indicates the epochs of training, starting from purple to yellow. We observe \emph{valley jumping} from the \emph{descent lemma} under GD, which are suppressed and stabilized quickly under RMSprop. }
    \label{fig:ada:gd_rms_cartoon}
\end{figure}

\noindent
The contraction can be quantified by:
\[
\eta \lambda_{\max}(\mH_t)\,\min_i (\vq_{t,i}+\epsilon)^{-1/2}
\;\leq\; \eta \lambda_1(\mH_{\text{eff},t})
\;\leq\; \eta \lambda_{\max}(\mH_t)\,\max_i (\vq_{t,i}+\epsilon)^{-1/2}.
\]
Because $\vv_{t,i}$ grows in directions with consistently large gradients, sharp directions are suppressed, and the instability threshold is pushed upwards. Adaptive optimizers therefore damp out oscillatory or divergent modes (including training instabilities) that would otherwise emerge in plain gradient descent. We show a visualization of the different optimizers in a $1$D toy example in Figure~\ref{fig:ada:gd_rms_cartoon}.

\section{Stability Thresholds for Clipped-Ada}
\label{app:adaptive:clipped-ada-deriv}

Recall that adaptive methods such as RMSprop and Adam maintain second-moment estimates $\vq_t = \{p_{i,t}\}$ and apply the preconditioner
\begin{equation}
    \mQ_t = \mathrm{diag}\!\big((\vq_t+\epsilon)^{-1/2}\big).
\end{equation}
For Clipped-Ada, each element is capped by a threshold $q^{\text{thresh}}$:
\begin{equation}
    q^{\text{mod}}_{i,t} = \min(p_{i,t}, q^{\text{thresh}}),
    \qquad
    \mQ^{\text{mod}}_t = \mathrm{diag}\!\big((\vq^{\text{mod}}_t+\epsilon)^{-1/2}\big).
\end{equation}

Let $\mH_t$ denote the empirical Hessian and $\vv_1$ its leading eigenvector with eigenvalue $\lambda_1$.  
Preconditioning rescales each coordinate direction, so the effective curvature along $\vv_1$ becomes
\begin{equation}
    \lambda_{\text{eff}}(\vv_1) 
    \;=\;
    \frac{\vu_1^\top \mQ^{\text{mod}}_t \mH_t \mQ^{\text{mod}}_t \vv_1}
         {\vu_1^\top \vv_1}.
    \label{eq:ada:lambdaeff-def}
\end{equation}
Expanding the diagonal preconditioner,
\begin{equation}
    \lambda_{\text{eff}}(\vv_1) 
    = 
    \lambda_1
    \frac{\sum_i u_{1,i}^2}{\sum_i v_{1,i}^2 (q^{\text{mod}}_{i,t}+\epsilon)}.
    \label{eq:ada:lambdaeff-sum}
\end{equation}
This expression shows that curvature suppression depends on how $\vv_1$ aligns with coordinates heavily scaled by $\vq^{\text{mod}}_t$.  

When the leading eigenvector aligns with a single coordinate $e_i$, Eq.~\eqref{eq:ada:lambdaeff-sum} simplifies to
\begin{equation}
    \lambda_{\text{eff}} \;\approx\; 
    \frac{\lambda_1}{q^{\text{mod}}_{i,t}+\epsilon}.
\end{equation}
The classical discrete-time stability condition $\eta\,\lambda_{\text{eff}} < 2$ thus generalizes to
\begin{equation}
    \eta\,\lambda_1 < 2\,\sum_i v_{1,i}^2\,(q^{\text{mod}}_{i,t}+\epsilon).
    \label{eq:ada:instab-cond}
\end{equation}
Smaller $q^{\text{mod}}_{i,t}$ values (i.e., tighter clipping) reduce the effective damping, lowering the barrier for instability.  
Consequently, the clipping threshold $q^{\text{thresh}}$ controls the degree to which adaptive preconditioning suppresses or restores instabilities.

Equation~\eqref{eq:ada:instab-cond} provides a quantitative link between preconditioner clipping and training stability.  
When $q^{\text{thresh}}=\infty$, adaptive methods fully suppress sharp-direction updates; decreasing $q^{\text{thresh}}$ increases $\lambda_{\text{eff}}$ and can reintroduce beneficial instability, leading to flatter minima and improved generalization.

\newpage

\section{Experimental Details}\label{app:exp-dets}

This appendix summarizes the datasets, model architectures, and training procedures used in this study. 
All experiments, presented in figures or tables, are implemented with \texttt{JAX}-based training code, published in the Supplementary Material. These experiments were conducted on NVIDIA RTX A6000 GPUs. Models, with the exception of explicit mentions of mini-batching (Sections~\ref{ch:sgd}, ~\ref{ch:ada}), were trained using deterministic gradient-based optimizers without stochastic regularization, in order to isolate the effects of learning rates, curvature, and instability. 

\subsection{Datasets}

Five dataset settings were employed throughout the experiments:

\begin{itemize}
    \item \textbf{Fashion-MNIST (fMNIST). \cite{xiao2017fashionmnist}} 
    We use a subset containing the first four classes of the standard Fashion-MNIST dataset. 
    The training and evaluation sets each contained approximately $1{,}000$ samples ($\approx250$ per class). 
    Images were normalized to zero mean and unit variance. 

    \item \textbf{CIFAR-10 (5k subset). \cite{krizhevsky2009learning}} 
    A reduced training set of $5{,}000$ randomly sampled images ($\approx500$ per class) and an evaluation set of $1{,}000$ samples. 
    Standard normalization was applied, without stochastic augmentations.

    \item \textbf{CIFAR-10 (50k full). \cite{krizhevsky2009learning}} 
    The complete $50{,}000/10{,}000$ train–test split, preprocessed using standard CIFAR-10 normalization.

    \item \textbf{CIFAR-10 (500k augmented). \cite{krizhevsky2009learning}} 
    Constructed by applying ten fixed augmentations (random crops and horizontal flips) to each image in the $50{,}000$ training set, yielding $500{,}000$ training samples.

    \item \textbf{Toy / Numerical datasets.} 
    Synthetic or analytically tractable numerical settings, including two-parameter diagonal linear networks (DLNs), 
    used for simulation and phase-space visualization experiments.
\end{itemize}

\subsection{Architectures}

All architectures were intentionally kept small to facilitate exact spectral and curvature analyses.

\begin{itemize}
    \item \textbf{Multilayer Perceptron (MLP).} 
    Five layers with $32$ hidden units per layer and ReLU activations; no dropout or normalization. 
    Total parameters: $\sim 28{,}480$.

    \item \textbf{AlexNet (modified for CIFAR-10–5k).} 
    Two convolution–maxpool blocks (64 filters, $(5\times5)$ kernels, stride $(2,2)$; $(3\times3)$ pooling) followed by two dense layers 
    ($382$ and $196$ hidden units) and an output layer. 
    ReLU activations; no dropout or batch normalization.

    \item \textbf{VGG10 (small).} 
    Three VGG blocks, each with three convolutional layers, with channel widths $\{8, 16, 32\}$. 
    GhostBatchNorm (batch size $1{,}024$) and a linear two-epoch learning rate warmup. 
    $\sim 47{,}892$ parameters.

    \item \textbf{VGG19 (deep).} 
    Six VGG blocks (three convolutional layers each) with channels $\{16, 32, 64, 128, 256, 512\}$. 
    GhostBatchNorm (batch size $1{,}024$) and two-epoch warmup. 
    $\sim 335{,}277$ parameters.

    \item \textbf{ResNet20.} 
    Standard architecture following \textcite{he2015residual}, using identity skip connections. 
    GhostBatchNorm (batch size $1{,}024$), no dropout. 
    $\sim 271{,}117$ parameters.

    \item \textbf{Toy / DLN models.} 
    Two-parameter, one-layer deep linear networks used for closed-form analyses and visualization of dynamics and instability.
\end{itemize}

\subsection{Training and Evaluation Procedures}

Training followed either full-batch (FB) or mini-batch (MB) gradient descent, depending on the experiment. 

\begin{itemize}
    \item \textbf{Optimizers.} 
    Deterministic full-batch gradient descent (FB) or mini-batch SGD without momentum, RMSProp, or Adam. 

    \item \textbf{Learning rate and warmup.} 
    A linear warmup of two epochs was applied for convolutional networks (VGG, ResNet). 
    Learning rates were varied systematically to explore stable, transitional, and unstable regimes.

    \item \textbf{Epoch compensation.} 
    For fMNIST experiments, smaller learning rates ($\eta \leq 0.05$) were compensated with proportional increases in epoch count to ensure comparable optimization trajectory lengths.

    \item \textbf{Early stopping.} 
    Validation loss was monitored post hoc, and stopping points were determined by the first reversal in validation loss after the global minimum, 
    ensuring consistent comparison across seeds.

    \item \textbf{Hessian computation.} 
    Hessian–vector products were obtained using \cite{pearlmutter1997fastexact}'s trick, and spectra computed via the Lanczos algorithm 
    \parencite{Lanczos1950method} with $100$ iterations and full re-orthogonalization. 
    The implementation adapted the \texttt{spectral-density} baseline provided by Google, accelerated using \texttt{JAX}.

    \item \textbf{Reproducibility.} 
    Each configuration was trained with $3$--$5$ random seeds. Divergent trajectories (under large $\eta$, observed post-hoc) were excluded from metric correlations.
\end{itemize}

\subsection{Hardware and Runtime}

All experiments were executed on a workstation equipped with four NVIDIA RTX A6000 GPUs. 
Average GPU-hour requirements ranged from negligible (toy models) to $\sim100$~h for the lowest learning-rate VGG experiments. Total compute time in the order of magnitude of $\sim1{,}000$s of hours.
Scripts and data generation notebooks are available in the accompanying code release.

\newpage

\bibliographystyle{elsarticle-harv}
\bibliography{references}




\end{document}